%% file: main.tex
\documentclass[10pt,twocolumn,letterpaper]{article}
\usepackage[pagenumbers]{cvpr} %

\input{preamble}

\input{defs}

\begin{document}
    
    \maketitle
    
    \input{sec/0_abstract}   
    \input{sec/1_intro}

    \input{sec/2_related_work}

    \input{sec/3_background}

    \input{sec/4_method}

    \input{sec/5_experiments}
    \input{sec/6_conclusion}

    \section*{Acknowledgements}
    T. Birdal was supported by a UKRI Future Leaders Fellowship [grant number MR/Y018818/1].
    S. Foti and S. Zafeiriou were supported by the EPSRC Project GNOMON (EP/X011364/1) and the Turing AI Fellowship MAGAL (EP/Z534699/1).
    
    {
        \small
        \bibliographystyle{ieeenat_fullname}
        \bibliography{main,suppl}
    }
    
    \input{sec/X_suppl}

\end{document}

%% file: preamble.tex
\usepackage{xspace}

\usepackage{soul}
\setuldepth{foobar}

\usepackage{graphicx}
\usepackage{tikz}
\usepackage{multirow}
\usepackage{comment}
\usepackage{amsmath,amssymb,amsthm,mathtools} %
\usepackage{booktabs}
\usepackage{bm}
\usepackage{color}
\usepackage[dvipsnames]{xcolor}
\usepackage{fancyvrb}
\usepackage{algorithm,algpseudocode}
\usepackage{mathrsfs}
\algrenewcommand\algorithmicrequire{\textbf{In:}}
\algrenewcommand\algorithmicensure{\textbf{Out:}}

\usepackage{array} %

\usepackage{wrapfig}

\definecolor{cvprblue}{rgb}{0.21,0.49,0.74}
\definecolor{myteal}{RGB}{128, 166, 192}
\definecolor{mypink}{RGB}{241, 109, 161}
\usepackage[pagebackref,breaklinks,colorlinks,allcolors=cvprblue]{hyperref}

\usepackage[capitalize]{cleveref}

\usepackage{lineno}

\newcommand{\samethanks}[1][\value{footnote}]{\footnotemark[#1]}

\def\paperID{32370}
\def\confName{CVPR}
\def\confYear{2026}

\title{
Parallelised Differentiable Straightest Geodesics for 3D Meshes
}

\author{Hippolyte Verninas \quad Caner Korkmaz \quad Stefanos Zafeiriou \quad Tolga Birdal\thanks{Joint senior authors, equal contribution} \quad Simone Foti\samethanks \\
Imperial College London \\
}

%% file: defs.tex
\renewcommand{\vec}[1]{\boldsymbol{#1}}
\newcommand{\mat}[1]{\mathbf{#1}}

\newcommand{\x}{\mathbf{x}}
\newcommand{\p}{\mathbf{p}}
\newcommand{\y}{\mathbf{y}}

\newcommand{\Y}{{\mathbf{Y}}}

\newcommand{\R}{\mathbb{R}}
\newcommand{\Nat}{\mathbb{N}}

\newcommand{\z}{\mathbf{z}}

\newcommand{\N}{\mathcal{N}}
\newcommand{\q}{\mathbf{q}}

\renewcommand{\u}{\mathbf{u}}
\renewcommand{\v}{\mathbf{v}}
\newcommand{\J}{\mathbf{J}}
\newcommand{\w}{\mathbf{w}}

\newcommand{\V}{\mathbf{V}}

\newcommand{\versor}[2][e]{\hat{\mathbf{#1}}_{#2}}

\newcommand{\Vertices}{\mathbf{X}}

\newcommand{\Faces}{\mathbf{F}}

\newcommand{\mesh}{\mathscr{M}}
\newcommand{\vertex}[1][i]{\mathbf{x}_{#1}}

\newcommand{\face}[1][i]{\mathbf{f}_{#1}}
\newcommand{\faceindex}{\mathrm{f}}
\newcommand{\vertexindex}{\mathrm{x}}
\newcommand{\bary}{\mathbf{b}}

\newcommand{\curve}{\gamma}
\newcommand{\dcurve}{\dot{\gamma}}
\newcommand{\geo}{\curve_\v}
\newcommand{\dgeo}{\dcurve_\v}

\newcommand{\Man}{\mathcal{M}}

\newcommand{\TpM}{\T_{\p}\Man}

\newcommand{\T}{\mathcal{T}}

\newcommand{\Exp}{\mathrm{{Exp}}}
\newcommand{\Log}{\mathrm{{Log}}}

\newcommand{\connection}{\nabla}
\newcommand{\covderiv}[2]{\connection_{#1} {#2}}
\newcommand{\curvevecfield}{\mathcal{W}_{\curve}}

\crefname{equation}{eq.}{eq.}
\Crefname{equation}{Eq.}{Eq.}
\crefname{theorem}{thm.}{thms.}
\Crefname{Theorem}{Thm.}{Thms.}
\crefname{conjecture}{conj.}{conjs.}
\Crefname{Conjecture}{Conj.}{Conjs.}
\crefname{proposition}{prop.}{props.}
\Crefname{proposition}{Prop.}{Props.}
\crefname{definition}{dfn.}{dfn.}
\Crefname{definition}{Dfn.}{Dfn.}
\crefname{remark}{remark}{remark}
\Crefname{Remark}{Remark}{Remark}
\Crefname{algorithm}{Alg.}{Alg.}

\crefname{section}{Sec.}{Secs.}
\Crefname{section}{Sec.}{Secs.}
\crefname{equation}{Eq.}{Eqs.}
\Crefname{equation}{Eq.}{Eqs.}
\crefname{figure}{Fig.}{Figs.}
\Crefname{figure}{Fig.}{Figs.}
\crefname{table}{Tab.}{Tabs.}
\Crefname{table}{Tab.}{Tabs.}
\crefname{thm}{Thm.}{Thms.}
\Crefname{thm}{Thm.}{Thms.}
\crefname{conj}{Conj.}{Conjs.}
\Crefname{conj}{Conj.}{Conjs.}
\crefname{dfn}{Dfn.}{Dfns.}
\crefname{dfn}{Dfn.}{Dfns.}
\crefname{remark}{remark}{remarks}
\Crefname{Remark}{Remark}{Remarks}
\crefname{prop}{Prop.}{Prop.}
\Crefname{prop}{Prop.}{Prop.}
\Crefname{algorithm}{Alg.}{Alg.}
\crefname{appendix}{App.}{Apps.}
\Crefname{appendix}{App.}{Apps.}
\crefname{appsec}{appendix}{appendices}
\Crefname{appsec}{Appendix}{Appendices}

\renewcommand{\paragraph}[1]{{\vspace{0.6mm}\noindent \bf #1}.}

\newlength{\savedintextsep}
\newlength{\savedcolumnsep}

\newcommand{\StoreParams}{
    \setlength{\savedintextsep}{\intextsep}
    \setlength{\savedcolumnsep}{\columnsep}
}

\newcommand{\StartTightFigureParagraph}{
    \StoreParams
    \setlength{\intextsep}{5pt}  
    \setlength{\columnsep}{7pt}
}

\newcommand{\EndTightFigureParagraph}{
    \setlength{\intextsep}{\savedintextsep}
    \setlength{\columnsep}{\savedcolumnsep}
}

\newlength{\arrowwidth}
\newcommand{\FixedRightArrow}[1]{
    \xrightarrow{\makebox[\arrowwidth]{\scriptsize $#1$}}
}

%% file: sec/0_abstract.tex
\begin{abstract}
    Machine learning has been progressively generalised to operate within non-Euclidean domains, but geometrically accurate methods for learning on surfaces are still falling behind. The lack of closed-form Riemannian operators, the non-differentiability of their discrete counterparts, and poor parallelisation capabilities have been the main obstacles to the development of the field on meshes. 
    A principled framework to compute the exponential map on Riemannian surfaces discretised as meshes is straightest geodesics, which also allows to trace geodesics and parallel-transport vectors as a by-product.
    We provide a parallel GPU implementation and derive two different methods for differentiating through the straightest geodesics, one leveraging an extrinsic proxy function and one based upon a geodesic finite differences scheme. 
    After proving our parallelisation performance and accuracy,
    we demonstrate how our differentiable exponential map can improve learning and optimisation pipelines on general geometries.
    In particular, to showcase the versatility of our method, we propose a new geodesic convolutional layer, a new flow matching method for learning on meshes, and a second-order optimiser that we apply to centroidal Voronoi tessellation. 
    Our code, models, and pip-installable library ($\mathtt{digeo}$) are available at: \href{https://circle-group.github.io/research/DSG}{circle-group.github.io/research/DSG}.

\end{abstract}

%% file: sec/1_intro.tex
\section{Introduction}
    \label{sec:intro}

\begin{figure}[t]
        \centering
        \vspace{-9px}
        \includegraphics[width=1\linewidth]{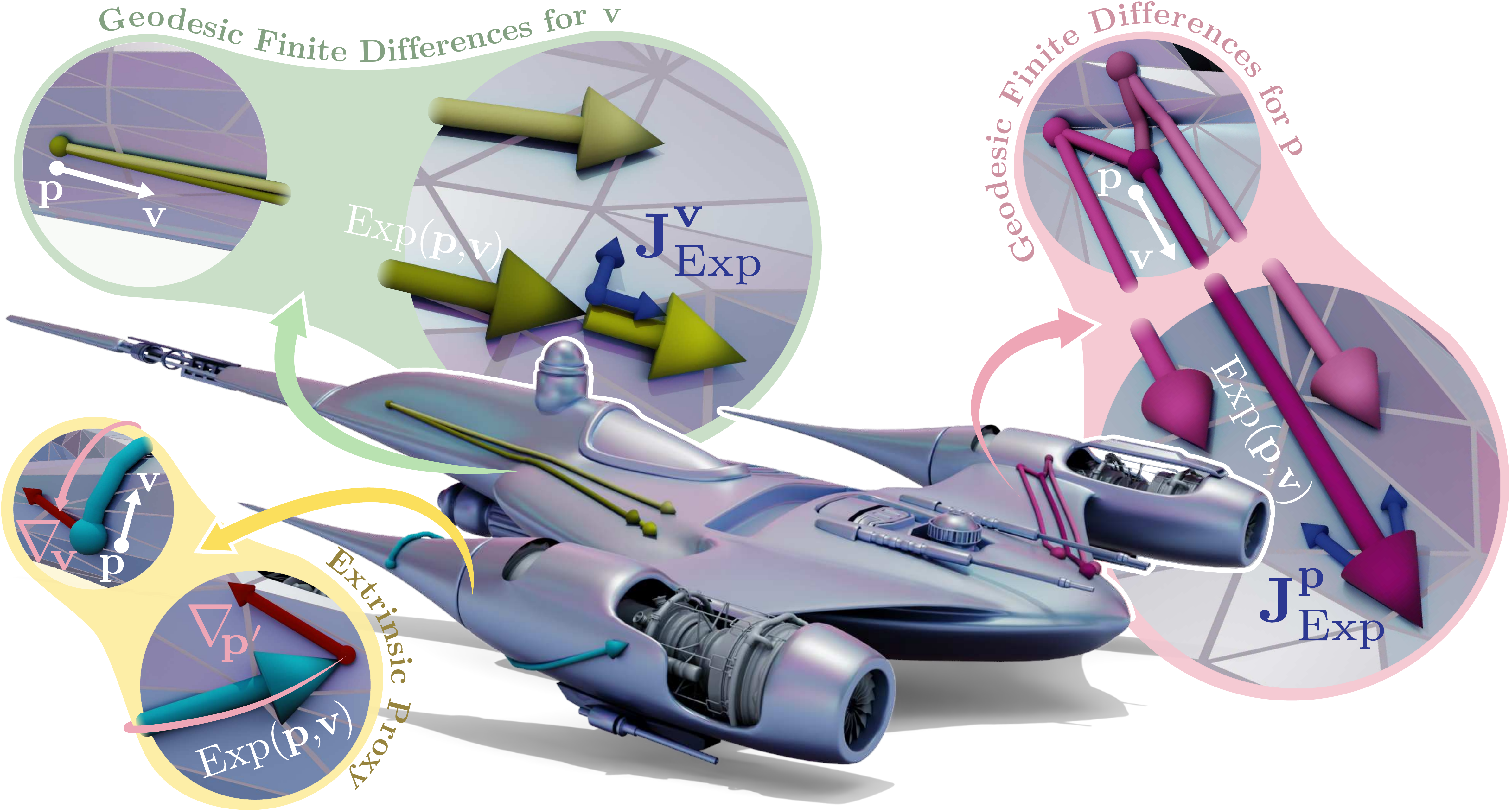}
        \vspace{-16px}
        \caption{
        Our GPU-parallelised schemes to differentiate the Exponential map and improve learning and optimisation on meshes: the Extrinsic Proxy (EP) and Geodesic Finite Differences (GFD).
        \vspace{-5mm}}
        \label{fig:teaser}
    \end{figure}

    What do molecules, proteins, the Earth, joint articulations, orthonormal matrices and a spaceship (like the one in \cref{fig:teaser}) have in common? They all have fundamental properties that can be modelled through Riemannian geometry. Tremendous progress has recently been made possible by integrating the geometry of these spaces into learning~\cite{bose2023se, bonev2025fourcastnet,yu2025nrmf, huguet2024sequence,yim2023fast, morehead2025flowdock,he2024nrdf} and by the availability of software libraries providing Riemannian operators for analytically defined spaces admitting closed-form-solutions~\cite{townsend2016pymanopt, miolane2020geomstats, meghwanshi2018mctorch, kochurov2020geoopt}. 
    Significant efforts have also been made to learn on surfaces~\cite{foti2024uv, diffusionnet, gcnn, mdgcnn,bouritsas2022improving, pointnet++, riemannian-fm}, but when Riemannian operators are needed, they are pre-computed~\cite{gcnn, mdgcnn}, or replaced by slow numerical solvers~\cite{riemannian-fm}. 
    In this work, we take an important step towards enabling the adoption of Riemannian operators directly into learning and optimisation methods for meshes.  

    We focus on developing a differentiable exponential map, arguably the most important Riemannian operator,
    a function that takes a tangent vector at a starting point and finds the end point of the geodesic traced according to that vector.
    While closed-form solutions of the exponential map exist for simple canonical shapes like spheres~\cite{lee2006riemannian}, more complex shapes require the use of advanced computational techniques. On dscrete surfaces, tracing the geodesic curve that determines the solution of the exponential map requires numerically solving the initial value problem, a system of second order differential equations, later detailed in \cref{sec:bg}.
    Unfortunately, this approach not only is rather cumbersome~\cite{cheng216ivp}, but also computationally inefficient and primarily suited for smooth continuous surfaces. In computing, most surfaces are discretised into meshes and different techniques are needed~\cite{polthier2006straightestgeod, crane2020survey, cheng216ivp}. To date, the straightest geodesics method introduced in~\cite{polthier2006straightestgeod} is the de-facto choice for computing the exponential map and geodesic tracing on triangular meshes. All available implementations are non-differentiable and conceived to operate sequentially on the CPU. 
    Consequently, the existing formulation of the exponential map is not suitable for integration within modern learning and optimisation frameworks for 3D meshes. 
     
    We derive two different implementations of the differential (\cref{fig:teaser}): a faster version using an extrinsic proxy (EP) and a slower yet more accurate geodesic finite-differences (GFD) scheme. 
    EP uses a proxy function to emulate the tracing result in Euclidean space, thus providing an analytical formulation that can be easily differentiated. GFD is based upon finite differences, which we adapt to respect the geometry of the mesh. Both schemes are agnostic to the forward computation of the exponential map, and can provide differentiability to any exponential map.  
    Additionally, we propose a parallel GPU implementation of the straightest geodesics that can handle tens of thousands of points on multiple meshes, yielding results in just a few milliseconds.
    We thoroughly benchmark the speed and accuracy of both our forward and backward step against existing implementations and closed-form exponential maps.

    Then, we focus on demonstrating our methods by proposing three new competitive applications. 
    We introduce a novel geodesic convolution that dynamically learns the patch size during training. Our convolution outperforms fixed-patch methods and most leading intrinsic techniques. We then propose a flow matching-inspired method for arbitrary meshes that reduces inference time by orders of magnitude while using a fraction of GPU memory, and outperforming the state-of-the-art. Finally, we introduce a second order LBFGS-based~\cite{rbfgs1, rlbfgs-imp} Riemannian optimiser for 3D meshes that we use for computing the centroidal Voronoi tessellation~\cite{cvt}. Also in this case our method achieves faster convergence and superior minimisation performance. 
    It is worth noting that beyond our proposed applications, we expect our contributions to benefit a wide variety of domain-specific applications. 
    
    \noindent To summarise, our key contributions are:
    \begin{enumerate}[noitemsep, leftmargin=*, topsep=0.1pt]
        \item A \textbf{differentiable straightest geodesics} method to compute exponential maps, trace geodesics, and parallel-transport vectors on the surface of 3D meshes that is efficiently parallelised on the GPU.  %
        \item A \textbf{PyTorch-compatible software library featuring C++ CUDA kernels} that enables seamless integration of discrete Riemannian geometry into modern learning pipelines for meshes. 
        \item \textbf{Adaptive Geodesic Convolutions (AGC)} that learn their patch size for each channel and layer.
        \item \textbf{MeshFlow}, a new flow-matching-inspired method based on our exponential map and optimal transport. 
        \item \textbf{Mesh-LBFGS}, a second order LBFGS optimiser for meshes that we use to efficiently solve the centroidal Voronoi tesselation problem. 
    \end{enumerate}

%% file: sec/2_related_work.tex
\section{Related Work}
    \label{sec:related_work}

   Numerous existing works on geodesics has focused on solving the Boundary Value Problem (BVP) to compute the shortest paths between two points on a discrete surface~\cite{balasubramanian2008exact, o1985shortest, mitchell1987discrete, chen1990shortest, surazhsky2005fast, fast-marching, heat-geodesics, crane2017hmd, huberman2023deep, lichtenstein2019deep}. Widely researched are also algorithms for computing the logarithm map and create local parametrisations~\cite{vectorheat, soliman2025affine, herholz2019efficient, schmidt2006interactive}, which are particularly useful for texturing. As noted in \cite{soliman2025affine}, the graphics community has often misused the logarithm and exponential map terminology. We here adhere to the Riemannian geometry definitions. 
    Therefore, when looking at the IVP literature for tracing straightest geodesics and computing the exponential map, we note that it is quite prolific on continuous surfaces and considerably under-explored on meshes. 
    Iterative integration of the IVP ODE~\cite{ying2006geodesic-ode} is possible on smooth continuous surfaces, but computationally inefficient. For this reason, closed-form solutions are always preferred when available, like on $\mathrm{SO}(3)$~\cite{yu2025nrmf, he2024nrdf}, $\mathrm{SE}(3)$~\cite{bose2023se}, or the sphere~\cite{mathieu2020riemannian, riemannian-fm}. These and other equally well-defined simple spaces are readily available in famous libraries for Riemannian geometry~\cite{townsend2016pymanopt, miolane2020geomstats, meghwanshi2018mctorch, kochurov2020geoopt}. 
    
    On meshes, \cite{polthier2006straightestgeod} still represents the gold standard. The isolated attempt to improve \cite{polthier2006straightestgeod} by approximating the continuous surface discretised by the mesh is more computationally expensive, still non-differentiable, and produces off-mesh results. Beyond meshes, \cite{genest2025implicit} extended exponential maps to implicit functions. Its approximate nature and a mandatory representation conversion make it unsuitable for meshes. Finally, \cite{Madan2025local-parameterizations} introduced a representation-agnostic exponential map obtained through an iterative projection from the tangent space. However, their step-size selection imposes a critical trade-off between accuracy and speed.

%% file: sec/3_background.tex
\section{Riemannian Geometry}
    \label{sec:bg}

    In this section we provide a self-contained summary of the Riemannian geometry concepts related to our work and recommend \cite{lee2006riemannian, guigui2023introriemannian} for a broader review of the field.
    
    Let $\Man \subset \R^3$ be a Riemannian manifold embedded in 3D Euclidean space and $\T\Man = \bigcup_{\p \in \Man} \TpM$ be the tangent bundle set encompassing the tangent spaces of all points $\p$ on $\Man$. By definition, a manifold $\Man$ is locally homeomorphic to $\TpM \subset \R^3$. A curve $\curve: I \rightarrow \Man$ parametrised over an interval $I =[a, b]$ (with ${a,b} \in \R$), is a geodesic on a smooth Riemannian surface when it can be simultaneously characterised as being the shortest and straightest. Since for triangular meshes straightest and locally shortest geodesics differ from each other~\cite{polthier2006straightestgeod}, and this work focuses on the former, we here define geodesics according to the straightness criteria also in the continuous setting. To define straightness we need a way to characterise the absence of intrinsic curvature. In $\R^3$ we would just impose $\ddot{\curve} = 0$, but if we attempt to write $\ddot{\curve}$ as $\lim_{\Delta t \rightarrow 0} \frac{\dcurve(t +\Delta t) - \dcurve(t)}{\Delta t}$ it becomes immediately obvious that the tangent vectors $\dcurve (t+ \Delta t)$ and $\dcurve (t)$ cannot be directly compared as they belong to two different tangent planes, $\T_{\p_1}\Man$ and $\T_{\p_0}\Man$, with $\p_1= \curve(t + \Delta t)$ and $\p_0= \curve(t)$. Therefore, to correctly compute this second derivative we need to take into account how tangent vectors $\dcurve(t)$ turn as they move along the tangent direction $\dcurve(t)$. This can be achieved with the Covariant derivative, which is usually computed using the metric preserving and torsion-free Levi-Civita connection $\connection$. The Riemannian equivalent of $\ddot{\curve} = \frac{d^2 \curve(t)}{dt^2} = \frac{d}{dt} \dcurve(t)$ can therefore be rewritten as $\nabla_{\dcurve(t)} \dcurve(t)$ and it equals zero $\forall t \in [a,b]$ if $\gamma(t)$ is a geodesic. This intrinsic formulation could also be reframed from an extrinsic point of view by allowing $\curve$ to have curvature only in the normal direction of $\Man$. 

    A fundamental problem in Riemannian geometry is to identify the unique geodesic curve $\geo(t):[a, b] \rightarrow \Man$ originating in $\p \in \Man$ given an initial tangent velocity $\v \in \TpM$. This amounts to solving the initial value problem (IVP) represented by the following second order system of ordinary differential equations:
    \begin{equation}
        \label{eq:ivp}
        \geo(t) = 
            \begin{cases}
                \nabla_{\dcurve(t)} \dcurve(t) = 0 \\
                \curve(a) = \p \\
                \dcurve(a)=\v
            \end{cases}
    \end{equation}
    The \emph{exponential map} $\Exp_\p: \TpM \rightarrow \Man$ is the map that computes the end-point of the geodesic obtained by solving the IVP. In other words, given $\p \in \Man$ and a vector $\v \in \TpM$, the exponential map determines the point on the manifold obtained walking until the end of the geodesic originating in $\p$ with direction $\frac{\v}{\|\v\|}$ and length $\|\v\|$. Therefore we have: $\Exp_\p (\v) = \geo(b)$.
    The inverse of the exponential map is called \emph{logarithm map} and is defined over a neighbourhood $\N(\p)$ of $\p \in \Man$ where $\Exp_\p$ is invertible. Given any point $\p' \in \N(\p)$, $\Log_\p : \N(\p) \in \Man \rightarrow \TpM$ finds the shortest vector $\v$ such that $\Exp_\p (\v) = \p'$. 

    As previously mentioned, the Levi-Civita connection provides a rule for comparing tangent vectors at different points, thereby defining the notion of \emph{parallel transport}. 
    A vector field along $\curve$ is a smooth map $\curvevecfield: I \rightarrow \T_{\!\curve(t)}\Man$ and is considered parallel if it is covariantly constant: $\covderiv{\dcurve(t)}{\curvevecfield} = 0$. 
    When $\nabla$ is Levi-Civita, for every $\mathcal{C}^1$ curve $\gamma$, any $t_0 \in I$, and any initial vector $\w \in \T_{\curve(t_0)}\Man$ there exists a unique parallel vector field $\curvevecfield$ satisfying the initial condition $\curvevecfield(t_0) = \w$. 
    Given $\{t_0, t_1\} \in I$, the parallel transport of $\w$ along $\curve$ from $t_0$ to $t_1$ is thus written as $\prod_{\curve,t_0}^{t_1} = \curvevecfield(t_1)$ and is computed by simultaneously solving the system of differential equations for the geodesic curve (\cref{eq:ivp}) and the parallel vector field therein defined.

%% file: sec/4_method.tex
\section{Differentiable Straightest Geodesics}
    \label{sec:dpsg}

    After detailing the computation of $\geo$, $\Exp_\p (\v)$, and $\prod_{\curve,t_0}^{t_1}$ on meshes using straightest geodesics~\cite{polthier2006straightestgeod}, we introduce our two methods for differentiating through $\Exp_\p (\v)$ and our GPU parallelisation strategy. Our differentiable and efficient operators build the foundations for a differentiable Riemannian geometry stack on meshes.

    Instead of continuous manifolds $\Man$ we now consider meshes $\mesh = \{ \Vertices, \Faces \}$, with $\Vertices \in \R^{N \times 3} = [\vertex[1] \dots \vertex[N] ]^T$ representing the $N$ vertices $\vertex = [x_i, y_i, z_i]^\top$, and $\Faces \in \Nat^{F \times 3}$ the $F$ triangular faces. Values in each $\face \in \Faces$ index the vertices to define the mesh's connectivity. 
    The mesh is \emph{manifold} if each edge is incident to either one or two faces, and faces around each vertex form a single connected fan.
    Points on the surface of $\mesh$ are defined by the pair $(\faceindex, \bary)$, where $\faceindex \in \Nat$ is the index of the face on which a point lies, and $\bary = [b_u, b_v, b_w]^T$ its \emph{barycentric coordinates} on $\faceindex$. These coordinates belong to the 2-simplex $\Sigma_2 = \big\{ \bary \in [0,1]^3 | \sum_{k = (u, v, w)} b_k = 1 \big\}$. Given $\vertex[0], \vertex[1], \vertex[2]$ as the vertices of $\face[\faceindex]$, the coordinates of $\p \in \mesh$ are $\p = b_u \vertex[0] + b_v \vertex[1] + b_w \vertex[2]$, where $b_w = (1 - b_u - b_v)$ according to the definition of $\Sigma_2$.
    
\begin{wrapfigure}[8]{R}[0pt]{0.49\linewidth}
            \centering
            \vspace{-17pt}
            \includegraphics[width=\linewidth]{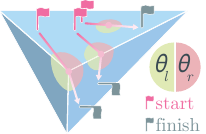}
            \vspace{-20pt}
            \caption{Left $\theta_l$ and right $\theta_r$ angles on a vertex, edge, and face.}
            \label{fig:angles}
        \end{wrapfigure}

    \StartTightFigureParagraph
    \paragraph{Straightest geodesics} 
        Given a point $\p \in \mesh$ and a vector $\v \in \T_\p \mesh$, the straightest geodesic algorithm uses the angle preservation criteria (\cref{fig:angles}) to iteratively trace a geodesic $\geo$ from one face onto another.  Like in the continuous setting $\Exp_\p (\v) = \geo(b)$.  
        
        The discrete analogue to imposing $\nabla_{\dcurve(t)} \dcurve(t) = 0$ is to impose left and right curve angles ($\theta_l, \theta_r$) to be the same for every $\p$ along $\curve(t)$~\cite{polthier2006straightestgeod}. 
        $\theta_l$ and $\theta_r$ are the angles that incoming and outgoing segments of the curve $\gamma$ form with the total angle space $\theta(\p)$ at point $\p$. 
        While the straightness condition is $\theta_l = \theta_r, \forall \p \in \mesh$, we need to distinguish how the total angle definition changes in the three separate cases depicted in \cref{fig:angles}: when a point is within a face, on an edge, or on a vertex.
        For all points within faces we are effectively on a plane, so the total angle is $\theta(\p) = 2 \pi$.
        Considering that two connected faces can easily be made coplanar, the neighbourhood of a point on an edge is isometric to a plane and thus $\theta(\p) = 2 \pi$ also for all points on edges. For vertices, on the other hand, the total angle is the sum of all interior angles of the $m$ faces meeting at the vertex: $\theta(\p) = \sum_{i=1}^m \theta_i(\p)$. Therefore, at vertices, $\theta(\p) = 2 \pi$ if all faces connected to the vertex are coplanar, $\theta(\p) < 2\pi$ for spherical (cone-like) vertices, and $\theta(\p) > 2\pi$ for hyperbolic (saddle-like) vertices.   
        Further details on the implementation of the geodesic step are provided in \cref{alg:sg-step}. The approach exploits the extrinsic geometry to accelerate computations, relying on the normal vectors of faces and vertices. This does not pose any disadvantage, since 3D meshes are almost always embedded in a Euclidean space with a canonical frame.

        Note that the original algorithm of \cite{polthier2006straightestgeod} stops tracing geodesics when reaching the boundary of the mesh $\partial \mesh$ (i.e., when reaching an edge incident only to one face or a vertex belonging to it) even if the desired length $|\v\|$ is not reached. While correct, we notice that this behaviour can negatively affect performance during learning and optimisation in the presence of small holes and defects (\cref{sec-supp:otexpflow}). We thus give the option to temporarily interrupt the straightness criteria, trace along the boundary, and resume tracing as soon as possible with the last straightest direction recorded when $\partial \mesh$ was reached.   

    \noindent\textbf{Parallel Transport}
        can be defined via
        the discrete equivalent of $\covderiv{\dcurve(t)}{\curvevecfield} = 0$, which corresponds to imposing the normalised angle $\alpha$ between the vector field $\curvevecfield$ and the curve's tangent vector $\dcurve(t)$ to be constant $\forall t \in I$, with $\alpha \big(\curvevecfield(t), \dcurve(t) \big) := \frac{2\pi}{\theta(\p)} \angle \big(\curvevecfield(t), \dcurve(t) \big)$~\cite{polthier2006straightestgeod}. Given $\{t_0, t_1\} \in I$, the parallel transport of $\w$ along $\curve$ from $t_0$ to $t_1$, $\prod_{\curve,t_0}^{t_1}$, is obtained while iteratively tracing the geodesic. When crossing vertices $\curvevecfield$'s orientation is corrected according to $\alpha$. When crossing an edge, $\curvevecfield$'s orientation is simply maintained relative to the unfolded plane because $\theta(\p) = 2 \pi$ and the normalised angle is then identical to the Euclidean angle $\angle$.

    \EndTightFigureParagraph

    \paragraph{Differentiation} %
        As shown in \cite{chen2022projective}, automatic differentiation in Riemannian settings is problematic in nature because naive Euclidean gradients produce off-manifold results. Furthermore, the piecewise linear nature of straightest geodesics on $\mesh$ results in a discontinuous derivative ($\dgeo(t) \notin \mathcal{C}^1$ at face transitions), which inherently prevents automatic back-propagation through $\Exp_\p (\v)$ and makes its adoption in optimisation or learning problems infeasible without an explicit differentiation scheme. More details on non-differentiability are provided in \cref{sec-supp:geostep}. 
        
        Given a composite function, $h(\z)=f(\Exp(g(\z)))=\y$, sending an input $\z \in \R^{n_z}$ to $\y \in \mathbb{R}^{n_y}$, we have:
        \begin{equation}
            h: \R^{n_z} \FixedRightArrow{g} (\mesh \times \T\mesh) \FixedRightArrow{\Exp} \mesh \FixedRightArrow{f} \mathbb{R}^{n_y}.
        \end{equation}
        $h(\z)$ represents the general scenario in which $\Exp$ is used within a more complex framework involving other differentiable functions $f$ and $g$ , which could also be parametrised by neural networks. If we need to differentiate through $h$ we can use the chain rule. Indicating with $\J \in \R^{m \times n}$ the Jacobian matrix 
        whose rows are the $m$ transposed gradients ($\nabla \in \R^n$) of the function's $m$ output components, we have:
        \begin{equation}
            \label{eq:chain}
            \J_h = \J_f \cdot \big[ \J_\Exp^\p \cdot \, \J_g^\p + \J_\Exp^\v \cdot \, \J_g^\v \big],
        \end{equation}
        where each Jacobian is computed with respect to the input of the function it is referred to. If a function has either multiple inputs or multiple outputs, the superscript is used to select a specific input/output sub-block of the full Jacobian.
        If a function is not used, its Jacobian is the identity matrix, making the formulation of $h$ valid also when $f$ and $g$ are not used.
        When a loss function $\mathcal{L}$ is computed on $\y = h(\z)$, the full chain rule becomes $\nabla_\z \mathcal{L} = \nabla_\y \mathcal{L} \cdot \, \J_h(\z)$.
        Nevertheless, what emerges from \cref{eq:chain} is that we need $\J_\Exp^\v$ and $\J_\Exp^\p$ to enable full differentiability of $\Exp_\p (\v)$. 
        Refer to \cref{fig:teaser} and the corresponding insert figures for a visual description of all the differentiation schemes detailed hereafter.
        
    \StartTightFigureParagraph
    \paragraph{Differentiation via Extrinsic Proxy (EP)} 
        The first of our two proposed differentiation schemes uses a proxy function $\varphi(\p, \v)$ that leverages $\Exp (\p, \v)$ ---computed with the non-differentiable straightest geodesics method--- to define a proxy endpoint $\p'_{\text{proxy}} (=\p')$ that 
        we can differentiate. In particular, $\varphi$ is defined as:
        \begin{equation}
            \label{eq:proxy_func}
            \varphi(\p, \v) = \p'_{\text{proxy}} = \big[\mat{R}^\text{fix} \cdot (\p + \v) \big] + \vec{t}^\text{fix}.
        \end{equation}
        
        \begin{wrapfigure}[5]{L}[0pt]{0.45\linewidth}
            \centering
            \vspace{-12pt}
            \includegraphics[width=\linewidth]{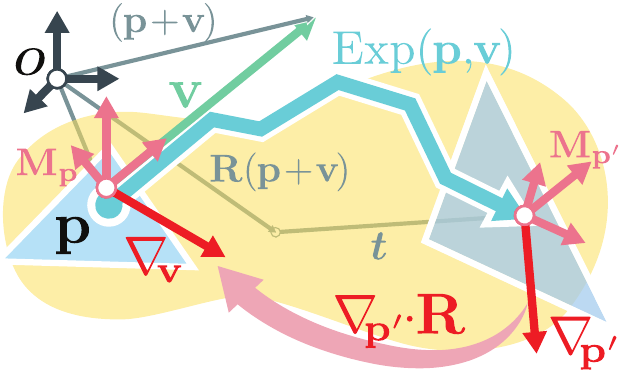}
        \end{wrapfigure}
        \noindent Here, $(\p + \v)$ is the Euclidean movement of $\p$ according to $\v$. $\mat{R} = \mat{M}_\p^T \mat{M}_{\p'}$ represents the total cumulative rotation accumulated by the geodesic path as it travels along the curved surface and acts as a differentiable surrogate for parallel transport in Euclidean space. It is computed as the rotation between the orthonormal frames defined at the start and end of the geodesic. 
        Here, the axes of the starting frame $\mat{M}_\p$ (and columns of the matrix) are: $\versor{\parallel} = \frac{\v}{\|\v\|}$, $\versor{\perp} = \versor[n]{} \times \versor{\parallel}$, and $\versor[n]{}$ ---which is the normal to the tangent plane where $\v$ is defined. $\mat{M}_{\p'}$ is built following the same procedure, but replacing $\v$ with its parallel transported version in $\p'$, which we indicate as $\prod_{\p}^{\p'}(\v)$ using a small abuse of notation to improve clarity in the discrete setting.
        Finally, $\vec{t} = \p' - \mat{R} \cdot (\p + \v)$ is the fixed bias translation vector that forces $\varphi$ to perfectly match the position of $\p'$ and the ``fix" superscript indicates a gradient detachment. %

        Note that the extrinsic proxy is specifically constructed to model differential changes of $\v$. In fact, if we differentiate $\varphi$ with respect to $\v$ we obtain $\J_\Exp^\v = \mat{R}^\text{fix} \cdot \mat{I}$. Therefore, %
        in the backward pass, $\mat{R}^\text{fix}$ backward parallel transports the upstream Jacobian (e.g. $\J_f$ in \cref{eq:chain}) from $\p'$ to $\p$.
        Unfortunately, if we differentiate $\varphi$ with respect to $\p$, the result is unchanged, showing that changes in the initial position would also be treated as rigid rotations while completely ignoring how the geodesic shifts when its starting point $\mathbf{p}$ changes. Observing \cref{eq:chain}, it becomes obvious that this incorrectness can jeopardise the optimisation even for problems that seek to optimise only with respect to the $\v$ component of $\Exp(\p, \v)$. Therefore, we void the contributions of $\p$ by zeroing the elements of $\J_\Exp^\p$. Instead of computing the proxy and letting $\mathtt{autograd}$ differentiate $\varphi$, we directly use $\Exp$ in the forward step and use the manually computed Jacobian of $\varphi$ in the backward step.

    \paragraph{Differentiation via Geodesic Finite Differences (GFD)}
        \begin{wrapfigure}[6]{R}[0pt]{0.5\linewidth}
            \centering
            \vspace{-10pt}
            \includegraphics[width=\linewidth]{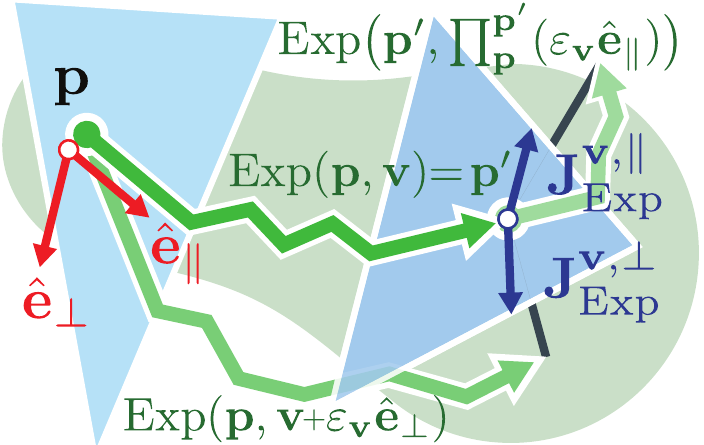}
        \end{wrapfigure}
        To introduce our finite differences scheme, we first define the local reference frames of our intial conditions. For vectors $\v$ in the tangent bundle we use $\versor{\parallel}$ and $\versor{\perp}$, as defined for the EP differentiation approach.
        For points, we define a non-orthogonal barycentric coordinates system as $\versor{u} = \frac{\vertex[1] - \vertex[0]}{\| \vertex[1] - \vertex[0]\|}$ and $\versor{v} = \frac{\vertex[2] - \vertex[0]}{\| \vertex[2] - \vertex[0]\|}$. This system is consistent with the barycentric definition of points and simplifies perturbations for finite differences. Since it is still defined in the tangent of a face, it also allows to project from ambient space ($\R^3$) onto the local frame $\mat{M} = [\versor{u} \; \versor{v}] \in \R^{3\times2}$ by using its pseudoinverse (Moore–Penrose inverse) $\mat{M}^\dagger = (\mat{M}^T\mat{M})^{-1}\mat{M}^T$.
        
        Given an extrinsically defined $\p$ and a vector $\v$ in its local reference system, we have $\J_\Exp^\v \in \R^{2\times2}$, where columns correspond to perturbations along the axes of the local reference system. In order to apply a forward finite differences scheme we need to introduce perturbations along the axes of $\v$'s frame and project the Jacobian in ambient space to the local frame:
        \begin{equation}
            \label{eq:jv}
            \J_\Exp^\v \approx \mat{M}^\dagger_{\p'} \cdot
                \Big[ 
                    \frac{\scriptstyle \Exp(\p, \v + \varepsilon_\v \versor{\parallel}) - \p'}{\varepsilon_\v}
                    \;
                    \frac{\scriptstyle \Exp(\p, \v + \varepsilon_\v \versor{\perp}) - \p'}{\varepsilon_\v} 
                \Big]
        \end{equation}
        where $\varepsilon_\v$ is a small constant and $\mat{M}^\dagger_{\p'}$ is the projection performed on the barycentric frame at $\p' = \Exp (\p, \v)$ to project the finite differences to the same coordinate system. Note that the perturbation along $\versor{\parallel}$ is aligned with the direction of $\v$ and is equivalent to increasing the length of the trace by $\varepsilon_\v$. Therefore, instead of computing the full $\Exp(\p, \v + \varepsilon_\v \versor{\parallel})$, we can trace a significantly shorter geodesic from $\p'$ with initial vector equal to the parallel transported incremental displacement. In other words $\Exp(\p, \v \! + \! \varepsilon_\v \versor{\parallel}) = \Exp \big(\p', \prod_{\p}^{\p'}(\varepsilon_\v \versor{\parallel}) \big)$.

        \begin{wrapfigure}[7]{L}[0pt]{0.5\linewidth}
            \centering
            \vspace{-3pt}
            \includegraphics[width=\linewidth]{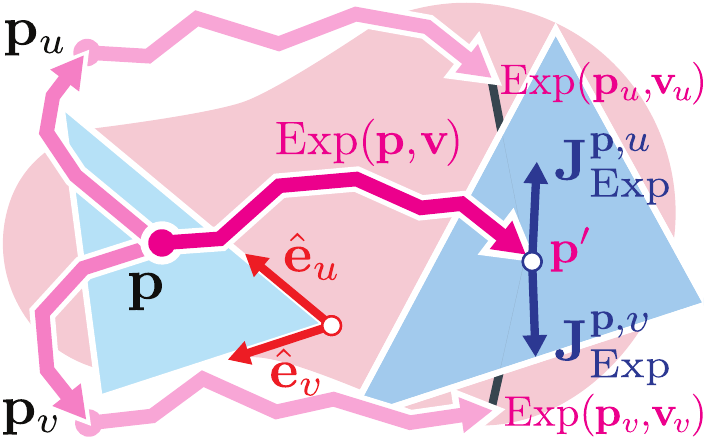}
        \end{wrapfigure}
        The procedure to obtain $\J_\Exp^\p$ is very similar, but a perturbation of $\p$ in its barycentric system may move the origin of the geodesics onto new faces. Therefore, we also need to compute the new starting points and their corresponding parallel-transported starting vectors. We first perturb $\p$ and compute $\p_u = \Exp ( \p, \varepsilon_\p \versor{u})$ and $\p_v = \Exp ( \p, \varepsilon_\p \versor{v})$, as well as their parallel transported directions $\v_u = \prod_{\p}^{\p_u}(\v)$ and $\v_v = \prod_{\p}^{\p_v}(\v)$. Then, we have:
        \begin{equation}
            \label{eq:jp}
            \J_\Exp^\p \approx \mat{M}^\dagger_{\p'} \cdot
                \Big[ 
                    \frac{\scriptstyle \Exp(\p_u, \v_u) - \p'}{\varepsilon_\p}
                    \quad
                    \frac{\scriptstyle \Exp(\p_v, \v_v) - \p'}{\varepsilon_\p} 
                \Big].
        \end{equation}

    \input{algorithms/straightest_geodesics_GPU_short}

    \paragraph{Parallelisation on the GPU}
        To efficiently compute $\Exp(\p, \v)$ for large batches, we parallelise the tracing algorithm on the GPU. The core design principle is to leverage the independence of each geodesic path. We launch a single CUDA kernel where each thread is responsible for tracing one complete geodesic path from a starting point $\p$ with an initial vector $\v$ (\cref{alg:gpu_parallel}). Each thread executes its own tracing loop, iteratively advancing the point across the mesh faces as described in \cref{sec:dpsg}. This one-thread-per-geodesic strategy naturally accommodates paths of varying complexity and length. Threads assigned to shorter geodesics (i.e., smaller $\|\v\|$) or simpler paths will complete their while loop and write their result, becoming idle sooner. Meanwhile, threads tracing longer or more complex paths (e.g., crossing many vertices) continue executing. This model avoids complex inter-thread synchronisation. The GPU's streaming multiprocessors remain saturated with active warps tracing the remaining paths, ensuring high utilisation until the longest path in the batch is complete. 
        
        Since the tracing algorithm only depends on the local faces, it can also simultaneously process meshes heterogeneously batched together. Ensuring that the indices within face matrices ($\Faces$) and intrinsically defined points $\mathcal{P} = \{ (\faceindex, \bary)\}$ are correctly off-setted for each $\mesh$ is the only additional step required for parallelising across meshes.

%% file: algorithms/straightest_geodesics_GPU_short.tex
\begin{algorithm}[t]
    \caption{Parallel Geodesic Tracing CUDA Kernel}
    \label{alg:gpu_parallel}
    \begin{algorithmic}[0]
        \Require $\mesh$ (mesh); 
            $\mathcal{P},\V$ (batch of start points and vectors)
        \Ensure $\mathcal{P}',\V'$ (batch of end points and vectors)            
        \State  $i \gets \text{threadIdx.x} + \text{blockIdx.x} \times \text{blockDim.x}$
        \State  \textbf{if} $i \ge \text{length}(\mathcal{P})$ \textbf{then return}
        \State $\p \gets \mathcal{P}[i]$; $\v \gets \V[i]$
        \State  $(\v, \p) \gets \mathtt{initialise\_trace}(\mesh, \p, \v)$
        \State $L_{\text{target}} \gets \|\v\|$; $L_{\text{traced}} \gets 0$; $step \gets 0$
        \State  \textbf{while} $L_{\text{traced}} < L_{\text{target}}$ \textbf{and} $step < \text{MAX\_STEPS}$:
        \State  \qquad $(\p, \v, L_{\text{step}}) \gets \mathtt{geodesic\_step}(\mesh, \p, \v )$
        \State \qquad $L_{\text{traced}} \gets L_{\text{traced}} + L_{\text{step}}$; $step \gets step + 1$
        \State $\mathcal{P}'[i] \gets \p; \V'[i] \gets \v$
    \end{algorithmic}
\end{algorithm}

%% file: sec/5_experiments.tex
\section{Experimental Evaluation \& Applications}
    \label{sec:experiments}
    We report the accuracy and performance of our method in both its forward and backward steps in \cref{sec:acc-perf}. Then, we present our three novel applications in \cref{sec:geoconv,sec:flow,sec:opt2ord}.

    \begin{figure}[t]
        \vspace{-8pt}
        \centering
        \includegraphics[width=\linewidth]{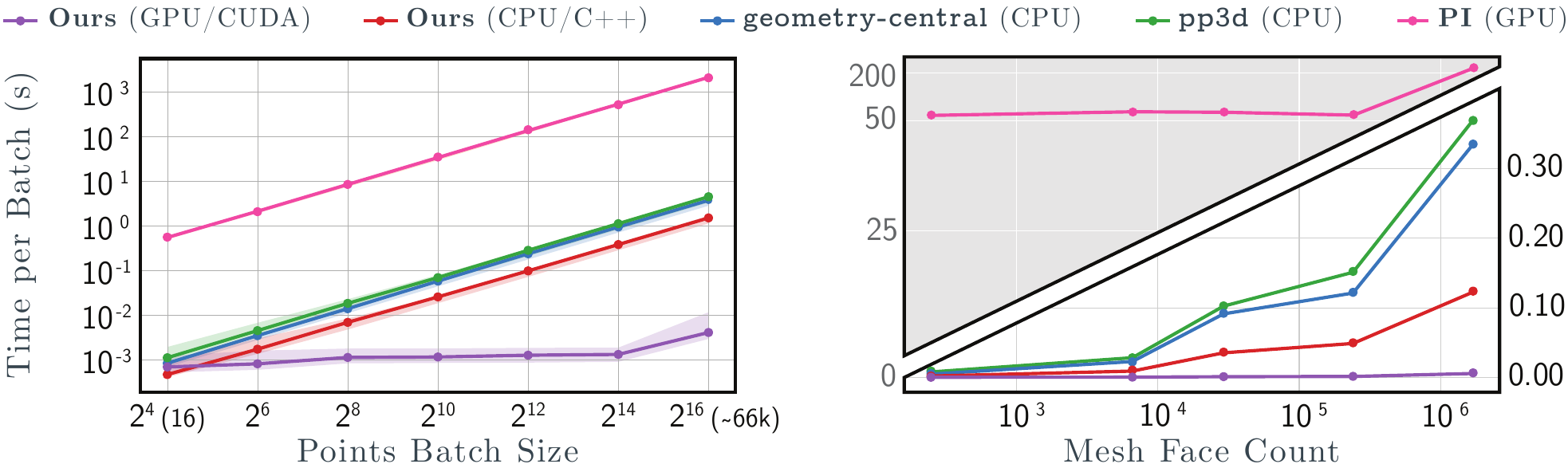}
        \vspace{-20pt}
        \caption{Median runtime comparisons with different batch sizes (\textit{left}) and face counts (\textit{right}). Medians are computed over 5 consecutive executions (as well as 5 different meshes for the \textit{left}). While variance is negligible with face counts, $25\%$ and $75\%$ intervals are reported for batch sizes. The time axis was split in \textit{right} to highlight the massive performance gap with PI. \vspace{-4mm}}
        \label{fig:runtime}
    \end{figure}

    \StartTightFigureParagraph
    \subsection{Accuracy \& Performance}   
        \label{sec:acc-perf}

        We test our $\Exp$ across the three back-end implementations we provide (Python, C++, and CUDA) against the SotA implementations of \cite{polthier2006straightestgeod} provided by the C++ library $\mathtt{geometry \! - \! central}$~\cite{geometrycentral} and its Python wrapper $\mathtt{potpourri3d}$ ($\mathtt{pp3d}$). We test on 5 different meshes (\cref{fig:meshes-benchmarking}) with a diverse face count in the approximate range of $F \in [200, 1.7\text{M}]$ and report accuracies comparable to $\mathtt{pp3d}$ (\cref{tab:accuracy-combined}, \textit{left}), while being $3$ \emph{orders of magnitude} faster on big batch sizes (\cref{fig:runtime},~\textit{left}). Our CUDA implementation also \emph{scales linearly} with the number of faces of a mesh, consistently outperforming by $2$ \emph{orders of magnitudes} other methods (\cref{fig:runtime}~\textit{right}). Our CUDA implementation also has nearly constant run-times while scaling either of those factors on a single NVIDIA RTX-4090.       
        
        As reported in \cref{tab:accuracy-combined}~(\textit{right}), both ours and $\mathtt{pp3d}$'s $\Exp$ report similar performance on the closed-form exponential map of the sphere $\Exp^\text{sphere}(\p, \v) = \p \cos(\|\v\|) + \frac{\v}{\|\v\|} \sin(\|\v\|)$ and the exponential map computed by solving the IVP on a torus (properly formulated in \cref{sec-supp:ivptorus}). Errors are here mostly caused by discretisation and they decrease as resolution increases and the mesh becomes a better approximation of the underlying continuous surface.
        
        \begin{wrapfigure}[9]{L}[0pt]{0.5\linewidth}
            \centering
            \vspace{-8pt}
            \includegraphics[width=\linewidth]{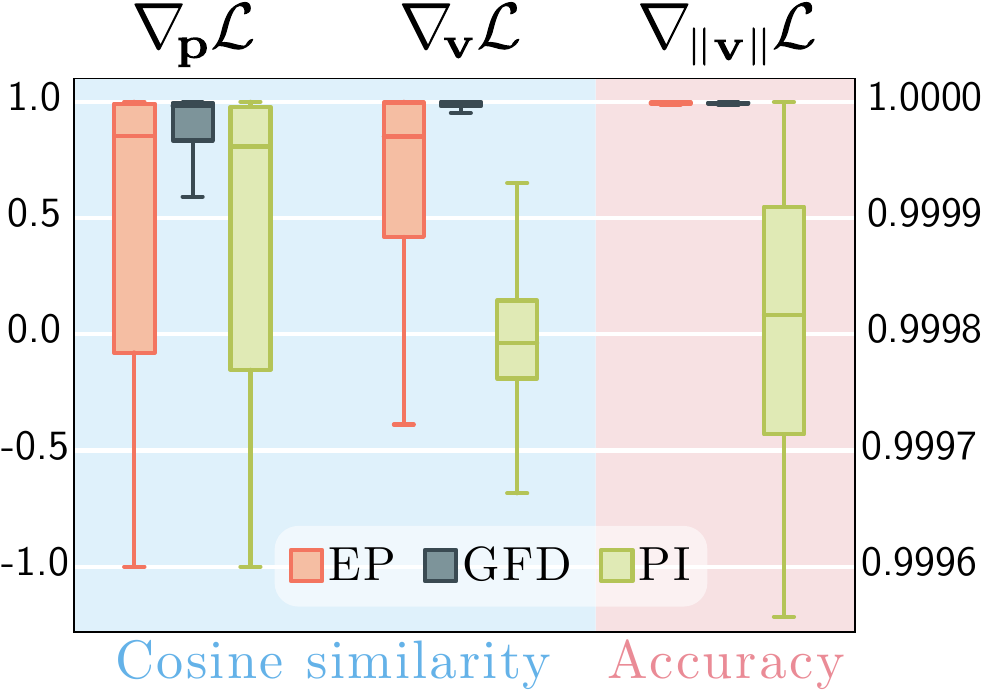}
            \vspace{-20pt}
            \caption{Differentiation correctness against closed-form.}
            \label{fig:diff-acc}
        \end{wrapfigure}
        
        The accuracy of our two differentiation schemes is tested against closed-form gradients computed on the sphere (\cref{fig:diff-acc}), which we derive in \cref{sec-supp:grad_sph}. 
        Our GFD scheme closely matches the gradients on the sphere for both $\v$ and $\p$. Results for our EP schemes corroborate our mathematical derivation by showing that gradients for $\v$ are a good approximation of the real ones, with their norm being particularly reliable. In addition, it also shows that the gradients for $\p$ are not reliable, further motivating our choice for voiding their contributions. Our experiments show that the backward pass with GFD is about 3.85 times more expensive than with EP. Overall while GFD should be the default choice for most applications, EP is preferable for applications requiring fast computations of gradients for $\v$ and where $\p$ are fixed (e.g., \cref{sec:geoconv}). %

        The projection integration (PI) we reproduced from~\cite{Madan2025local-parameterizations} exhibits similar accuracy to ours (\cref{tab:accuracy-combined},~\textit{right}), but we are a staggering $5$ to $6$ \emph{orders of magnitude} faster on meshes with a high face count and large batches, respectively (\cref{fig:runtime}).
        Note that PI can be auto-differentiated, but gradients (and speed) are of insufficient quality for optimisation (\cref{fig:diff-acc}). %

        \EndTightFigureParagraph
        
        \begin{table}
            \centering
            \caption{Comparison of Accuracy Metrics. \textit{Left}: mean distances between the endpoints of our different backends and $\mathtt{pp3d}$. \textit{Right}: mean distances against the $\text{Exp}$ map of the sphere and torus. 
            }
            \vspace{-6pt}
            \resizebox{\linewidth}{!}{
            \begin{minipage}[t]{0.45\textwidth}
                \huge
                \begin{tabular}{c|ccc}
                    & python & C++ & CUDA \\
                    \hline
                    float32 & $2.0\text{e-5}$ & $2.0\text{e-5}$ & $2.0\text{e-5}$  \\
                    float64 & $7.1\text{e-9}$ & $1.0\text{e-9}$ & $1.0\text{e-9}$  \\
                \end{tabular}
            \end{minipage}%
            \hspace{60pt}
            \begin{minipage}[t]{0.55\textwidth} %
                \raggedright
                \huge
                \centering
                \begin{tabular}{c|ccc}
                    & Ours & $\mathtt{pp3d}$ & PI~\cite{Madan2025local-parameterizations} \\
                    \hline
                    Sphere & $5.7\text{e-4}$ & $5.7\text{e-4}$ & $5.4\text{e-4}$ \\
                    Torus & $8.9\text{e-3}$ & $8.9\text{e-3}$  &$ 8.9\text{e-3}$ 
                \end{tabular}
            \end{minipage}
            }       
            \vspace{-4mm}
            \label{tab:accuracy-combined}
        \end{table}

    \subsection{Adaptive Geodesic Convolutions (AGC)}
        \label{sec:geoconv}
    
        \paragraph{Problem \& Setup}
            Geodesic Convolutional Neural Networks (GCNN) \cite{gcnn} extend the convolution operation to manifolds by replacing the Euclidean metric with geodesic distances defined on the surface. %
            Most follow-up works~\cite{gcnn, pc_gcnn, mdgcnn, hsn_gcnn, fc_gcnn} tried to overcome the absence of a local coordinate system. Nevertheless, a key limitation remains:
            their reliance on a fixed patch size, typically determined by precomputing geodesic distances or aggregating a predetermined set of neighbouring vertices. This rigid definition limits the model’s ability to adapt to local geometry. 

        \StartTightFigureParagraph
        \paragraph{Approach}
            \begin{wrapfigure}[6]{O}[0pt]{0.6\linewidth}
                \centering
                \vspace{-5pt}
                \includegraphics[width=\linewidth]{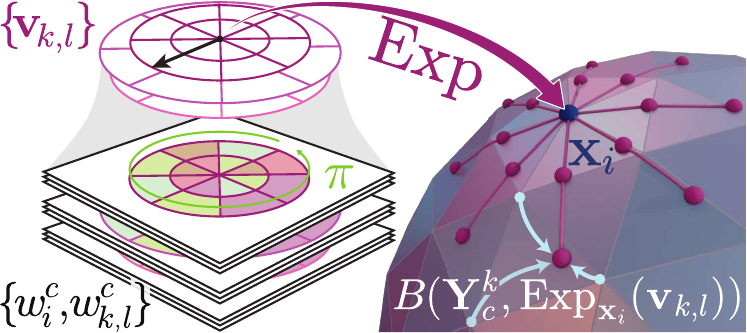}
                \label{fig:agc}
            \end{wrapfigure}
            To address this, we propose leveraging our differentiable straightest geodesics to dynamically learn patch sizes during training. This allows the receptive field to adjust based on the underlying surface geometry, resulting in a more flexible and expressive model capable of capturing features at the most relevant scales.  Note that unlike DiffusionNet~\cite{diffusionnet}, which also learns kernel's sizes for different channels, our filters are not curvature dependent ---a side effect of using the Laplace Beltrami Operator to diffuse heat.
        
            We define a set of planar vectors by rotating an arbitrary base versor ($\versor[v]{base}$) $n_\theta$ times while modulating its length with $n_\rho$ different values: $\v_{k,l} =  k\rho_w \mat{R}(\theta_l) \versor[v]{base}$ for $k=1...n_\rho, \; l=1...n_\theta$. Values defined at the end-points of these vectors and their origin determine the learnable filter values $w^c_{l,k}$ and $w^c_i$, respectively. To apply filters, we first find the corresponding points on the surface $\p'_{k,l} = \Exp_{\x_i}(\v_{k,l}) \in \mesh$ and then barycentrically interpolate the values at the vertices of the faces on which the $\p'_{k,l}$ belong. If we represent the interpolation of vertex feature $\Y^\kappa_c$ at layer $\kappa$ and channel $c$ with $B(\Y^\kappa_c, \Exp_{\x_i}\!(\v_{k,l}))$, we have:
            \begin{equation*}
                \Y^{\kappa+1}_{i, c} \! \! = \! \Y^{\kappa}_{i, c} w^c_i + \max_\pi \! \sum_{l=1}^{n_\theta} \sum_{k=1}^{n_\rho} \! w^c_{\pi(l),k} B(\Y^\kappa_c, \Exp_{\x_i}\!(\v_{k,l})).
            \end{equation*}
            Rotational invariance is achieved by cyclically permuting angular indices with $\pi(l)$, effectively rotating the filters. To speed up computations, filters are grouped in $\nu$ subsets sharing the same $\{\v_{k,l}\}$. Thanks to our differentiable $\Exp$, along the filter values we also learn one $\rho_w$ for each subset of filters. Since we are optimising only the norm of the vector directions, we use the EP differentiation scheme, which is much faster to compute and quite accurate when optimising with respect to the norm of the direction vector.

            Our network takes as input a vector of real values for each vertex. Like in DiffusionNet~\cite{diffusionnet}, we considered two possible input types: the raw 3D coordinates of the vertices, and the Heat Kernel Signature (HKS)~\cite{hks}. 
            More details on our model and training setup are provided in~\cref{sec-supp:geonet}.

        \paragraph{Results} 
            We trained our AGCNN to segment the human body parts on the composite dataset~\cite{human-seg-dataset}, containing models from several other human shape datasets~\cite{dataset-adobe, dataset-faust, dataset-scape, dataset-shrec, dataset-mit}. We trained our model on the original dataset, predicting faces, by summing the logits of the face vertices. When using HKS as input, the model converges faster and reaches a higher accuracy. {\parfillskip0pt\par}
            
            \begin{wrapfigure}[7]{L}[0pt]{0.4\linewidth}
                \centering
                \vspace{-7pt}
                \includegraphics[width=\linewidth]{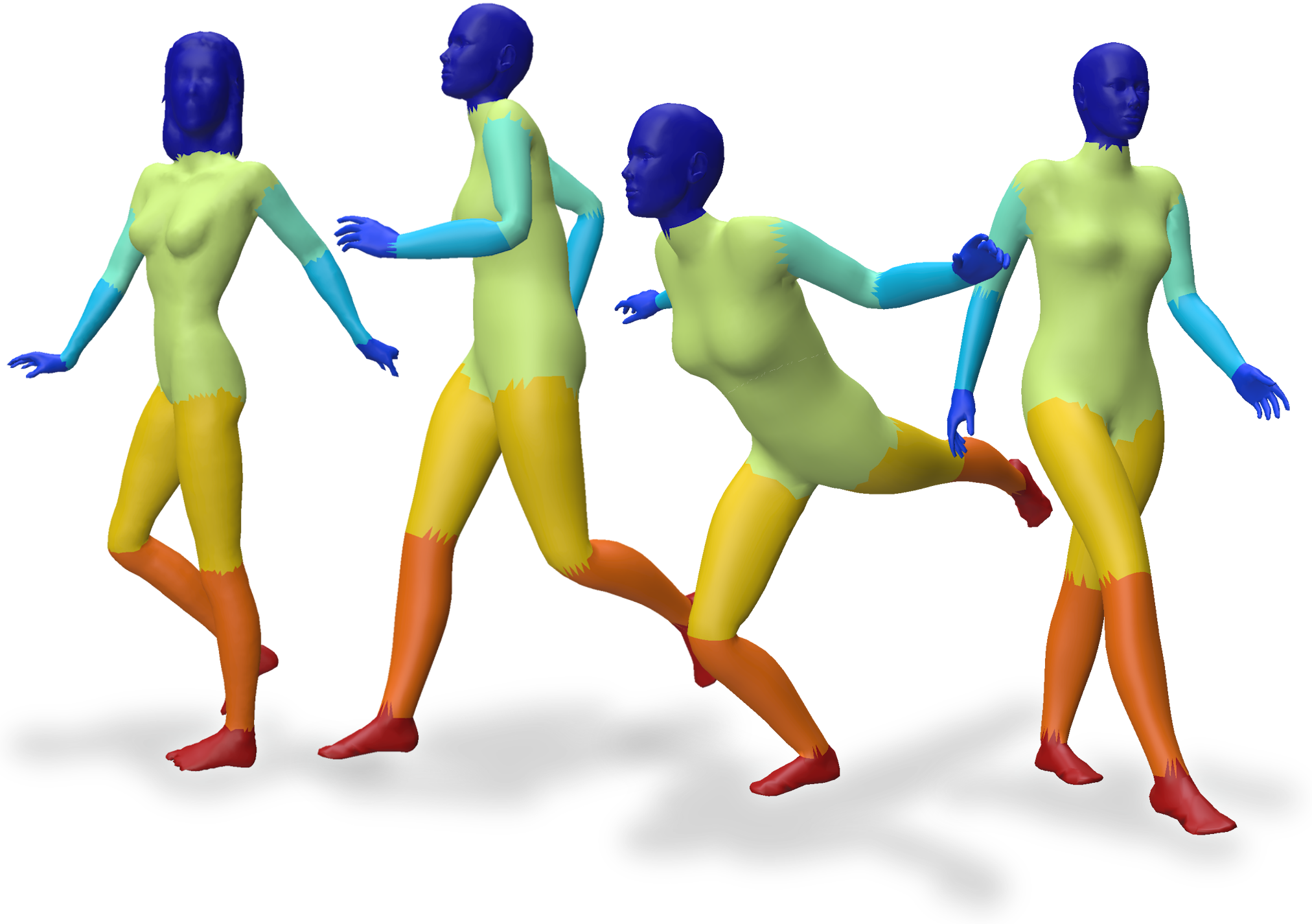}
                \vspace{-27pt}
                \caption{AGC (ours) on body parts segmentation.}
                \label{fig:body-res}
            \end{wrapfigure}
            \noindent
            \cref{fig:body-res} depicts some segmentations performed with our model, while \cref{tab:human_seg} reports a comparison against SotA methods using geodesic convolutions on meshes. %
            Even though our model is just an extension of GCNN~\cite{gcnn},
            it brings significant improvements over other methods and particularly over our closest competitors (i.e., GCNN~\cite{gcnn} and MDGCNN~\cite{mdgcnn}). The only model outperforming AGC was trained to predict vertex rather than face classes, making the comparison only partially fair.
            We trained our model on a Nvidia A100, taking $1$ minute and $21$ seconds per epoch and using around $6.8$GB of VRAM, while training on the full size meshes made of around $10$k vertices. The GFD scheme, took around $2$ minutes and $8$ seconds per epoch and $7.2$GB of VRAM, for similar accuracy.
        
            Our convolution layer is currently inspired by GCNN~\cite{gcnn}, which is one of the earliest and simplest mesh convolutions. As future study, we imagine adapting MDGCNN~\cite{mdgcnn} or FC~\cite{fc_gcnn} to use adaptive patch sizes like in AGC. Interestingly, by dropping the max pooling like in MDGCNN~\cite{mdgcnn} and learning a different $\rho_k$ for each angle, it would also be possible to learn anisotropic filters.
            \EndTightFigureParagraph
            
            \begin{table}
                \centering
                \footnotesize
                \caption{\label{tab:human_seg} 
                Body part segmentation accuracy. \textit{XYZ} and \textit{HKS} denote position coordinates and heath kernel signatures as inputs. $\mathtt{*}$ indicates models trained on a modified dataset for vertex predictions.
                \vspace{-4pt}}
                \resizebox{.6\linewidth}{!}{
                \begin{tabular}{lc}
                    \hline
                    \textbf{Method} & \textbf{Accuracy}\\
                    \hline 
                    GCNN \cite{gcnn} & 86.4\%\\      
                    MDGCNN \cite{mdgcnn} & 89.5\% \\
                    HSN* \cite{hsn_gcnn} & 91.1\% \\
                    FC* \cite{fc_gcnn} & \textbf{92.9\%} \\
                    DiffusionNet - XYZ \cite{diffusionnet} & 90.6\% \\
                    DiffusionNet - HKS \cite{diffusionnet} & 91.7\% \\
                    \hline
                    AGC (Ours) - XYZ & 91.7\% \\
                    AGC (Ours) - HKS & 92.3\% \\
                \end{tabular}
                }
                \vspace{-5mm}
            \end{table}

    \subsection{MeshFlow}
        \label{sec:flow}
        
        \paragraph{Problem \& Setup}
            Flow Matching~\cite{flowmatching} is a recent approach to constructing normalising flows~\cite{nice, density-nvp, glow, maf, iaf, nf-inference, nf-modeling, freeformcontinuousdynamics} by directly learning a vector field that transports one distribution onto another without requiring likelihood computation through an ODE solver.
            Riemannian flow matching (RFM)~\cite{riemannian-fm} generalises this framework to Riemannian manifolds by learning time-dependent tangent vector fields $v_{t, \theta}(\p) \in \TpM$ and moving points geodesically. Unlike on the simple manifolds, \cite{riemannian-fm} requires solving an ODE to identify a geodesic path on meshes. Since the Euler solver they use can drift off the surface, an additional projection step is required. Thanks to our differentiable intrinsic method we can avoid both steps and directly learn the vector field by back-propagating through $\Exp$. Therefore, our time-independent vector field does not need to match the marginal vector field of RFM. Differences and similarities between the two methods are detailed in \cref{sec-supp:otexpflow}.

        \paragraph{Approach}
            Rather than parametrising and optimising a time-dependent vector field, we instead learn a single stationary vector field defined on the manifold. The generative process is constructed by transporting samples from the base to the target distribution using pointwise correspondences obtained via optimal transport (OT). %
            The transport cost is defined as the squared biharmonic distance \cite{biharmonic}, which serves as a computationally efficient surrogate for the true geodesic distance. %
            We can then use the learned vector field to move the samples to the target distribution. Since we are using exponential maps, there are no projection errors, making the method useable with a low amount of steps, while still maintaining high accuracy. 
            Our objective is:
            \begin{equation}
                \label{eq:fmobj}
                \mathcal{L}(\theta) = \mathbb{E}_{\substack{\p_{i} \sim \mathcal{U}(\mesh)\\ \q_{i} \sim \mathcal{Q}(\mesh)}} \left[ 
                    \sum_{i=1}^B  d^2_\text{BH}  \Big(\Exp^{\circlearrowright K}_{\p_i}(\v_\theta), \q_{\sigma(i)} \Big)
                \right], 
            \end{equation}
            where $\p_{i} \in \mesh$ are initial samples from a uniform distribution over the mesh $\mathcal{U}(\mesh)$, $\q_{i}$ are samples from the target distribution $\mathcal{Q}(\mesh)$, $d_\text{BH}$ is the biharmonic distance, $\sigma$ is a permutation of $[1,B]$ representing the OT coupling for the batch, and $\Exp^{\circlearrowright K}$ is the $K$-steps exponential map breaking the trajectory into multiple steps and enabling non-straightest curves on $\mesh$. In practice, it iteratively computes $\p^{(k+1)}_i = \Exp_{\p^{(k)}_i} \big(\v_\theta (\p^{(k)}_i) / K \big)$ for $k = 0,\dots, K-1$, thus allowing changes in direction by re-evaluating the learned vector field at intermediate points. 
            To learn the vector field, we use a MLP with 3 hidden layers of size $512$, and $K=5$. The MLP takes as input the 3D position of the samples and outputs the corresponding vector. 
            We use the GFD differentiation scheme as we need to back-propagate also through positions.
            Full details in \cref{sec-supp:otexpflow}.
            \begin{figure}
                \centering
                \setlength{\tabcolsep}{4pt} %
                \renewcommand{\arraystretch}{-1.0} %
                \resizebox{\linewidth}{!}{
                    \begin{tabular}{@{}c *{6}{m{0.15\linewidth}}@{}}
                        \raisebox{-0.5\height}{\rotatebox{90}{\scriptsize GT}} &
                        \includegraphics[width=1.1\linewidth]{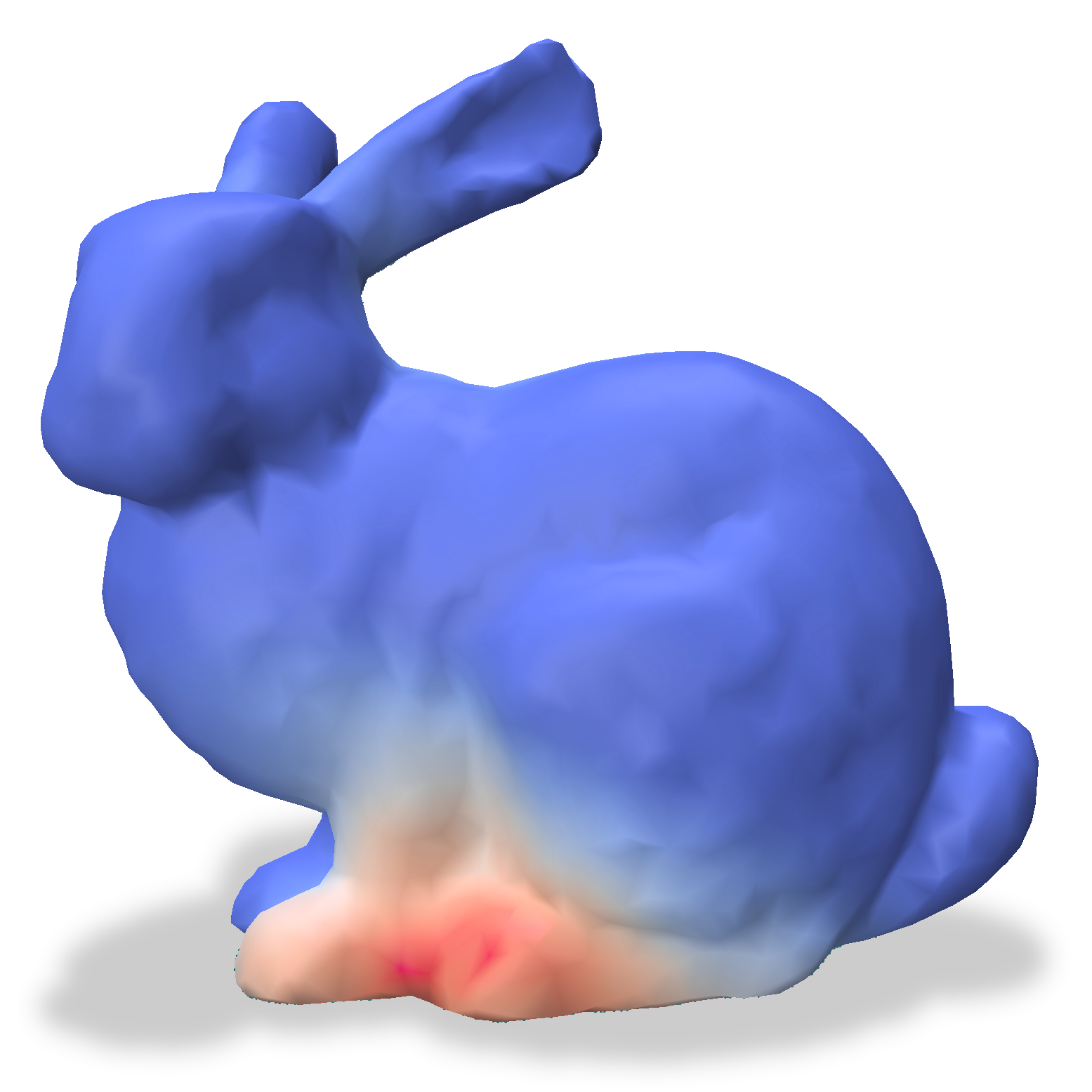} &
                        \includegraphics[width=1.1\linewidth]{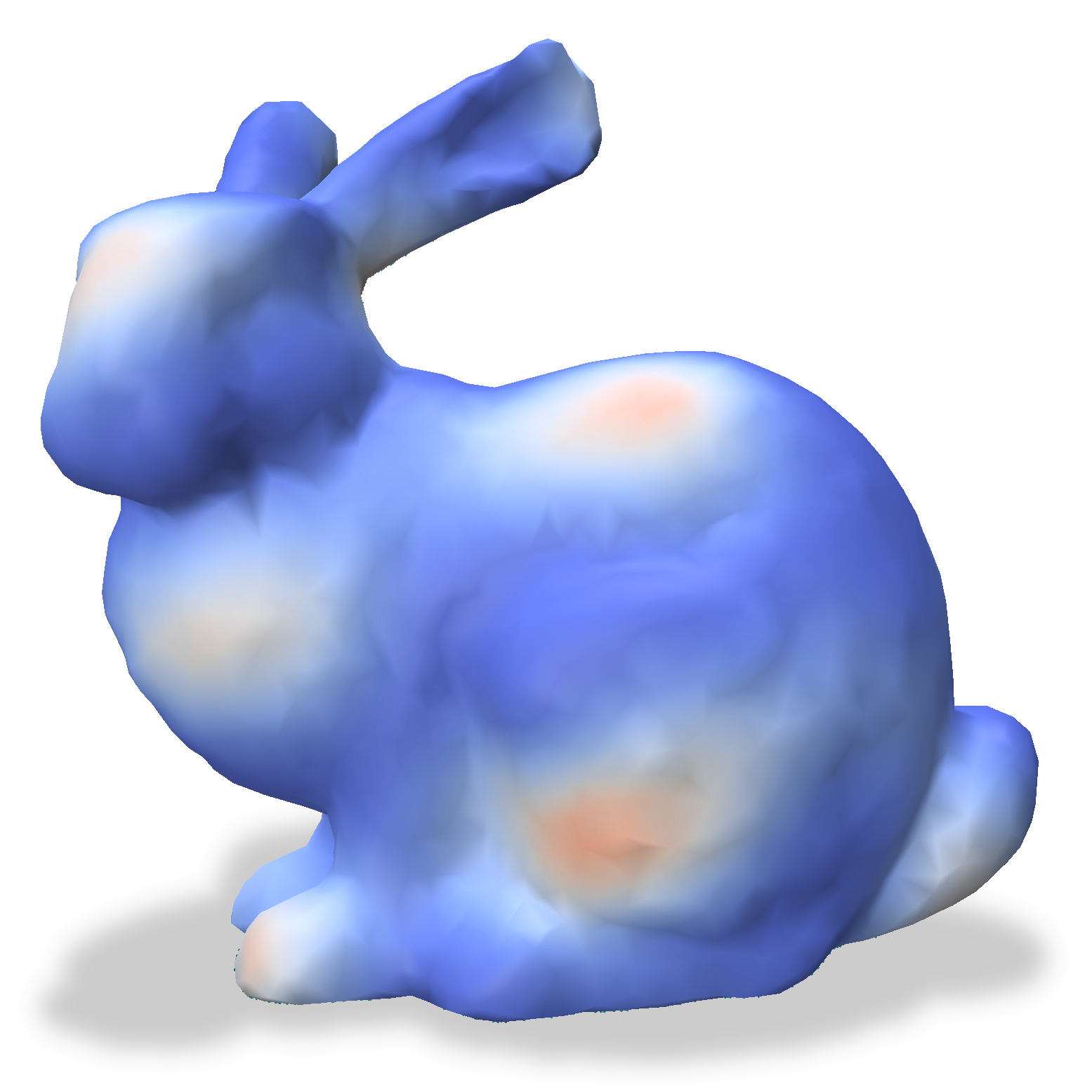} &
                        \includegraphics[width=1.1\linewidth]{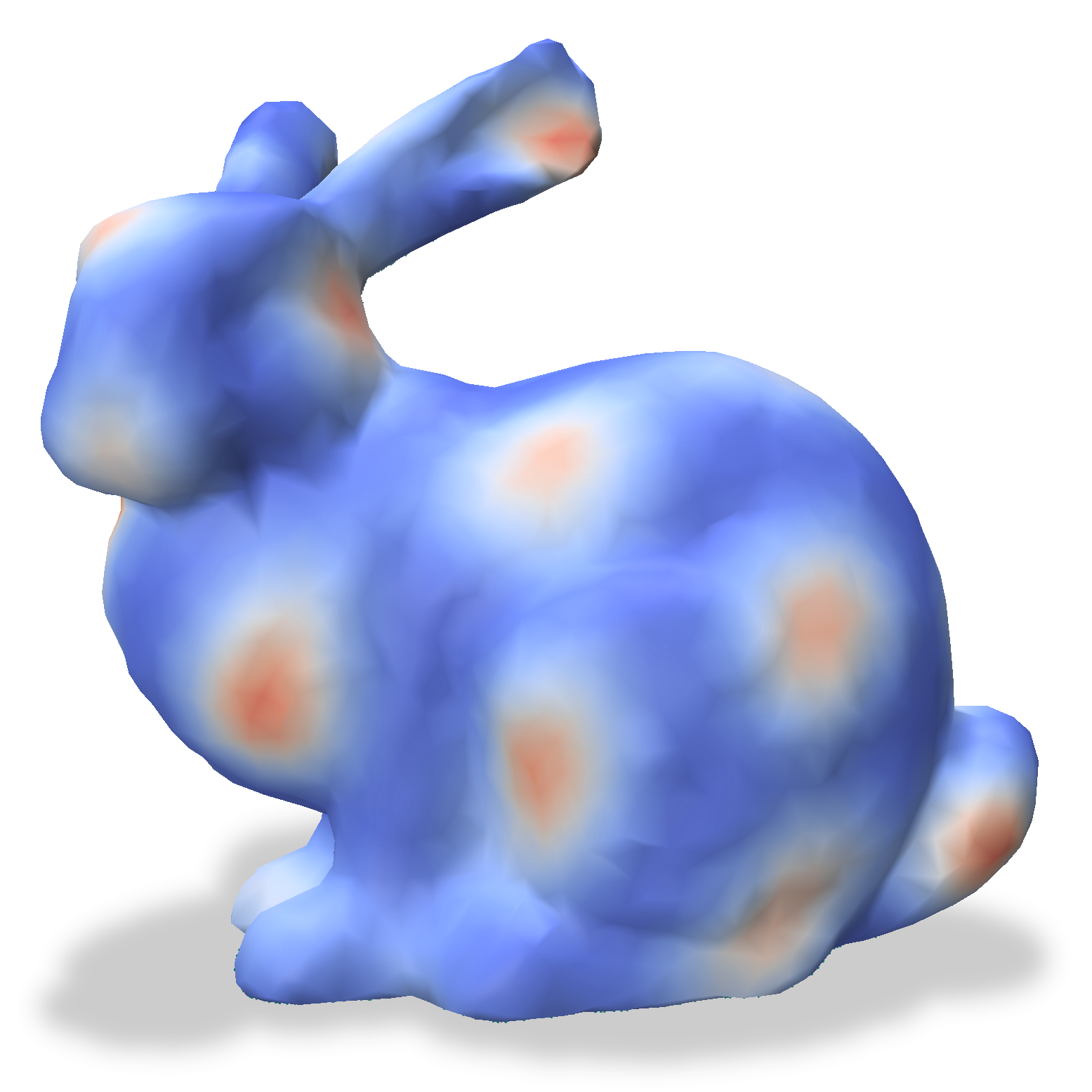} &
                        \includegraphics[width=1.1\linewidth]{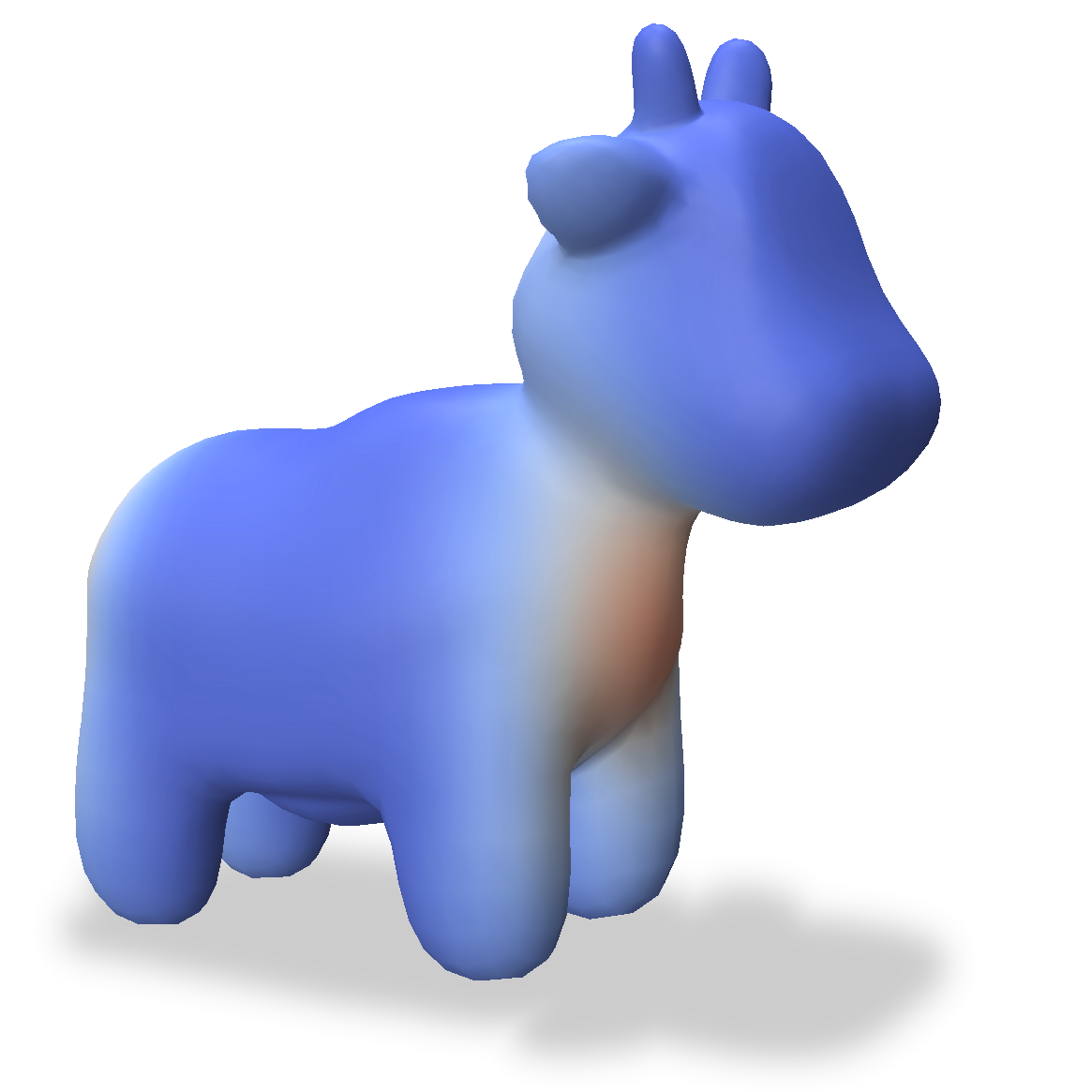} &
                        \includegraphics[width=1.1\linewidth]{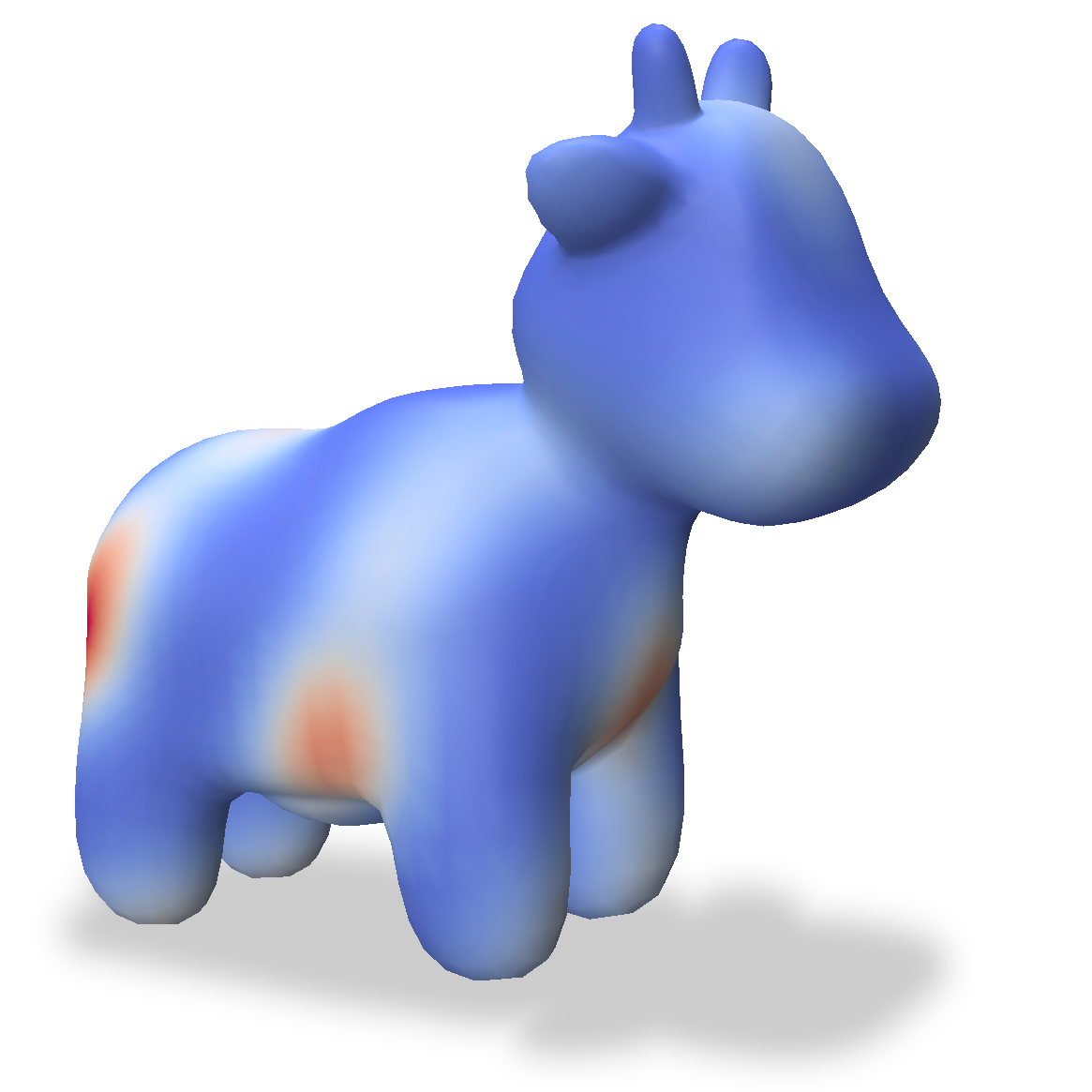} &
                        \includegraphics[width=1.1\linewidth]{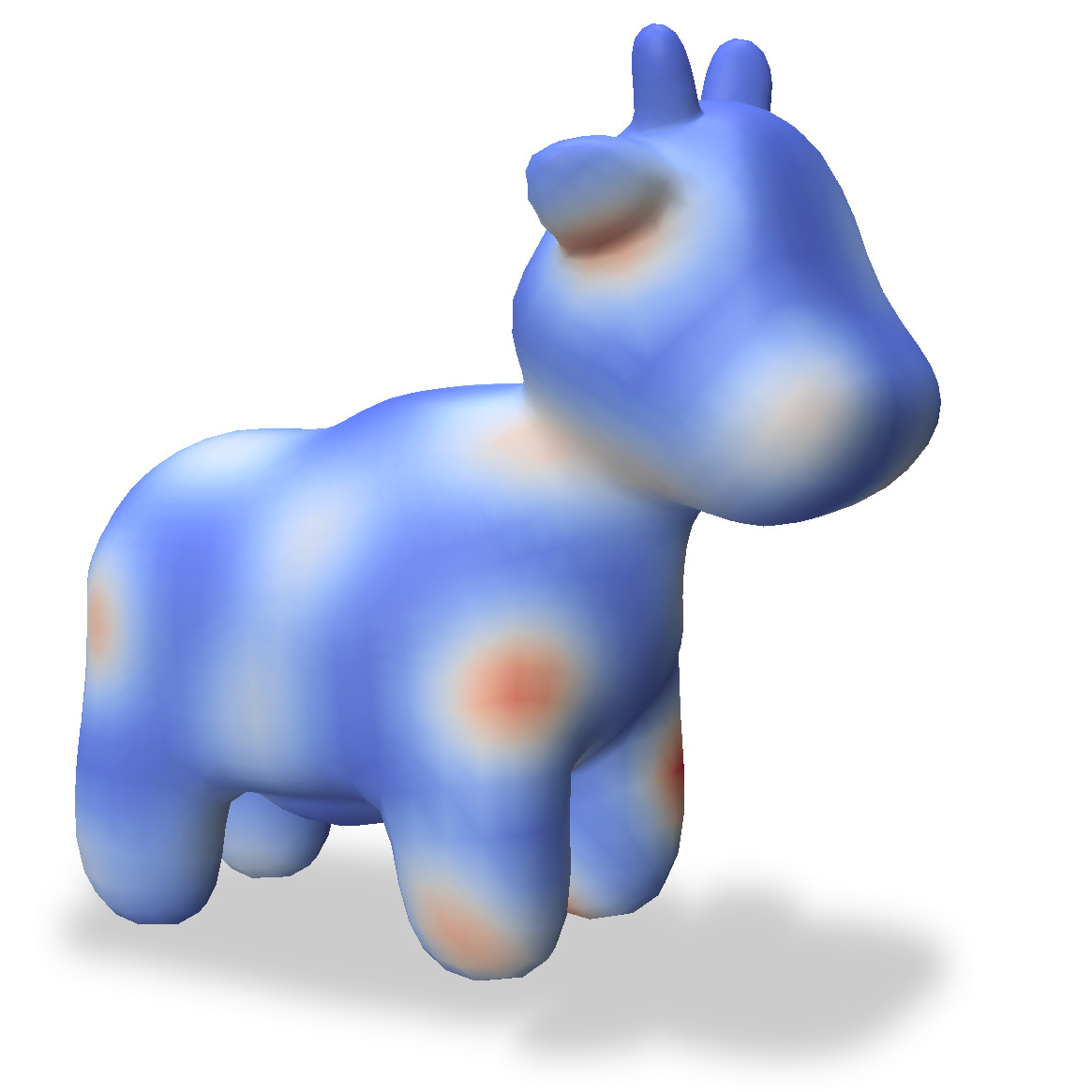} \\
                        
                        \raisebox{-0.5\height}{\rotatebox{90}{\scriptsize RFM}} &
                        \includegraphics[width=1.1\linewidth]{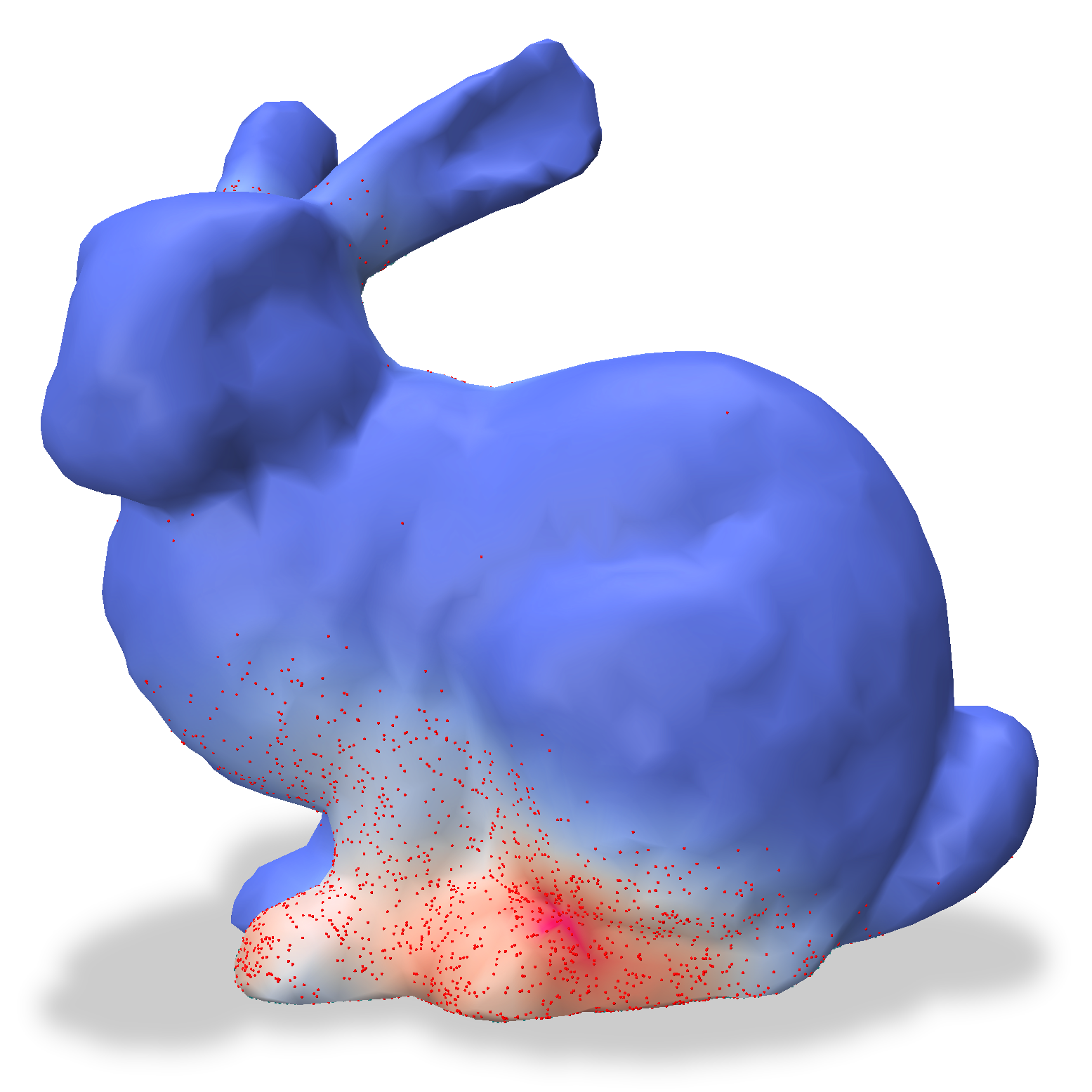} &
                        \includegraphics[width=1.1\linewidth]{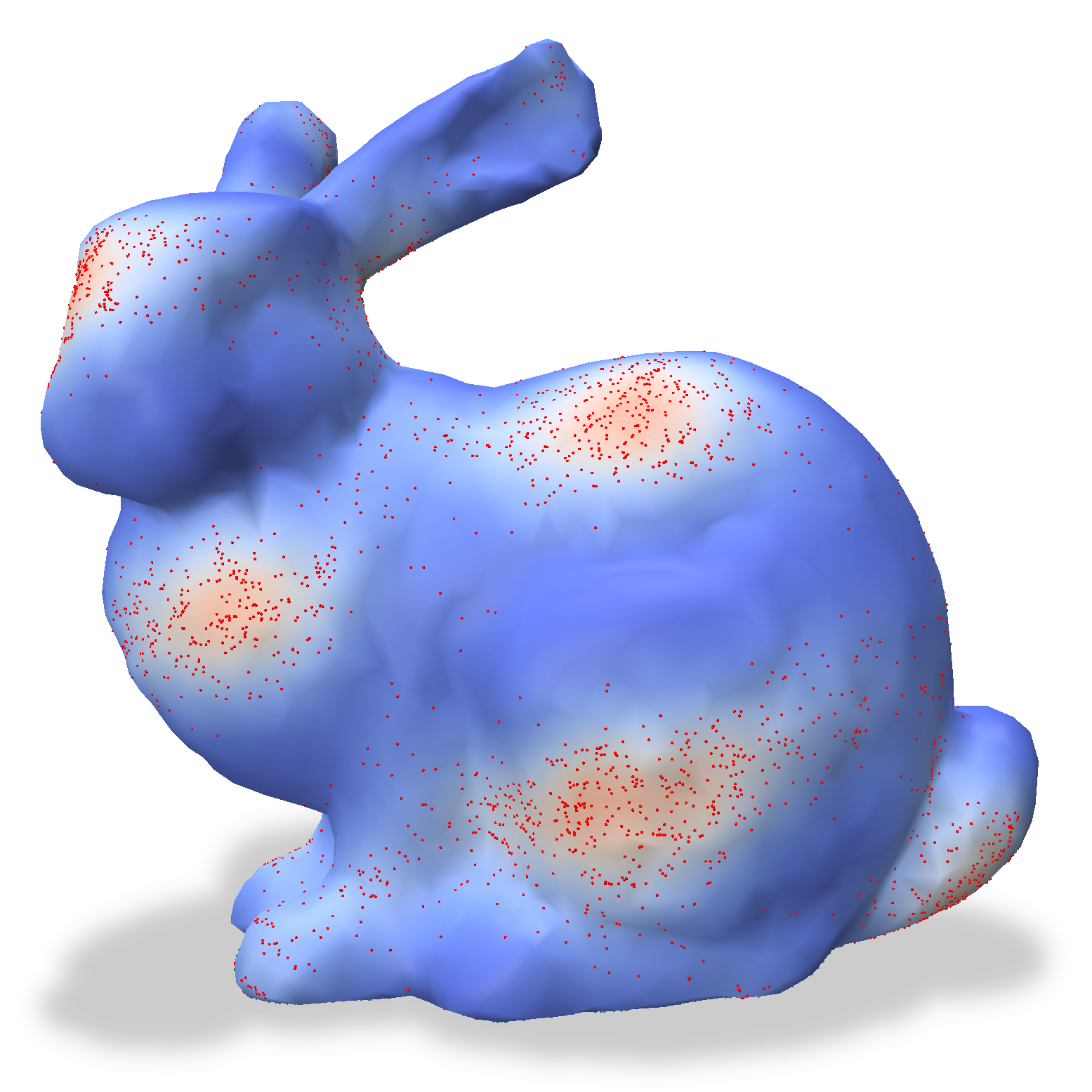} &
                        \includegraphics[width=1.1\linewidth]{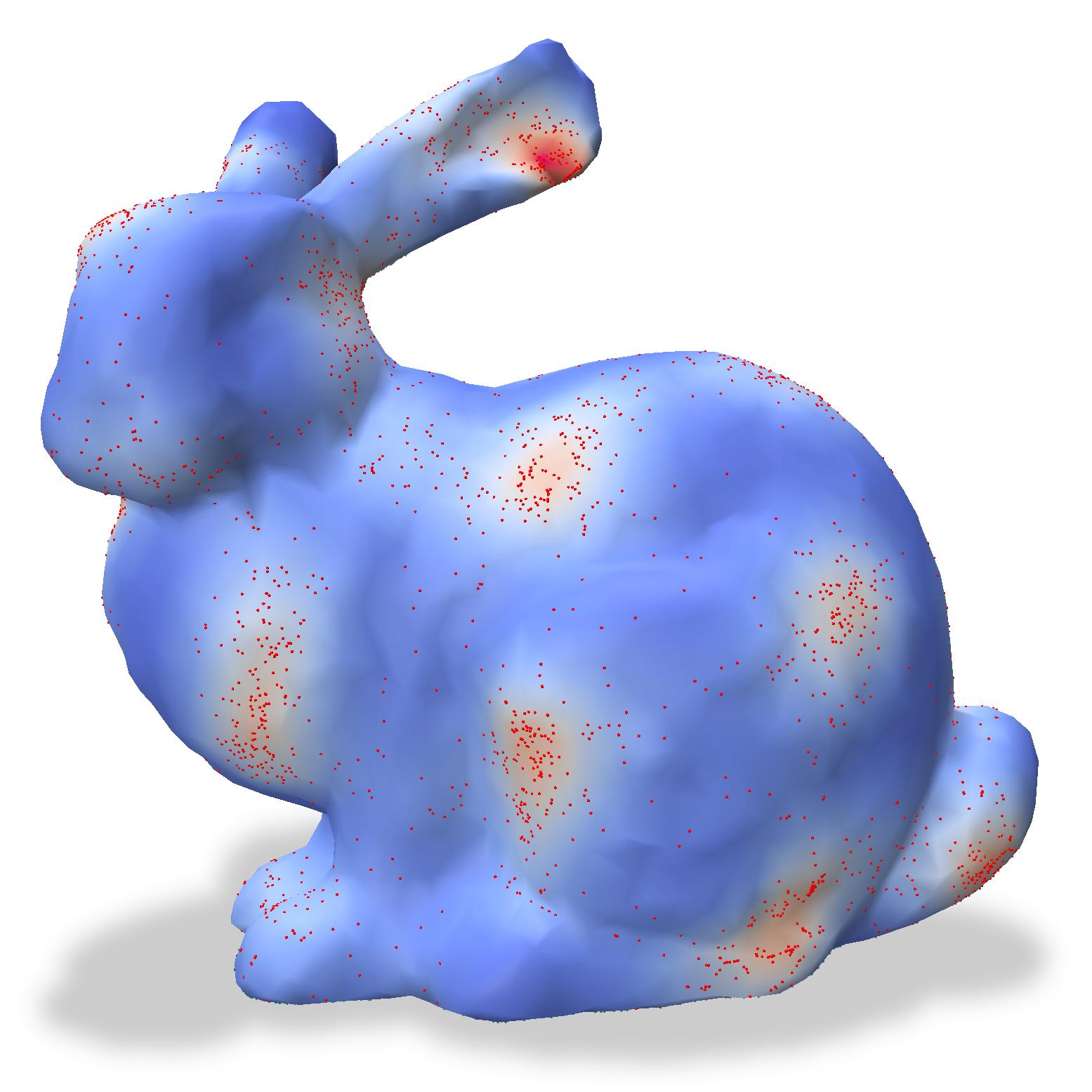} &
                        \includegraphics[width=1.1\linewidth]{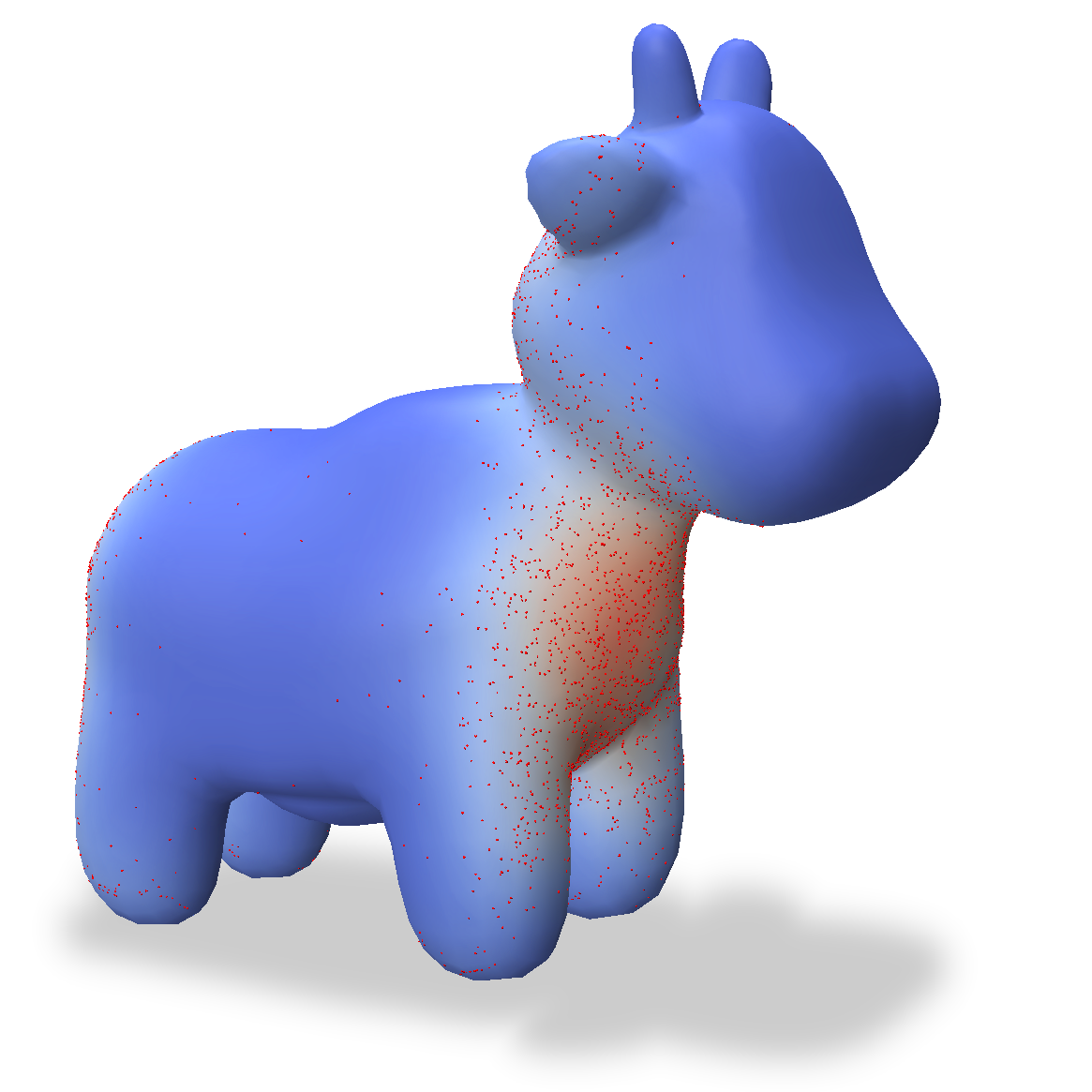} &
                        \includegraphics[width=1.1\linewidth]{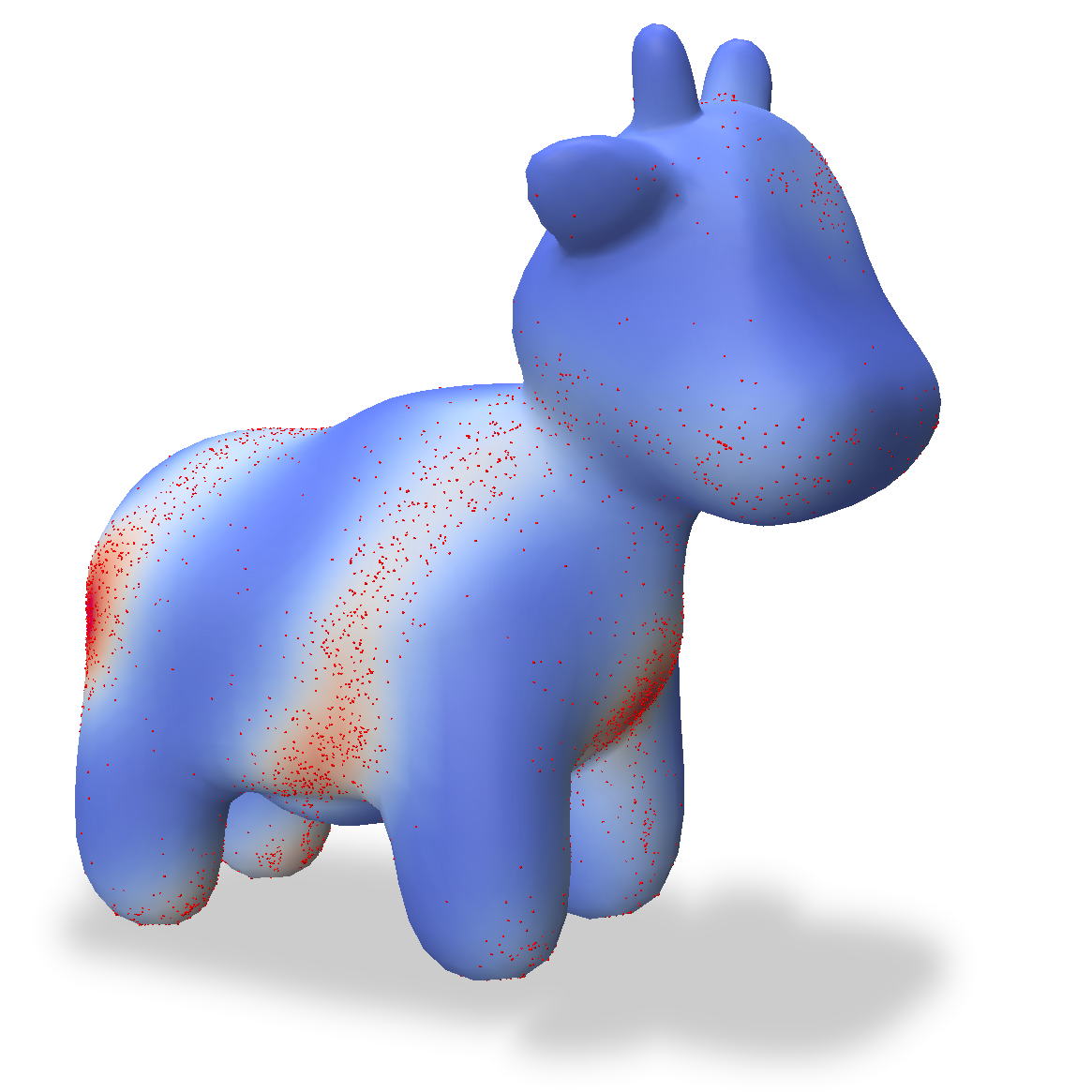} &
                        \includegraphics[width=1.1\linewidth]{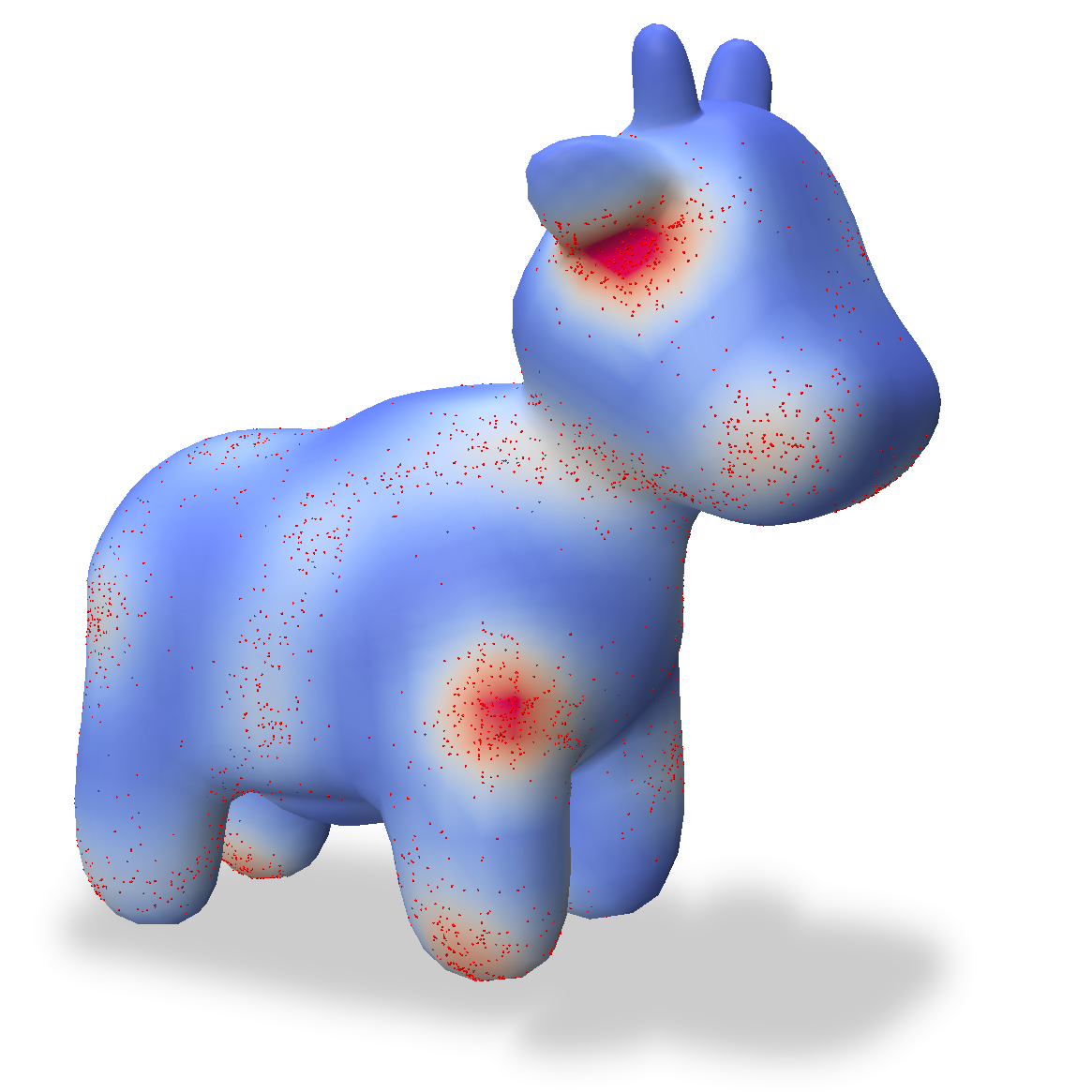} \\
                
                        \raisebox{-0.5\height}{\rotatebox{90}{\scriptsize Ours }} &
                        \includegraphics[width=1.1\linewidth]{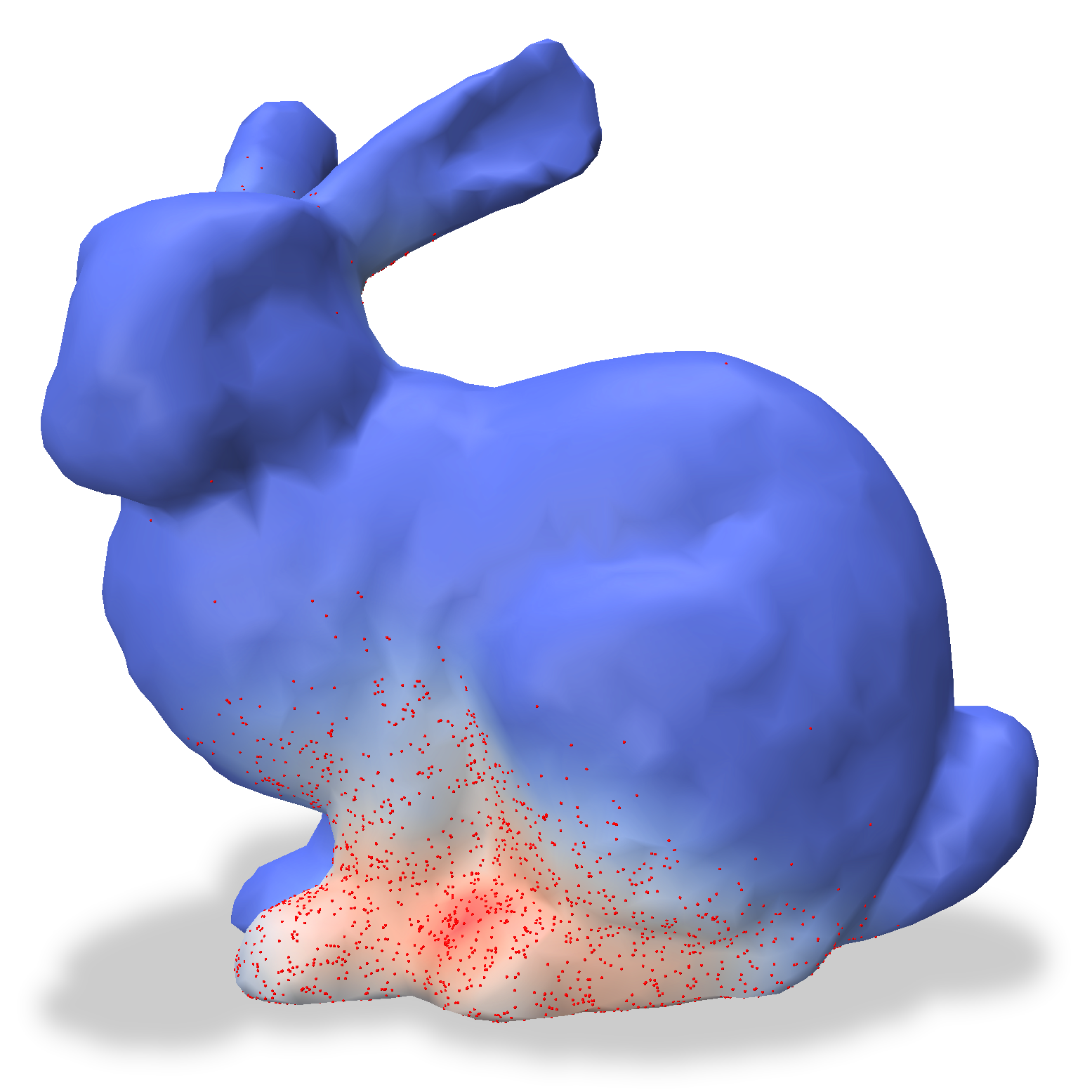} &
                        \includegraphics[width=1.1\linewidth]{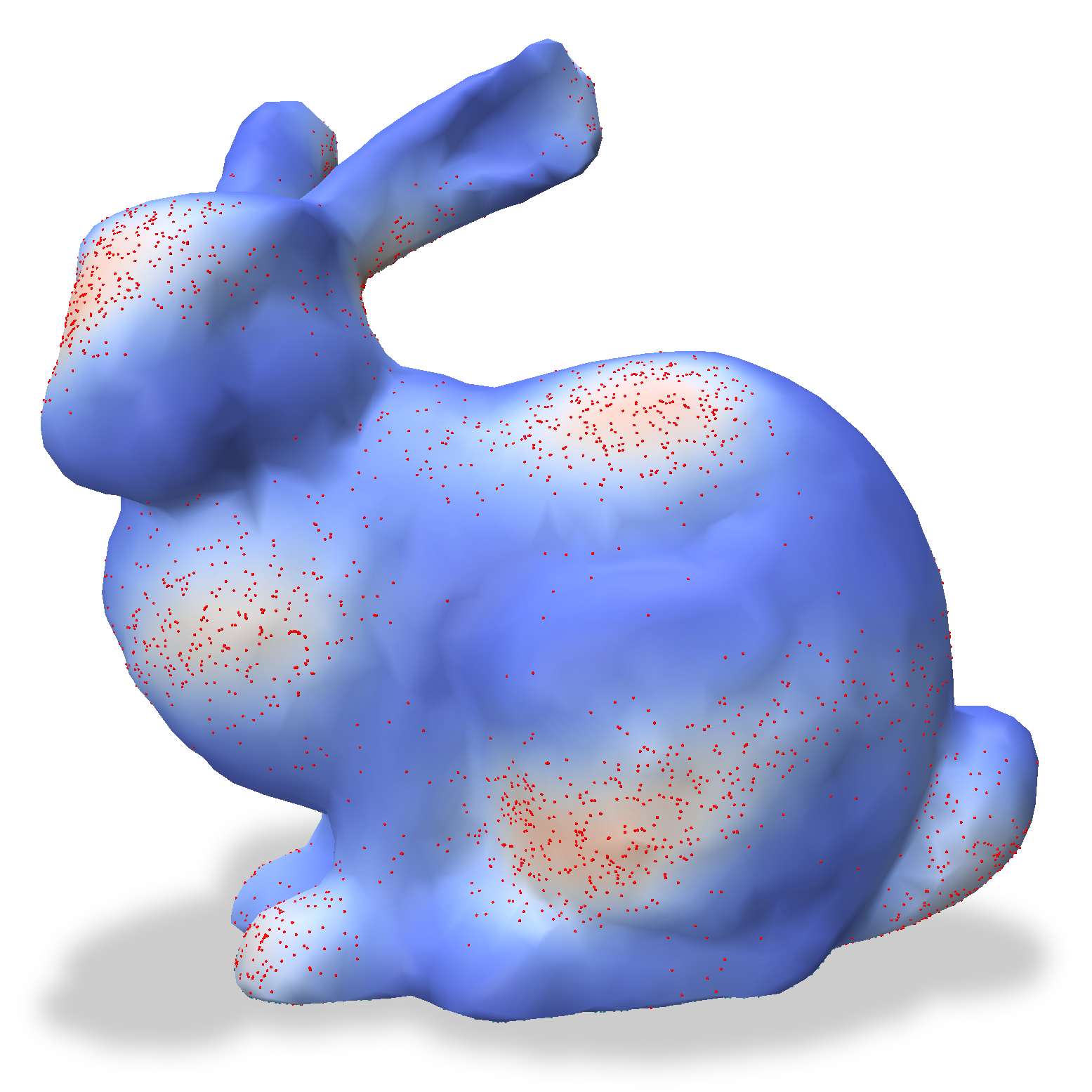} &
                        \includegraphics[width=1.1\linewidth]{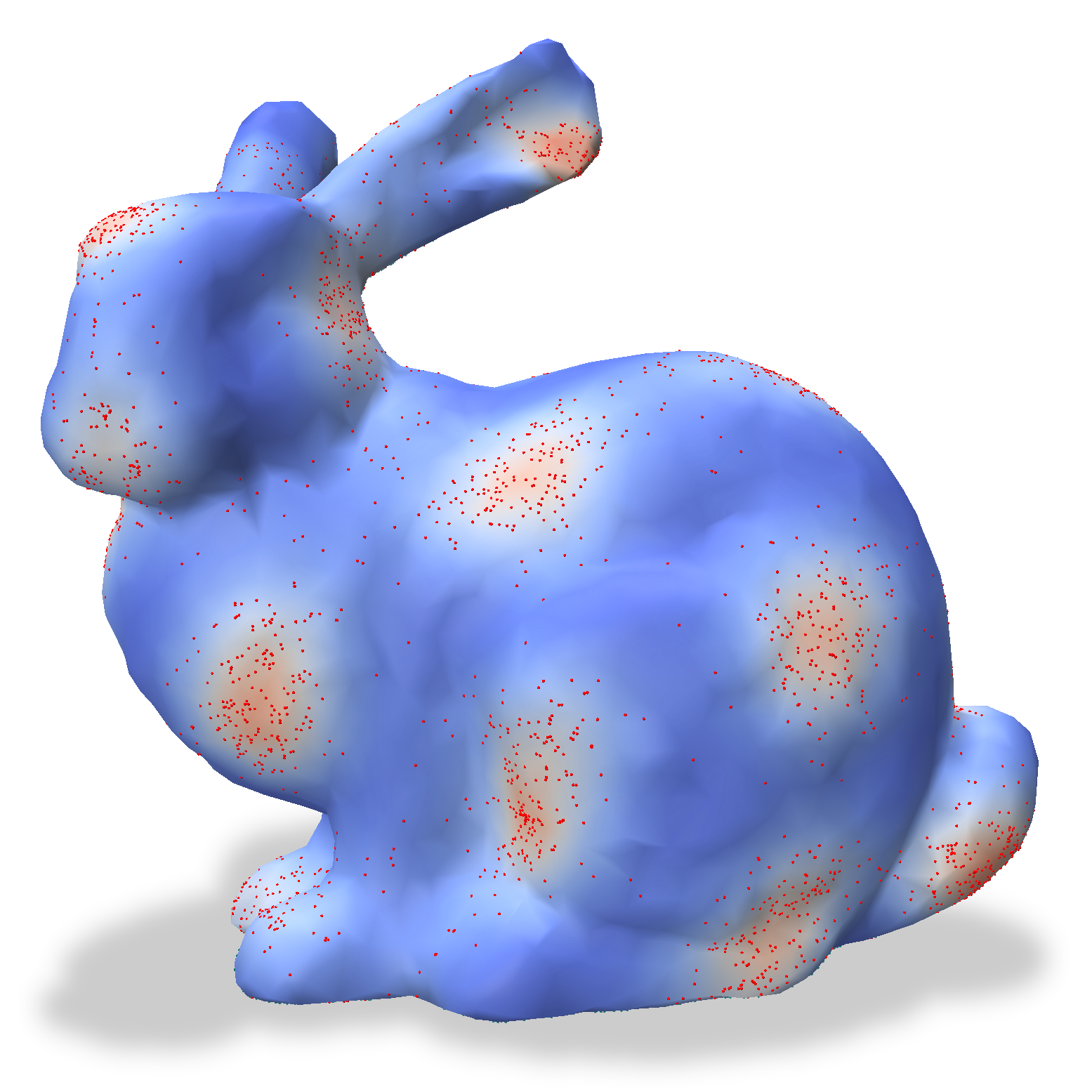} &
                        \includegraphics[width=1.1\linewidth]{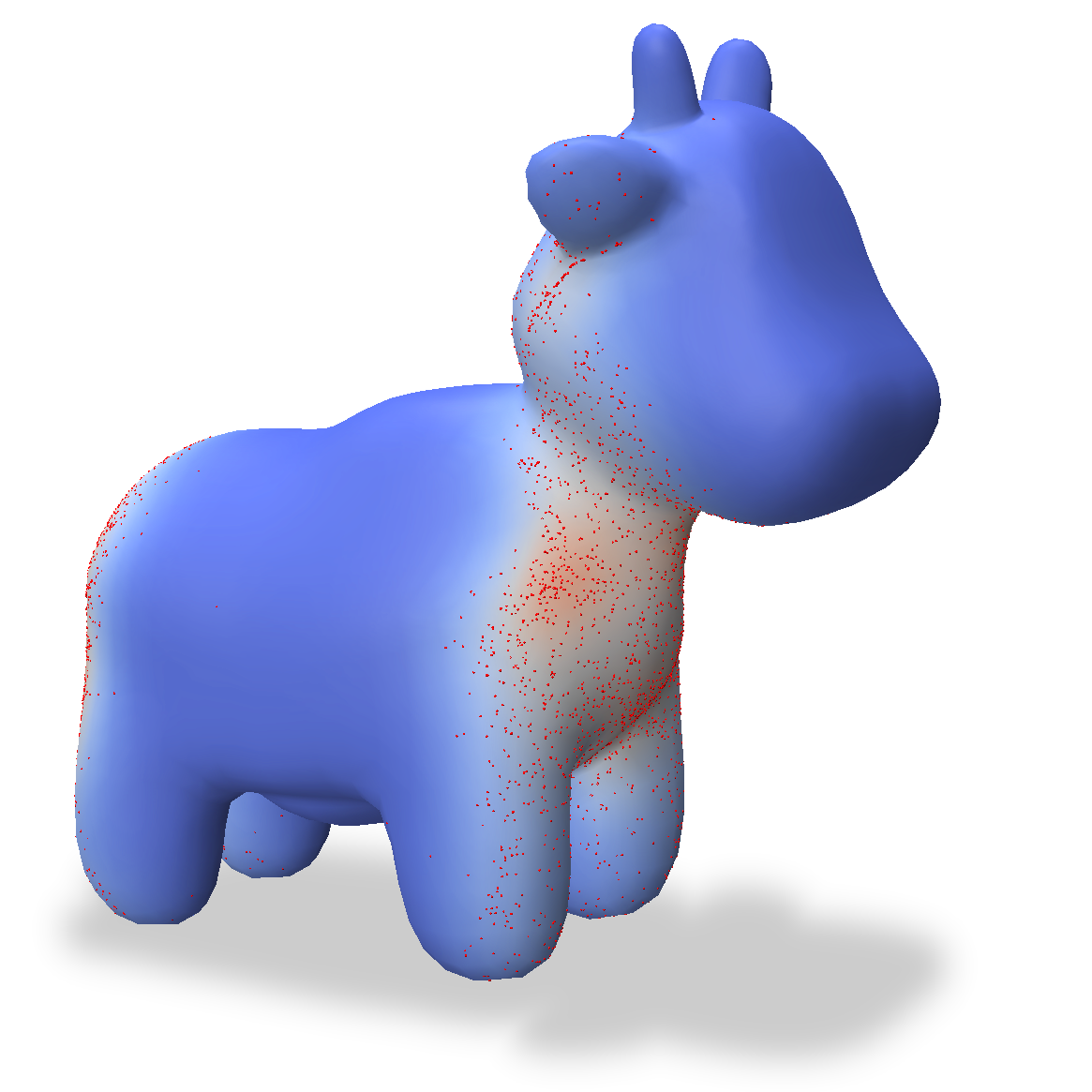} &
                        \includegraphics[width=1.1\linewidth]{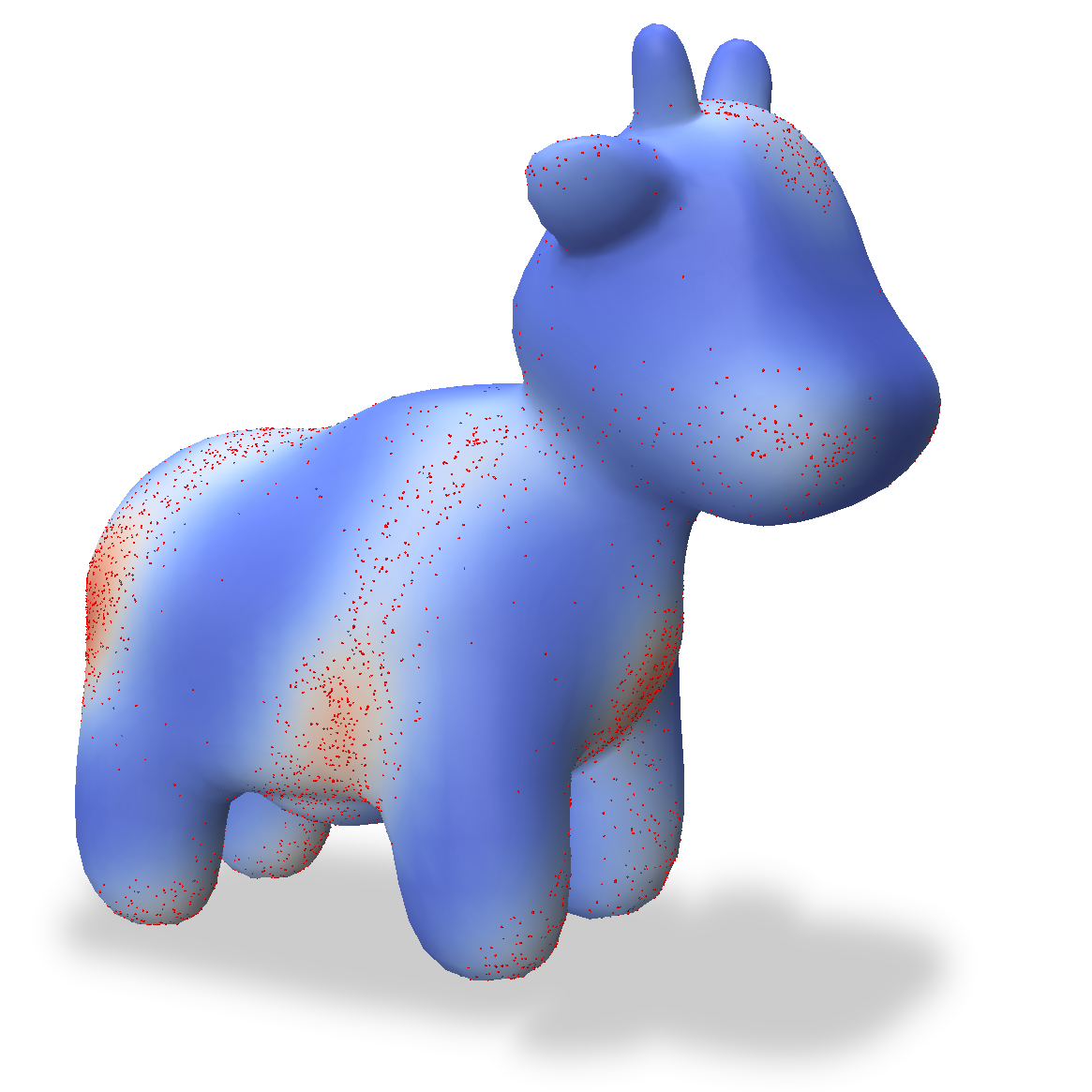} &
                        \includegraphics[width=1.1\linewidth]{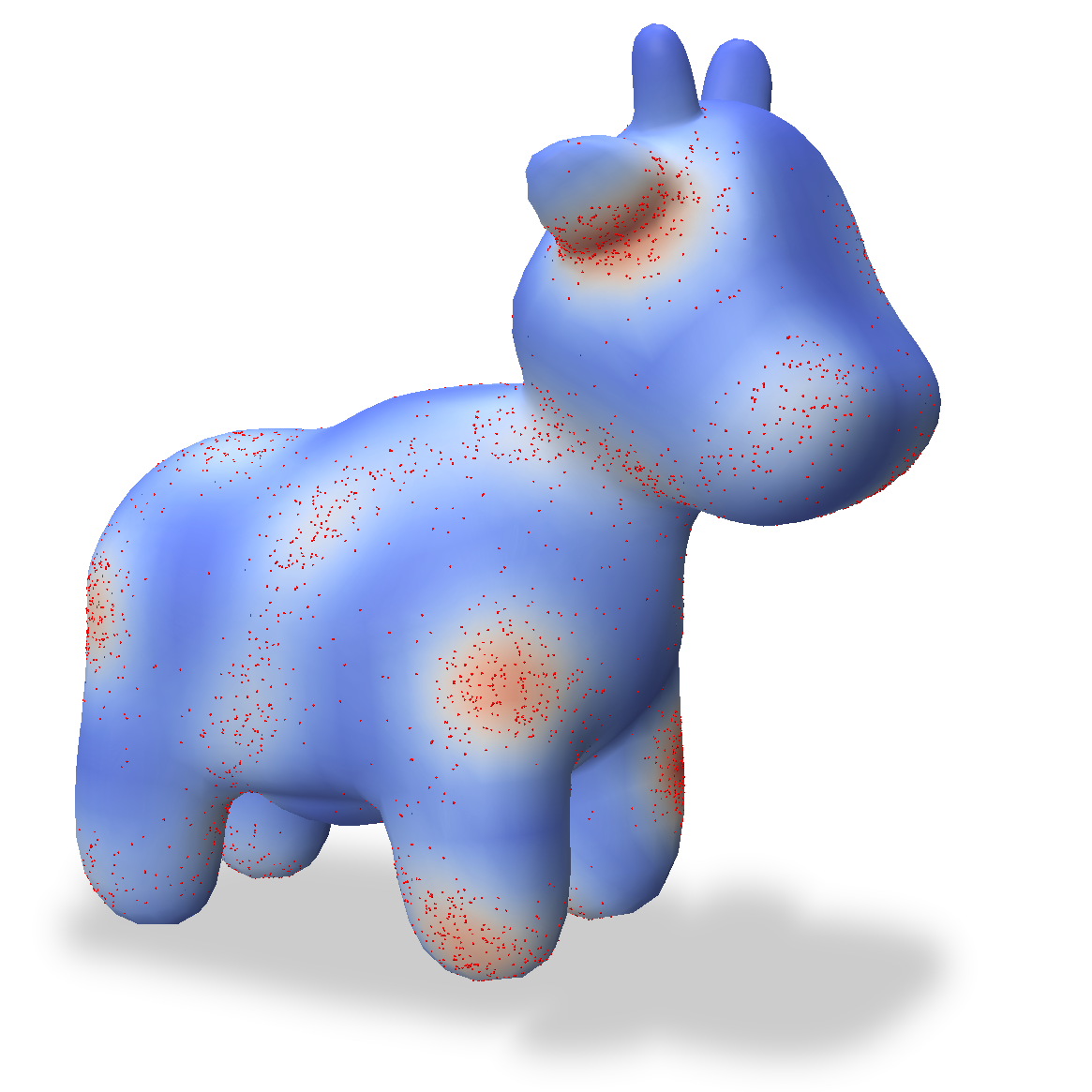} \\
                        
                        & \centering\scriptsize Bunny ($k$=10) & \centering\scriptsize Bunny ($k$=50) & \centering\scriptsize Bunny ($k$=100) 
                        & \centering\scriptsize Spot \\ ($k$=10) & \centering\scriptsize Spot \\ ($k$=50) & \centering\scriptsize Spot ($k$=100) \\
                    \end{tabular}
                }
                \vspace{-7pt}
                \caption{Comparison of eigenfunctions (GT), RFM predictions and our predictions on Bunny and Spot for different values of $k$.}
                \label{fig:eigen_vs_model}
                \vspace{-8pt}
            \end{figure}
            
        \paragraph{Results} 
            Like \cite{riemannian-fm, moserflow}, we use the eigenfunctions of the \emph{Stanford Bunny}~\cite{bunny-mesh} and \emph{Spot}~\cite{spot-mesh} as ground truth target distributions. %
            We report qualitative and quantitative comparisons in \cref{fig:eigen_vs_model} and \cref{tab:rfm-metrics}, respectively.
            \begin{table}[t]
                \centering
                \caption{\label{tab:rfm-metrics} Flow matching test metrics for different eigenvectors. $^*$~NLL is unreliable when comparing to RFM (see \cref{sec:flow}). \vspace{-4pt}}
                \resizebox{\linewidth}{!}{
                \begin{tabular}{clcccccc}
                \toprule
                Metric & Method &
                \multicolumn{3}{c}{\textbf{Stanford Bunny}} &
                \multicolumn{3}{c}{\textbf{Spot the Cow}} \\
                \cmidrule(lr){3-5} \cmidrule(lr){6-8}
                & &
                $k{=}10$ & $k{=}50$ & $k{=}100$ &
                $k{=}10$ & $k{=}50$ & $k{=}100$ \\
                \midrule
                                & RFM \cite{riemannian-fm}  & \textbf{1.22} & \textbf{1.44} & \textbf{1.55} & \textbf{1.03} & \textbf{1.14} & \textbf{1.26} \\
                \textbf{NLL}$^*$ ($\downarrow$) & MeshFlow-256 (ours)   & 1.49 & 1.74 & 1.75 & 1.34 & 1.42 & 1.38 \\
                                & MeshFlow-1024 (ours)  & 1.48 & 1.73 & 1.71 & 1.26 & 1.38 & 1.32 \\
                                
                \midrule
                                & RFM \cite{riemannian-fm}  & \textbf{0.050} & 0.11  & 0.13  & 0.087 & 0.065 & 0.084 \\
                \textbf{KLD} ($\downarrow$) & MeshFlow-256 (ours)   & 0.090 & 0.072 & 0.13  & 0.15  & 0.094 & 0.055 \\
                                & MeshFlow-1024 (ours)  & 0.060 & \textbf{0.036} & \textbf{0.058} & \textbf{0.050} & \textbf{0.044} & \textbf{0.034} \\
                \midrule
                                 & RFM \cite{riemannian-fm} & \textbf{0.014} & 0.018 & 0.020 & 0.013 & 0.014 & 0.014 \\
                \textbf{BCD} ($\downarrow$) & MeshFlow-256 (ours)             & 0.016 & 0.018 & 0.021 & 0.015 & 0.015 & 0.016 \\
                                 & MeshFlow-1024 (ours) & 0.015 & \textbf{0.015} & \textbf{0.017} & \textbf{0.012} & \textbf{0.013} & \textbf{0.013} \\
                \bottomrule
                \end{tabular}
                }        
                \vspace{-4mm}
            \end{table}
            Our method reports superior results on both KL divergence (KLD) and biharmonic-based Chamfer distances (BCD), which is the only geodesic measure. While we yield a higher negative log likelihood (NLL), this metric is inherently tied to the optimisation objective, making the NLL comparison informative only across the two variants of MeshFlow, which exhibits improvements when using more samples to compute the OT couplings.
            Our method is also $1.6 \cdot 10^4$ times faster during inference, and uses 97\% less GPU memory compared to RFM. We provide more details in \cref{sec-supp:otexpflow}.
        
    \noindent
    \subsection{Mesh-LBFGS optimiser}
        \label{sec:opt2ord}
        
        \paragraph{Problem \& Setup}
            The Broyden-Fletcher-Goldfarb-Shanno (BFGS) method~\cite{numeric-opti} is a classical quasi-Newton algorithm for unconstrained smooth optimisation. Quasi-Newton methods iteratively approximate the inverse Hessian of the objective function, thereby capturing curvature information without computing second derivatives explicitly. %
            However, the main drawback of the classical BFGS method is its storage cost. %
            L-BFGS~\cite{liu1989limited} avoids explicit storage, instead retaining only the most recent updates to gradient and position differences. %
            L-BFGS has become one of the most widely used large-scale optimisation algorithms and was adapted to Riemannian optimisation in~\cite{rlbfgs-imp, rbfgs2, yuan2016riemannian, birdal2019probabilistic} by replacing Euclidean operations with their manifold counterparts (\cref{sec:bg}). Here, we introduce the first Mesh LBFGS by using mesh operators like our Exponential map, which enables efficient execution of the optimiser thanks to its parallelisation capabilities.

            \noindent\hspace{\parindent} We apply our second order optimiser to Geodesic Centroidal Voronoi Tessellation (GCVT), which aims to divide a manifold into regions $\Omega_i$ based on the distance to a set of given points called seeds, $\mathbf{S} =\{\mathbf{s}_i \}_{i=1}^S \in \mesh^S$. Each region contains all locations closer to its seed than to any other. %
        
        \paragraph{Approach}
            We formulate the GCVT problem like in~\cite{vectorheat}, but we use our optimiser instead of the Lloyd's algorithm. We implement Mesh-LBFGS as described in~\cref{alg:mlbfgs}, adopting our differentiable straightest geodesics into \cite{rlbfgs-imp} to compute the exponential map and parallel transport. Instead of computing the gradient of a loss function, for this specific application, we compute the Karcher mean of the Voronoi regions like in~\cite{vectorheat}, as the vector representing its location relative to the current seed coincides with the direction of steepest descent. More details are provided in \cref{sec-supp:mlbfgs}.

\begin{figure}
                \centering
                \includegraphics[width=\linewidth]{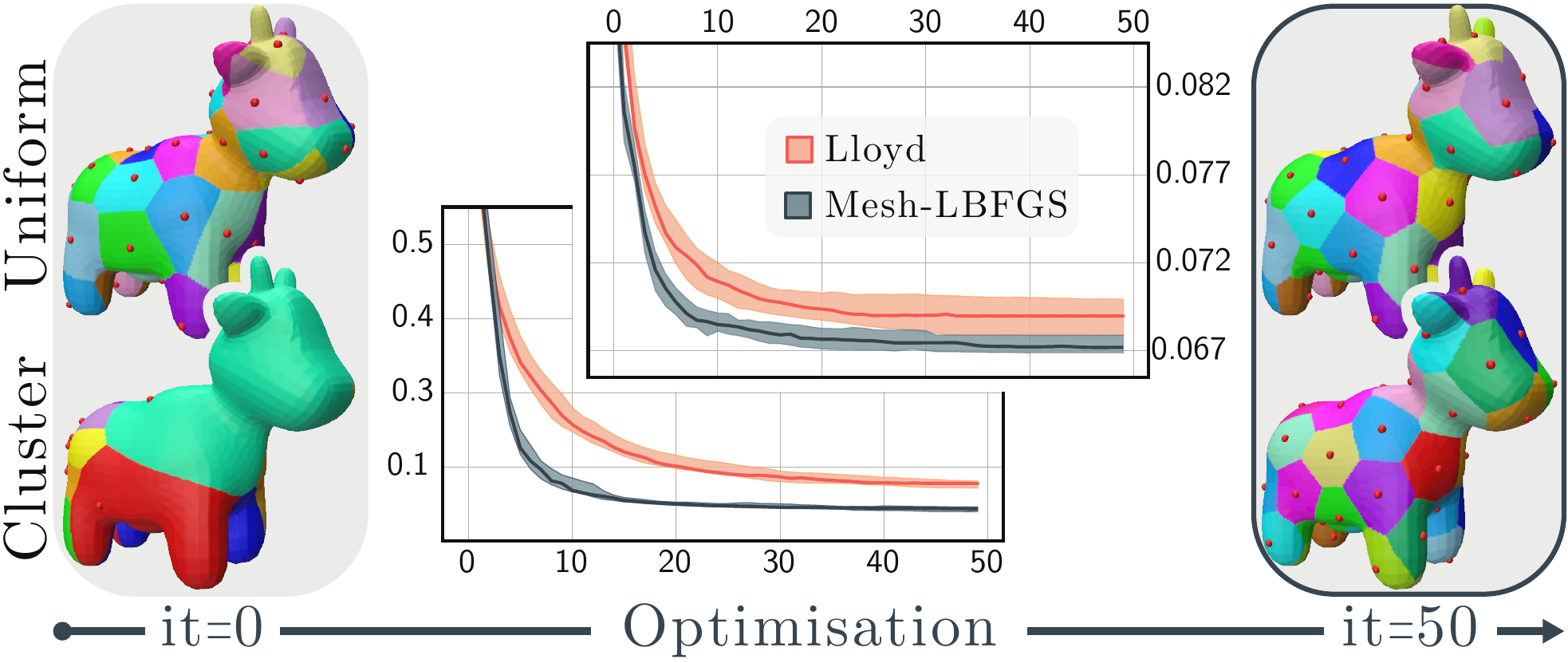}
                \caption{Mesh-LBFGS convergence with uniform and clustered initialisations against Lloyd (\textit{centre}). Tessellations are visualized before (\textit{left}) and after optimisation (\textit{right}) using our method.}
                \label{fig:gcvt}
                \vspace{-14pt}
            \end{figure}

        \paragraph{Results}
            We compare our implementation of Mesh-LBFGS with Lloyd's algorithm on two different cases for $50$ random seeds on Spot the Cow. In the first case seeds are uniformly sampled, in the second they are sampled on a small portion of the mesh (\cref{fig:gcvt}, \textit{left}). %
            Mesh-LBFGS uses an initial base learning rate $\eta_0$ and tests a maximum of three step lengths per iteration, $[\eta_0, 0.1\eta_0, 0.01\eta_0]$. Lloyd's algorithm, in contrast, makes exactly one function call per step.
            Mesh-LBFGS and Lloyd’s algorithm were tested on the same initial seeds over multiple runs, and the results are shown in \cref{fig:gcvt}. When comparing the two methods, it is important to consider not only the speed of convergence but also the number of function calls, since computing the Voronoi regions and the energy are the most expensive operations. Mesh-LBFGS often requires several function calls per step, but it compensates with faster convergence.
            The advantage of Mesh-LBFGS is particularly evident when the initial seeds are grouped together. In this case, the seeds start far from their optimal positions, and Mesh-LBFGS quickly corrects for this. By contrast, uniform sampling provides a better initial guess, so the improvement is less pronounced. Still, in both scenarios, our Mesh-LBFGS converges faster and achieves a better final solution (\cref{fig:gcvt}, \textit{centre} \& \textit{right}).

%% file: sec/6_conclusion.tex
\section{Conclusion}
    \label{sec:conclusion}

    We introduce the first fully differentiable and GPU-parallelized Exponential Map ($\text{Exp}$) for 3D meshes, based on straightest geodesics. Our method achieves orders-of-magnitude speed-ups across all discrete Riemannian operators it implements (Exp, tracing, and parallel transport). We derive two differentiation schemes: EP provides an efficient approximation for the initial vector derivative, while GFD offers a slower but more accurate derivative for both initial conditions. Both schemes generalize to alternative methods.
    
    The public release of our library, which includes our applications (convolution, flow model, and second-order optimizer), marks a key milestone for the community, accelerating future learning and optimization research on meshes.

%% file: sec/X_suppl.tex
\setlength{\columnsep}{0.3125in}
\twocolumn

\clearpage
\setcounter{page}{1}
\maketitlesupplementary
\appendix

\counterwithin{figure}{section}
\counterwithin{table}{section}
\counterwithin{equation}{section}
\counterwithin{algorithm}{section}

\section{Appendix}
    This document supplements our paper entitled \textbf{Parallelised Differentiable Straightest Geodesics for 3D Meshes} by providing further information on our core contributions, limitations and future directions, as well as additional implementation details for our differentiable straightest geodesics algorithm. We then describe the projection integration method we re-implement and show how our differentiation schemes can be used to improve their gradient estimation. We also provide a derivation of the closed-forms we use as benchmarks. Finally, for all our applications (i.e., AGC, MeshFlow, and Mesh-LBFGS) we provide more details and experimental results.

    \paragraph{Narrative description of our core contributions}
        Most modern machine learning tools struggle to correctly operate on surfaces discretised as meshes because the math behind them is slow and hard to differentiate. We bridge this gap by introducing a way to calculate Exponential maps on 3D meshes that is fully compatible with modern machine learning (\cref{sec:dpsg}). We first moved these straightest geodesics algorithm to the GPU, making it up to three orders of magnitude faster than current methods. More importantly, we created two new ways to let gradients flow through the Exponential map: a fast approximation and a high-accuracy version. For the first time, a machine learning model can learn the best way to move across a surface.
        
\begin{figure}[b]
            \centering
            \includegraphics[width=.8\linewidth]{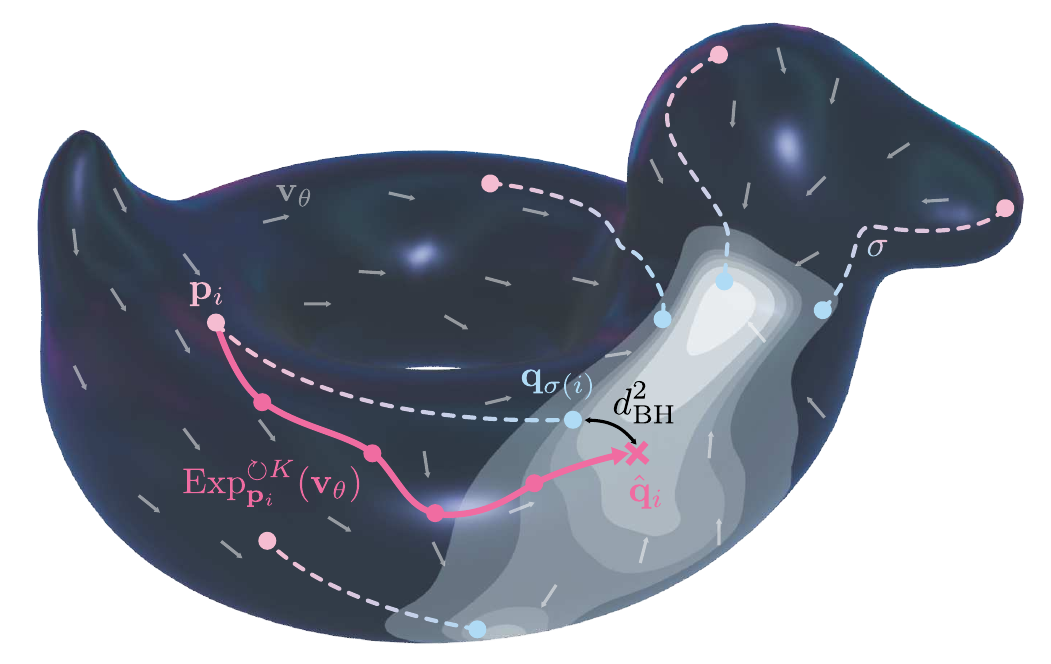}
            \vspace{-6pt}
            \caption{Visual representation of MeshFlow. Noise samples, $\p_i$, are displaced via our $K$-steps exponential map according to the directions dictated by our learnable static vector field, $\v_\theta$. The end-points, $\hat{\q}_i$, are compared with the OT-coupled training samples, $\q_{\sigma(i)}$, using a squared biharmonic distance function $d_{BH}^2$. }
            \label{fig:meshflow}
            \vspace{-10pt}
        \end{figure}

        To show what these tools can actually do, we applied them to three real-world problems that have traditionally been difficult to solve on 3D surfaces. First, we built Adaptive Geodesic Convolutions (AGC) (\cref{sec:geoconv}), which use our differentiable Exponential map to learn exactly how wide or narrow its receptive field should be for every single layer, allowing it to segment complex shapes like human body parts with much better accuracy than models with fixed kernel sizes. Next, we developed MeshFlow (\cref{fig:meshflow}), a generative model that can transform a learn how to flow distributions defined on a mesh through a single static velocity field. MeshFlow moves sampled points natively along the mesh and is incredibly efficient, being its work roughly $16,000$ times faster and using $97\%$ less memory than Riemannian Flow Matching (RFM)~\cite{riemannian-fm}. Finally, we introduced the Mesh-LBFGS optimiser (\cref{sec:opt2ord}), a high-speed second-order optimiser for meshes. Our optimiser uses the surface’s curvature to take much smarter, larger leaps, reaching a perfect distribution of points in far fewer steps.
    
    \paragraph{Limitations \& Future Work}
        In this work, we propose two differentiation schemes for the exponential map of 3D meshes, enabling backpropagation through this operator for the first time on meshes. Both methods are completely independent of the implementation of the exponential map, but each suffers from different limitations. The EP scheme is faster, but can be used only to approximate the Jacobian with respect to the initial vector $\v$, which is still less accurate than GFD. The GFD scheme can accurately differentiate with respect to both $\p$ and $\v$ at the expense of a slightly increased computational cost, caused by the requirement to trace multiple geodesics. While this is not a problem for our parallelised GPU implementation of the geodesic tracing algorithm, it may limit the applicability of alternative exponential maps in a learning framework (e.g., see our experiments with PI in \cref{fig:benchmark-backward}). Future work shall seek a differentiation scheme capable of achieving computational speeds comparable to EP, while retaining the accuracy of GFD.

    \paragraph{Note on differences with Eikonal solvers}
        The main difference between our method and Eikonal solvers~\cite{huberman2023deep, lichtenstein2019deep} is that while we solve an initial value problem, they solve boundary value problems.
        Eikonal solvers, solving Hamilton–Jacobi equations, explicitly work on minimising the distance (gradient descent after obtaining the distance field). Distance fields are scalar PDE objects not directly suited for computing exponential maps or parallel transport. Our approach does not provide a global distance field, but directly gives the discrete analogues of the Riemannian operators. Eikonal solvers such as the Deep Eikonal Solvers (DES) compute ``shortest" paths whereas we compute ``straightest" paths. While on smooth manifolds these coincide (0 covariant acceleration = length of the curve that minimises the path length), on discrete surfaces they do not. This is a geometric difference, which will naturally lead to algorithms computing a different quantity. At first, they look similar, but this can create very different behaviours when examined closely, especially on meshes with non-uniform triangle sizes.

        Moreover, naive fast marching (used for Eikonal solvers) is also non-differentiable and several works seek to make it so (like DES - deep eikonal solvers). They are also harder to parallelise as front ordering is naturally sequential (pop-min from wavefront) - ours is embarrassingly parallel (per traced geodesic). 

        Finally, (i) we directly discretise the geodesic equation itself, whereas Eikonal solvers do so indirectly - as a result, arguably, ours remain closer to the true notion of a ``geodesic" whatever it might mean in the discrete setting; (ii) our work also gives a natural connection therefore a parallel transport (which we use in LBFGS); these are harder to obtain for DES-like methods.

    \begin{figure}
        \centering
        \includegraphics[width=1\linewidth]{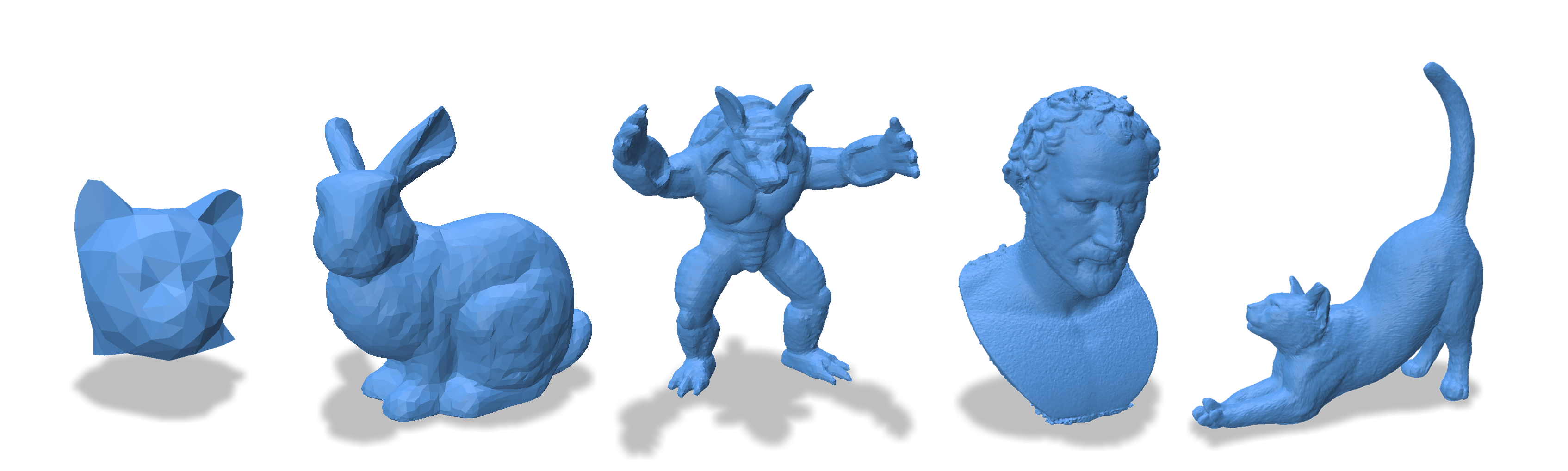}
        \caption{Meshes used to evaluate our method in \cref{sec:acc-perf}: cat head (248 faces), bunny (4,968 faces), armadillo (29,170 faces), Demosthenes (240,293 faces), and cat (1,698,248 faces) \cite{odedstein-meshes}}
        \label{fig:meshes-benchmarking}
    \end{figure}

    \subsection{Additional Details on Our Method}
        \label{sec-supp:geostep}

        \paragraph{Implementation of the Geodesic Step}
        While straightest geodesics can be formulated from a purely intrinsic perspective, as mentioned in \cref{sec:dpsg}, by using extrinsic 3D normals and rotations, we can replicate the intrinsic logic with more efficient tensor operations. The geodesic step reported in \cref{alg:gpu_parallel} and detailed in \cref{alg:sg-step} still distinguishes between three possible traces: on a face, across an edge, or across a vertex. On a face, the method uses barycentric coordinates to trace a straight line that terminates within the face when reaching the maximum length or continues until it intersects either a vertex or an edge. On edges, the angle preservation criteria corresponding to the unfolding of the triangles adjacent to the edge is replaced by computing the 3D dihedral angle between the face normals and extrinsically rotating the incoming vector around the common edge. At vertices, after identifying the correct outgoing edge by accumulating intrinsic angles to satisfy $\theta_l = \theta_r$, we determine the outgoing vector by applying a 3D rotation to that edge's vector on the new 3D face plane.

        \input{algorithms/geodesic_step}

        \StartTightFigureParagraph
        \paragraph{Main differences with SotA implementation}
            Geometry central processes meshes using a half-edge representation, thus requiring meshes to be orientable. However, in 
            {\parfillskip0pt\par}
            \begin{wrapfigure}[8]{R}[0pt]{0.25\linewidth}
                \centering
                \vspace{-8pt}
                \includegraphics[width=\linewidth]{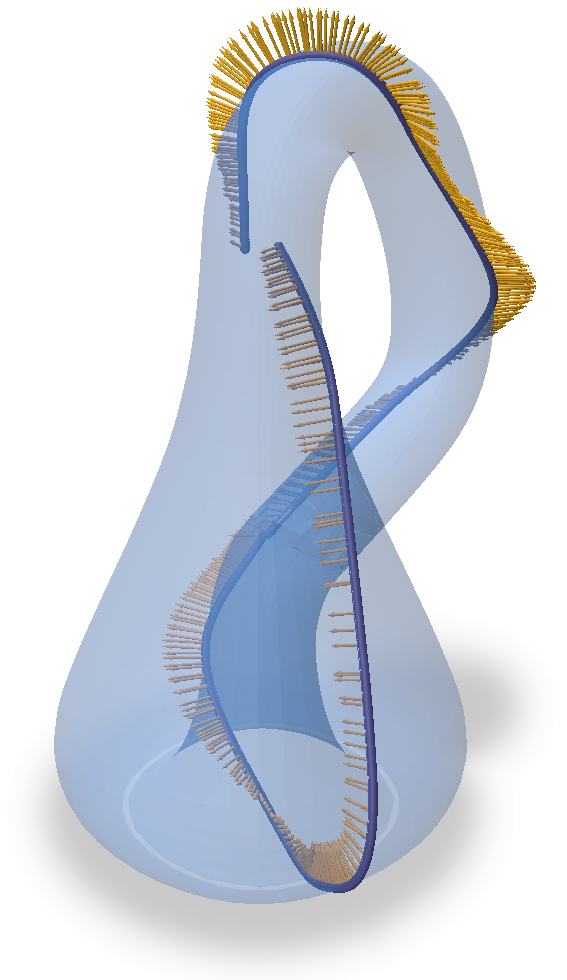}
            \end{wrapfigure}
            \noindent 
            our implementation, we only store triangle adjacencies to trace geodesics, making it compatible with non-orientable meshes, like a Möbius strip or Klein bottle. We here report a geodesic trace on the Klein bottle computed with our implementation while showing also the normals along the geodesic.
    
            Geometry central also does not handle traces crossing vertices, and only considers that the trace will cross edges. This can lead to some imprecisions when using high resolution meshes or low precision floats.

            An additional distinction, previously discussed in \cref{sec:dpsg} lies in the behaviour when encountering a mesh boundary. Instead of terminating the geodesic, we continue tracing along the boundary until the original tracing direction can be resumed. We highlight the essential modifications in blue in \cref{alg:sg-step-vertex} and \cref{alg:sg-step-edge}. We show how this can improve convergence performance on meshes with holes in \cref{sec-supp:otexpflow}. In \cref{alg:sg-step-vertex}, the transport error depends on two factors: it measures how far the direction deviates from the target face, and it ensures that the projected direction keeps the point inside that face. If no faces meet this criterion, we select the face adjacent to the hole and trace the geodesic along the edge.
            
            \EndTightFigureParagraph

            \input{algorithms/transport_vertex}

            \input{algorithms/transport_edge}

        \paragraph{Non-auto-differentiability of straightest geodesics}
            \begin{figure}[b]
                \centering
                \includegraphics[width=0.8\linewidth]{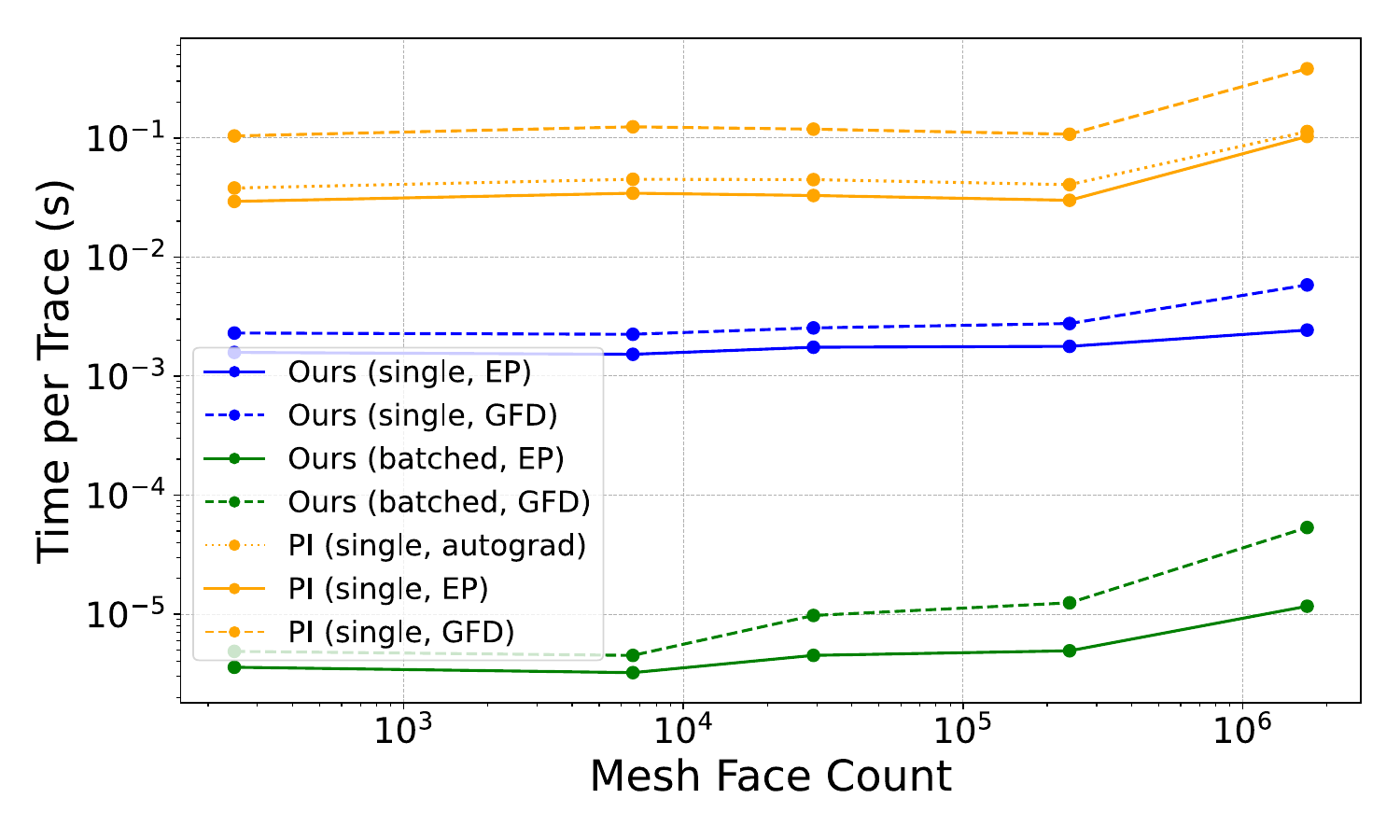}
                \vspace{-6pt}
                \caption{Mean time per trace for forward and backward pass using our method and PI with different differentiation schemes.}
                \label{fig:benchmark-backward}
                \vspace{-10pt}
            \end{figure}
            As mentioned in \cref{sec:dpsg}, proven in \cite{chen2022projective}, and confirmed by our experiments where we let $\mathrm{autograd}$ automatically differentiate through the exponential map of \cite{Madan2025local-parameterizations} (\cref{fig:diff-acc}), automatic differentiation is not suitable in non-Euclidean settings. 
            Even if we pretend to ignore this fact, all implementations of straightest geodesics are algorithms characterised by discrete combinatorial choices. As such, automatic differentiation cannot even be applied in practice. In \cref{alg:sg-step,alg:sg-step-edge,alg:sg-step-vertex} we highlight in red the non-differentiable operations that PyTorch’s $\mathrm{autograd}$ cannot handle automatically. In particular, searching for specific edges/faces and for indices that meet certain conditions would all require computing gradients with respect to integer indices, which cannot be handled automatically. In addition, moving points intrinsically defined by their barycentric coordinates, which change arbitrarily from face to face, requires hard resets and per-face computations. Another issue is differentiating with respect to $\v$ as the number of iterations is a dynamic function of $\|\v\|$ and $\mathrm{autograd}$ can't differentiate with respect to the number of times a loop runs.
            Finally, a native autograd approach is also prohibitively inefficient, as it would require storing intermediate states, which would increase the time and memory required to compute the gradient.
            To evaluate these issues, we compare $\mathrm{autograd}$ against our differentiable scheme using both our straightest-geodesic implementation and the PI exponential map. We compute geodesic traces on various meshes and optimise for an arbitrary target point using the Euclidean distance; this choice isolates the performance of the differentiation schemes themselves, without introducing additional complexity from using a geodesic metric such as the biharmonic distance. The results, shown in \cref{fig:benchmark-backward}, report the computational performance for our method with both differentiation schemes (EP and GFD), as well as for PI using EP, GFD, and $\mathrm{autograd}$. As discussed above, $\mathrm{autograd}$ is substantially slower because it must store intermediate values, whereas the EP scheme requires only the final state, making it far more efficient.

        \paragraph{Speed-up GFD and EP through batching}
            The computational speed of our GFD differentiation scheme is primarily influenced by the number of $\Exp$ that need to be computed. Nevertheless, the number of $\Exp$ calls can be significantly reduced by batching. In fact, $\Exp (\p, \v)$, $\Exp(\p, \v + \varepsilon_\v \versor{\perp})$, $\Exp ( \p, \varepsilon_\p \versor{u})$, $\Exp ( \p, \varepsilon_\p \versor{v})$ can all be combined in a single batched tracing operation by stacking the input arguments. A second execution of the tracing algorithm can be computed by combining the following exponential map computations: $\Exp \big(\p', \prod_{\p}^{\p'}(\varepsilon_\v \versor{\parallel}) \big)$, $\Exp(\p_u, \v_u)$, and $\Exp(\p_v, \v_v)$. As we can see in \cref{fig:benchmark-backward}, batching provides a significant reduction in computational time. On the other hand, EP does not require the additional tracing, but can easily be batched using batched tensor operations.

    \subsection{Projection Integration Exp Map}
        \input{algorithms/projection_exp_map}
        \StartTightFigureParagraph
        \noindent We re-implemented the Projection Integration (PI) exponential map following the formulation in \cite{Madan2025local-parameterizations} and compared it with our method as well as the implementation provided by Geometry Central \cite{geometrycentral}. Our implementation is detailed in \cref{alg:projection-exp}. The projection step computes the projection of the point onto every face of the mesh, identifies the faces that contain the point, and selects the one that minimises the distance. This operation is computationally expensive and must be performed at every step. While the number of steps is independent of the mesh resolution (unlike in straightest geodesic implementations), the projection step scales with mesh resolution, thereby increasing the overall computational cost of tracing. Implementing the projection on the 
        {\parfillskip0pt\par}
        \begin{wrapfigure}[10]{R}[0pt]{0.5\linewidth}
            \centering
            \vspace{-8pt}
            \includegraphics[width=\linewidth]{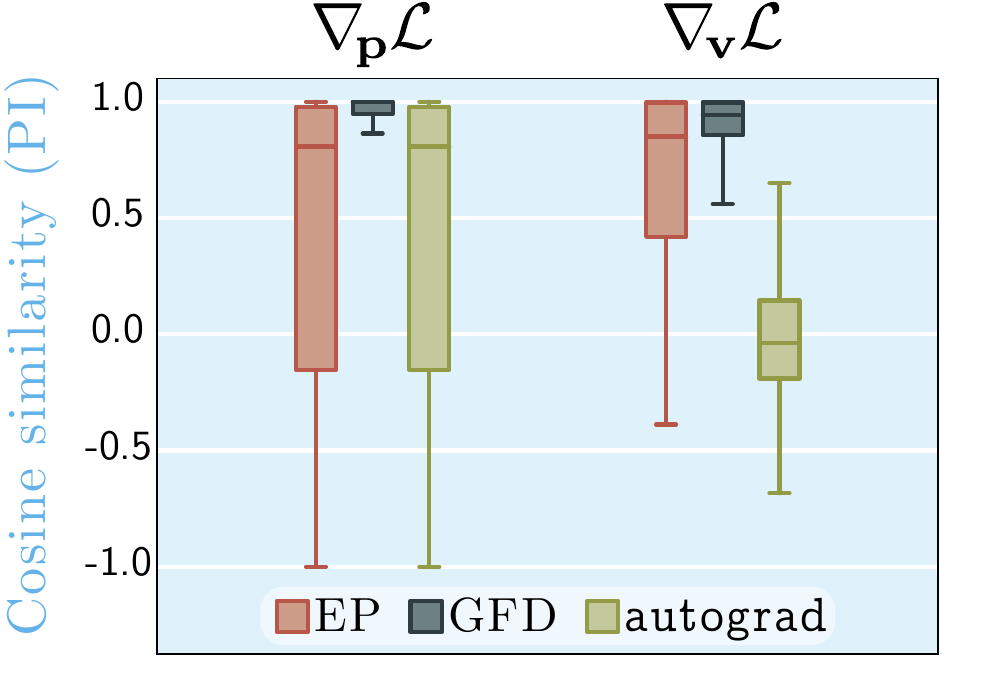}
            \vspace{-20pt}
            \caption{Differentiation correctness of PI across differentiation schemes.}
            \label{fig:PIdiff-acc}
        \end{wrapfigure}
        \noindent
        GPU is relatively straightforward, as it primarily involves tensor operations that are easily parallelisable. However, even with GPU acceleration, the computation of this Exp map remains substantially slower than the other methods.
        
        In \cref{fig:PIdiff-acc} we evaluate the differentiation correctness of PI when using $\mathrm{autograd}$ as well as our differentiation schemes. Cosine similarity results on PI remain consistent with previous findings leveraging our implementation of the straightest geodesics to compute the exponential map. This result is expected as the accuracy of both exponential maps was also comparable thanks to the choice of a small step size in PI (see \cref{tab:accuracy-combined}).
        \EndTightFigureParagraph

    \subsection{Exp map on the Sphere}
        \label{sec-supp:grad_sph}
        Given a starting point $\p \in \mathcal{S}^2$ and direction $\v \in \TpM$. We assume these are given in Cartesian coordinates. The exponential map on the sphere is then given by~\cite{birdal2018bayesian, riemannian-fm}:
        \begin{equation*}
            \Exp_{\p}(\v) = \cos(||\v||) \p + \sin(||\v||) \frac{\v}{||\v||}
        \end{equation*}
        The Jacobian with respect to $\p$ can be computed differentiating with respect to $\p$:
        {\small
        \begin{equation}
            \label{eq:jp_sphere}
            \J_p = \cos(\|\v\|)\mat{I}
        \end{equation}
        }
        Computing the Jacobian with respect to $\v$ is slightly more cumbersome, but still achievable via simple differentiation. Thus, we have:
        {\small
        \begin{equation}
            \label{eq:jv_sphere_start}
            \J_\v = \frac{\partial}{\partial \v}\Big[ \p \cos(\|\v\|) \Big] + \frac{\partial}{\partial \v}\left[ \frac{\v}{\|\v\|} \sin(\|\v\|) \right].
        \end{equation}
        }
        We now differentiate the two terms separately, while knowing that $\frac{\partial}{\partial \v} \|\v\| = \frac{\v^\top}{\|\v\|}$. The first term of \cref{eq:jv_sphere_start} is:
        {\small
        \begin{align}
            \label{eq:jv_partial1}
                \frac{\partial}{\partial \v}\Big[ \p \cos(\|\v\|) \Big] &= \p \frac{\partial}{\partial \v}\Big[ \cos(\|\v\|) \Big]
                = - \p \sin(\|\v\|) \frac{\partial}{\partial \v} \|\v\| \notag \\
                &= - \p \sin(\|\v\|) \frac{\v^\top}{\|\v\|}
        \end{align}
        }
        The second term of \cref{eq:jv_sphere_start} is:
        {\small
        \begin{equation*}
            \label{eq:jv_partial2}
                \frac{\partial}{\partial \v} \! \left[ \frac{\v}{\|\v\|} \sin(\|\v\|) \right] = \frac{\partial}{\partial \v} \! \left[ \frac{\v}{\|\v\|}\right] \sin(\|\v\|) + \frac{\v}{\|\v\|} \frac{\partial}{\partial \v} \sin(\|\v\|).
        \end{equation*}
        }
        If we first solve
        {\small
        \begin{equation*}
            \frac{\partial}{\partial \v} \! \left[ \frac{\v}{\|\v\|}\right] = \frac{\|\v\|\mat{I} - \v \frac{\v^\top}{\|v\|}}{\|\v\|^2} = \frac{\|\v\|^2\mat{I} - \v \v^\top}{\|\v\|^3} = \frac{\mat{I}}{\|\v\|} - \frac{\v\v^\top}{\|\v\|^3}
        \end{equation*}
        }
        \noindent we can then finish differentiating the second term of \cref{eq:jv_sphere_start}:
        {\small
        \begin{equation*}
                \frac{\partial}{\partial \v} \! \left[ \frac{\v}{\|\v\|} \sin(\|\v\|) \right] \! = \! 
                \left[\frac{\mat{I}}{\|\v\|} \! - \! \frac{\v\v^\top}{\|\v\|^3} \right]\!\sin(\|\v\|) + \frac{\v\v^\top}{\|\v\|^2}\!\cos(\|v\|)
        \end{equation*}
        }
        Putting everything back together in \cref{eq:jv_sphere_start}, we have:
        {\small
        \begin{equation}
            \label{eq:jv_sphere}
            \J_\v = \left[\frac{\mat{I} - \p\v^\top}{\|\v\|} \! - \! \frac{\v\v^\top}{\|\v\|^3} \right]\!\sin(\|\v\|) + \frac{\v\v^\top}{\|\v\|^2}\!\cos(\|v\|)
        \end{equation}
        }
        Then, the gradients of the loss $\mathcal{L}$ become $\nabla_\p \mathcal{L} = \nabla_\q \mathcal{L} \cdot \, \J_p$ and $\nabla_\v \mathcal{L} = \nabla_\q \mathcal{L} \cdot \, \J_v$

        \begin{figure*}
            \vspace{-1.6em}
            \centering
            \begin{subfigure}{0.33\textwidth}
                \centering
                \includegraphics[width=\linewidth]{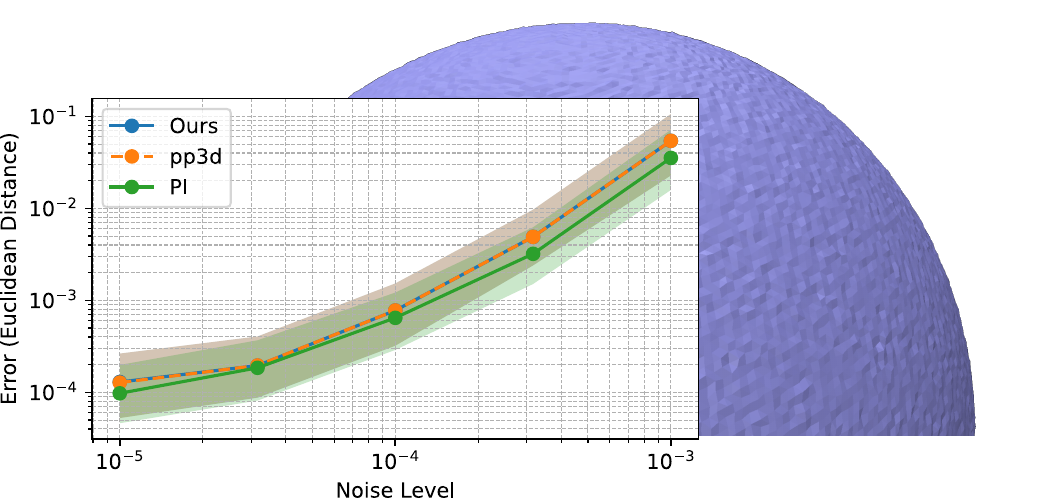}
                \caption{\label{fig:error_sphere_noise}Sphere with noise}
            \end{subfigure}
            \hfill
            \begin{subfigure}{0.33\textwidth}
                \centering
                \includegraphics[width=\linewidth]{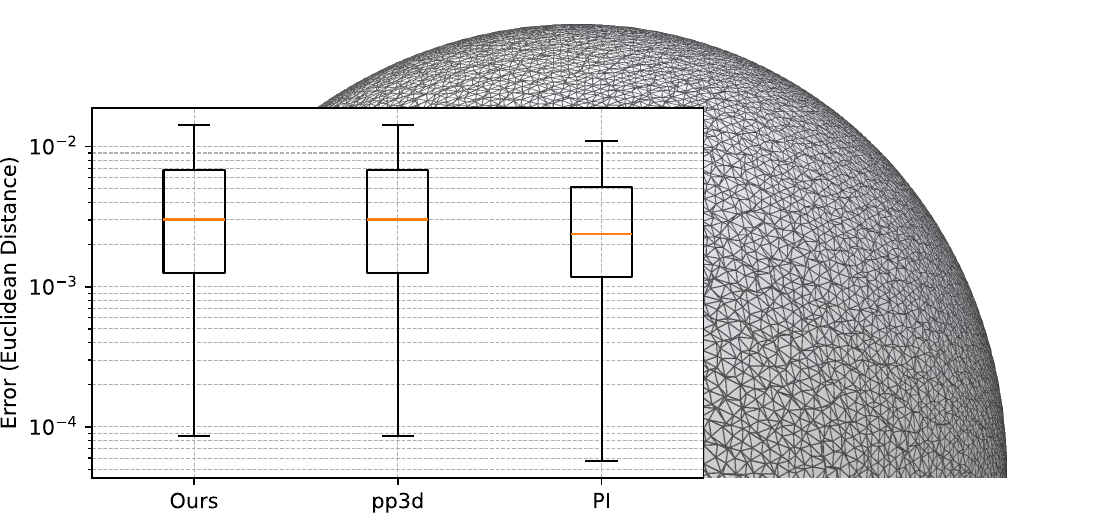}
                \caption{\label{fig:error_sphere_slim}Sphere with slim triangles}
            \end{subfigure}
            \hfill
            \begin{subfigure}{0.33\textwidth}
                \centering
                \includegraphics[width=\linewidth]{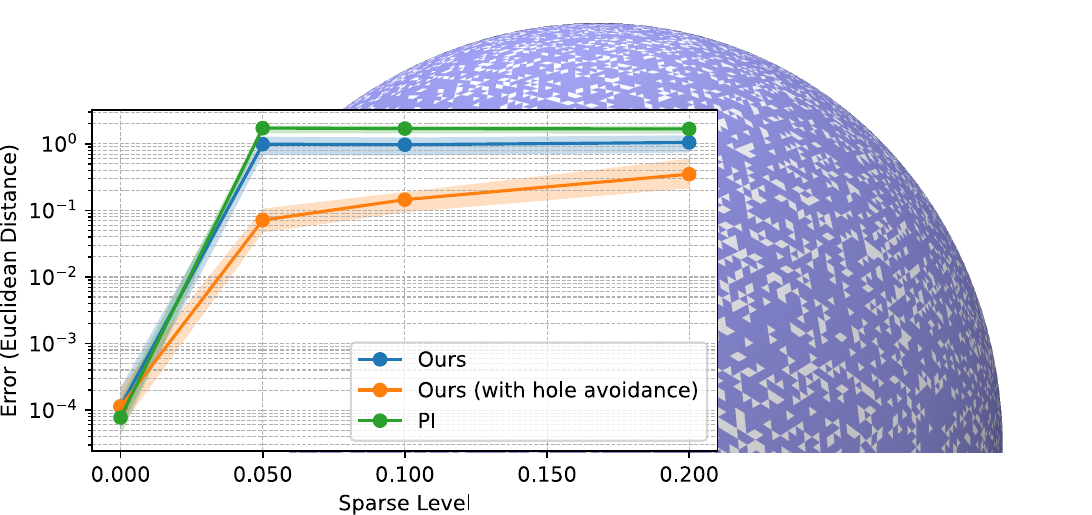}
                \caption{\label{fig:error_sphere_holes}Sphere with holes}
            \end{subfigure}
            \vspace{-2.2em}
            \caption{\label{fig:acc_corrupted_sphere}Exponential map accuracy on sphere meshes with different types and amounts of corruption.\vspace{-1mm}}
        \end{figure*}
        
        \begin{figure*}
            \centering
            \begin{subfigure}{0.3\textwidth}
                \centering
                \includegraphics[width=\linewidth]{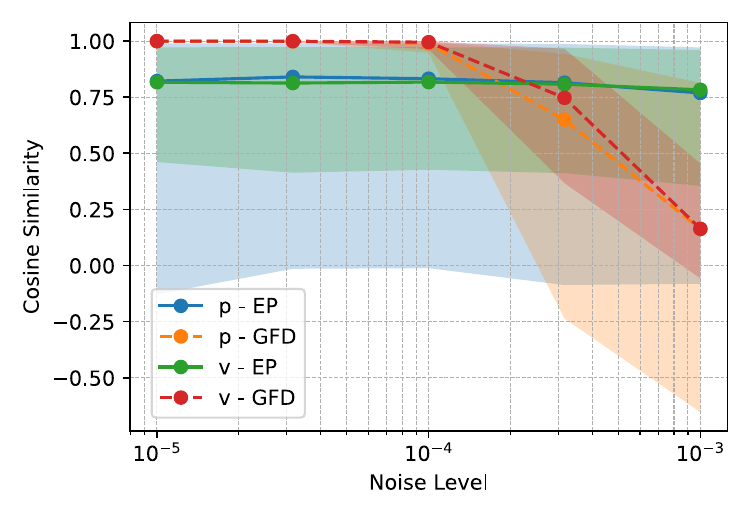}
                \caption{\label{fig:cossim_sphere_noise}Sphere with noise}
            \end{subfigure}
            \hfill
            \begin{subfigure}{0.3\textwidth}
                \centering
                \includegraphics[width=\linewidth]{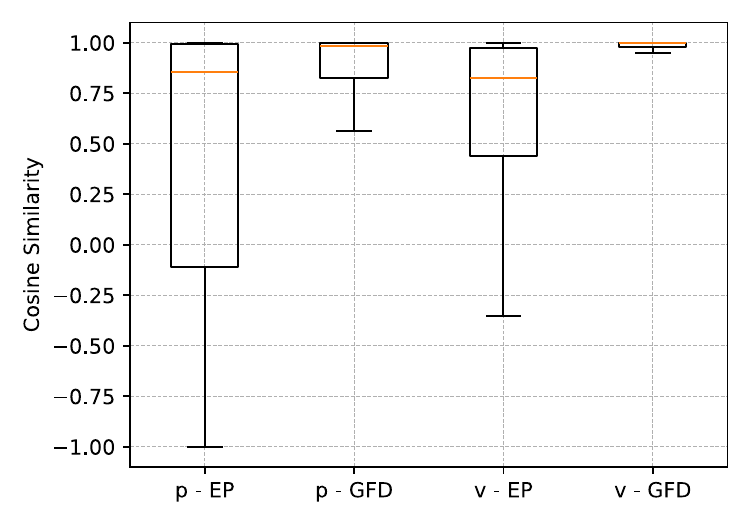}
                \caption{\label{fig:cossim_sphere_slim}Sphere with slim triangles}
            \end{subfigure}
            \hfill
            \begin{subfigure}{0.3\textwidth}
                \centering
                \includegraphics[width=\linewidth]{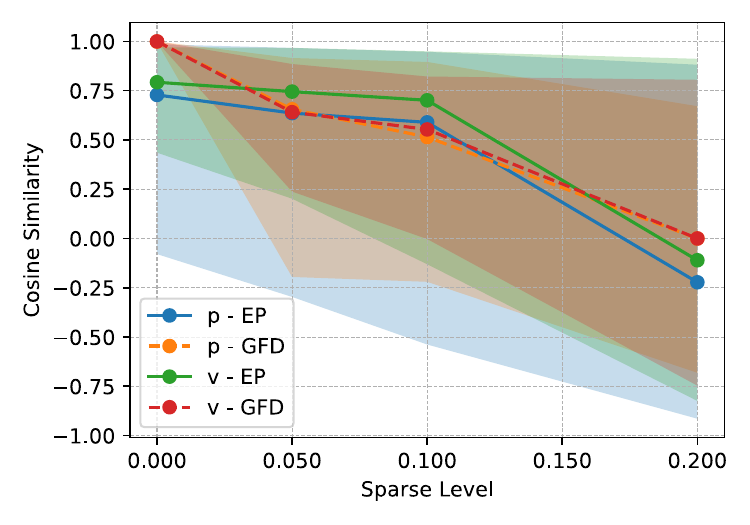}
                \caption{\label{fig:cossim_sphere_holes}Sphere with holes}
            \end{subfigure}
            \vspace{-.9em}
            \caption{\label{fig:cossim_corrupted_sphere}Cosine similarity of the gradients using our method on spheres (using the GT exp map) and other meshes (using ODEs).}
        \end{figure*}

    \subsection{IVP on Torus}
        \label{sec-supp:ivptorus}
        We consider a torus embedded in $\mathbb{R}^3$, with major and minor radius $R>r>0$. Its smooth parametrisation is given by:
        \begin{equation*}
            F(\alpha,\beta)=
            \begin{pmatrix}
                (R+r\cos \beta)\cos \alpha \\
                (R+r\cos \beta)\sin \alpha \\
                r \sin \beta
            \end{pmatrix}.
        \end{equation*}
        The tangent vectors are then given by:
        \begin{equation*}
            F_\alpha = \frac{\partial F}{\partial \alpha} = 
            \begin{pmatrix}
                -(R+r\cos \beta)\sin \alpha \\
                (R+r\cos \beta)\cos \alpha \\
                0
            \end{pmatrix}, 
        \end{equation*}
        \begin{equation*}
            F_\beta = \frac{\partial F}{\partial \beta} = 
            \begin{pmatrix}
                -r\sin \beta\cos \alpha \\
                -r\sin \beta\sin \alpha \\
                r \cos \beta
            \end{pmatrix}.
        \end{equation*}
        Given the smooth parametrisation of the torus, $F$, we can compute the metric tensor, which defines the local curvature,
        \begin{equation}
        \label{eq:torus-metric}
        \begin{aligned} 
        g &= 
        \begin{pmatrix}
            \langle F_\alpha, F_\alpha \rangle & \langle F_\alpha, F_\beta \rangle \\
            \langle F_\beta, F_\alpha \rangle & \langle F_\beta, F_\beta \rangle \\
        \end{pmatrix} \\
        &=
        \begin{pmatrix}
            (R+r\cos \beta)^2 & 0 \\
            0 & r^2 \\
        \end{pmatrix}
        \end{aligned}
        \end{equation}
        The Christoffel symbols are defined as:
        \[
        \Gamma^k_{\;ij}
        = \tfrac{1}{2} \, g^{k\ell}
        \left(
            \frac{\partial g_{\ell i}}{\partial x^j}
          + \frac{\partial g_{\ell j}}{\partial x^i}
          - \frac{\partial g_{ij}}{\partial x^\ell}
        \right)
        \]
        For vector fields $\mathcal{V},\mathcal{W} : \Man \to \mathcal{T}\Man$, the Christoffel symbols act as the correction terms needed to express differentiation while staying consistent with the manifold's geometry,
        \[
        (\nabla_\mathcal{V}\mathcal{W})^k = \sum_i \mathcal{V}^i\frac{\partial \mathcal{W}^k}{\partial x^i} +\sum_{i,j} \Gamma^k_{i,j}\mathcal{V}^i\mathcal{W}^j
        \]
        Using \cref{eq:torus-metric}, the non-zero Christoffel symbols can be computed as, 
        \begin{equation*}
            \Gamma^{\alpha}_{\alpha\beta} = \Gamma^{\alpha}_{\beta\alpha} = -\frac{r \sin \beta}{R+r\cos \beta},
            \quad 
            \Gamma^{v}_{\alpha\alpha} = \frac{(R+r\cos \beta)\sin \beta}{r}.
        \end{equation*}  
        Injecting the Christoffel symbols inside the geodesic equation, $\nabla_{\dot{\gamma}(t)}\dot{\gamma}(t)=0$, gives:
        \begin{equation}
        \label{eq:torus-geodesic}
        \left\{
        \begin{aligned}
            & \alpha'' - 2 \frac{r\sin \beta}{R+r\cos \beta}\, \alpha'\beta' = 0 \\
            & \beta'' + \frac{(R+r\cos \beta)\sin \beta}{r}\,(\alpha')^2 = 0
        \end{aligned}
        \right.
        \end{equation}
        Given a starting point $\p \in \Man$ and direction $\v \in \TpM$, we can compute the initial conditions, $\alpha(0), \beta(0), \alpha'(0), \beta'(0)$ and use \cref{eq:torus-geodesic} to compute $\alpha'', \beta''$ and integrate this differential equation to get $F(\alpha(||\v||),\beta(||\v||))$, the endpoint of the exponential map.

    \subsection{Additional Experiments on Our Method}

        \StartTightFigureParagraph
        \paragraph{Speed and accuracy comparison on more meshes} 
            To 
            {\parfillskip0pt\par}
            \begin{wrapfigure}[8]{R}[0pt]{0.4\linewidth}
                \centering
                \vspace{-8pt}
                \includegraphics[width=\linewidth]{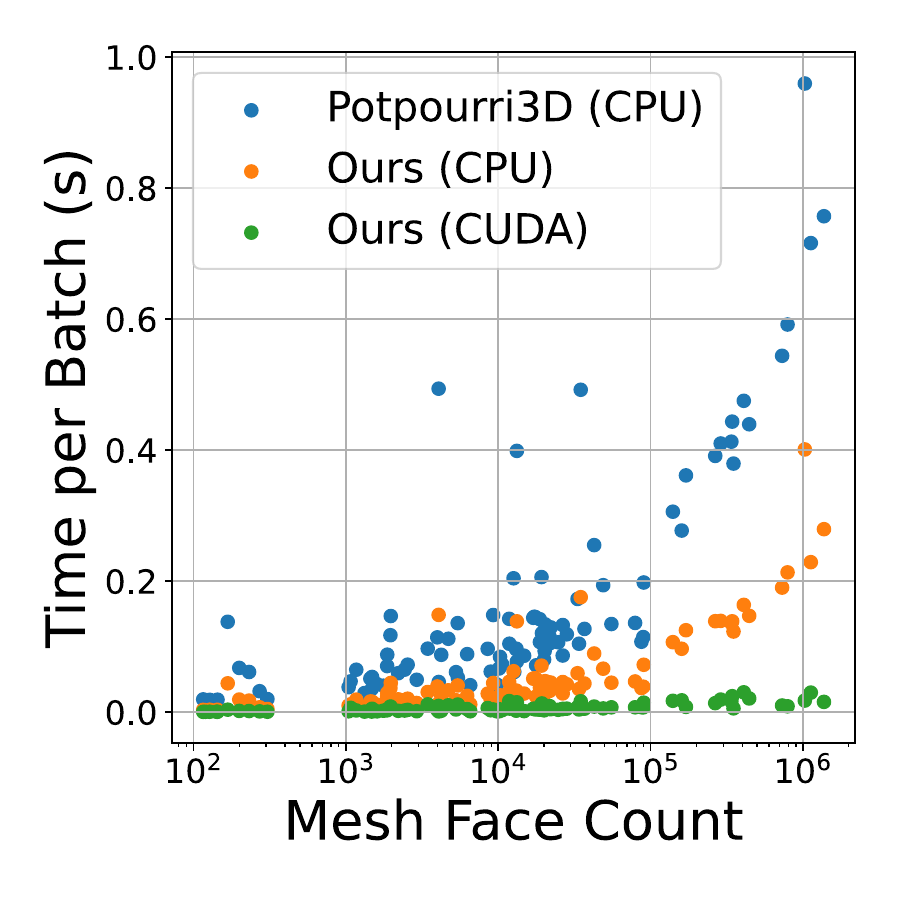}
            \end{wrapfigure}
            \noindent
            further corroborate our findings of \cref{fig:runtime} we evaluate our method also on $100$ meshes randomly sampled from the Thingi10K dataset~\cite{zhou2016thingi10k}. This plot reported as insert figure closely follows the trend and timings of \cref{fig:runtime}. Evaluating the accuracy against $\mathtt{pp3d}$ on the same set of $100$ meshes from Thingi10K, we report mean errors in the order of $10^{-6}$, thus closely aligning with \cref{tab:accuracy-combined}.

        \paragraph{Robustness to mesh quality}
            We assess the robustness to mesh quality for both our forward and backward steps, reporting results in \cref{fig:acc_corrupted_sphere,fig:cossim_corrupted_sphere}, respectively. Specifically, in \cref{fig:error_sphere_noise,fig:cossim_sphere_noise}, we progressively add noise to vertex positions; in \cref{fig:error_sphere_slim,fig:cossim_sphere_slim}, we deform approximately one third of the edges to produce slim triangles; in \cref{fig:error_sphere_holes,fig:cossim_sphere_holes}, we progressively remove triangles. The \emph{sparse level} on the x-axis of \cref{fig:error_sphere_holes,fig:cossim_sphere_holes} denotes the percentage of triangles removed. In this setting, $\mathtt{pp3d}$ fails to compute geodesics as some vertices belong to multiple boundary loops, whereas our method remains applicable. While in \cref{fig:error_sphere_holes} we report results for both the default and hole-avoidance (\cref{sec-supp:geostep}) versions of our method, gradients in \cref{fig:cossim_corrupted_sphere} are always evaluated using hole-avoidance.

        \EndTightFigureParagraph
        
\begin{figure}
                \vspace{-6pt}
                \centering
                \includegraphics[width=.8\linewidth]{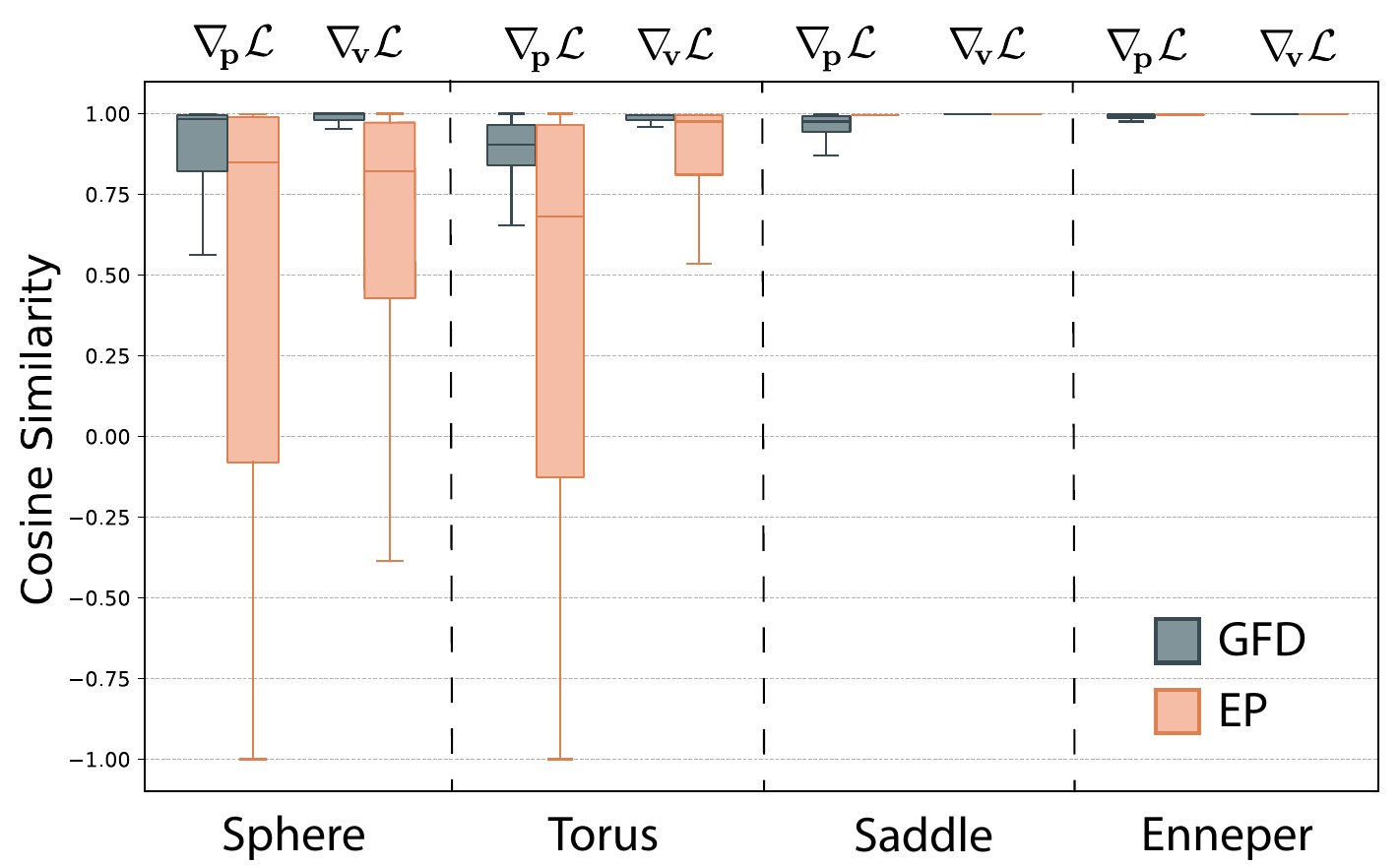}
                \caption{\label{fig:cossim_ode}Cosine similarity of the gradients using our method on different meshes}
                \vspace{-10pt}
            \end{figure}

        \paragraph{Gradient evaluation on more meshes}
            We conducted additional validations on multiple, positive and negative curvature surfaces,  as shown in \cref{fig:cossim_ode}. We address the lack of ground-truth (GT) gradients, by comparing against those from the ODEs using $\texttt{autograd}$ and compute the loss using an arbitrary point, $\mathcal{L}=  || \exp_\mathbf{p}(\mathbf{v})-\hat{\mathbf{p}}||^2$. While these numerical methods serve as a high-quality proxy, they remain approximations rather than exact GTs.
    
    \subsection{Additional Details on AGC Neural Network}
        \label{sec-supp:geonet}
        
\begin{figure*}
            \centering
            \includegraphics[width=0.9\linewidth]{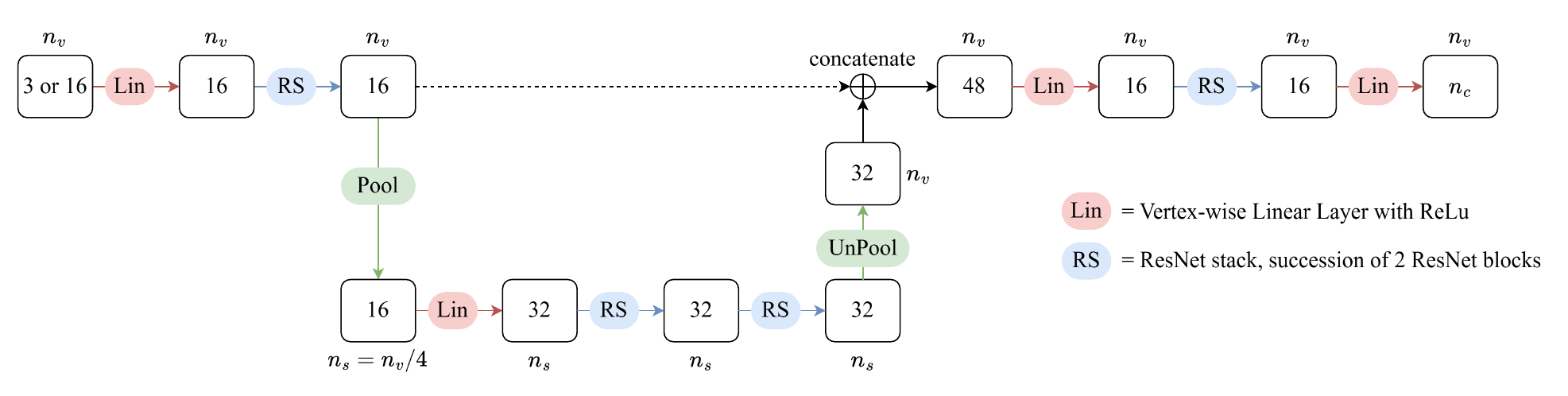}
            \vspace{-16pt}
            \caption{U-ResNet architecture used for segmentation. With $n_v$ the number of vertices, $n_s$ the subsampled vertices and $n_{c}$ the number of classes.}
            \label{fig:agcnn-unet}
            \vspace{-8pt}
        \end{figure*}

\begin{figure}
            \centering
            \includegraphics[width=\linewidth]{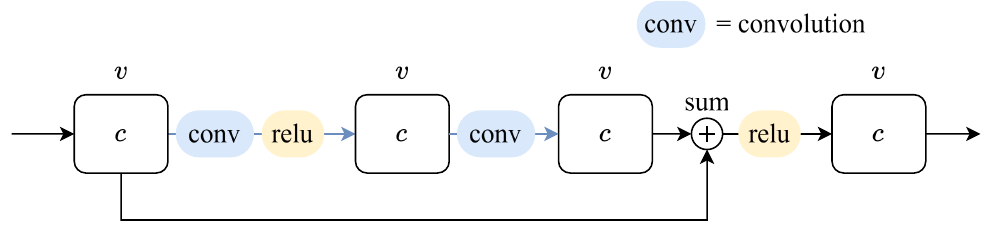}
            \caption{ResNet block architecture.}
            \label{fig:agcnn-resnetblock}
        \end{figure}

        We implemented our AGC Neural Network (AGCNN) in PyTorch~\cite{pytorch} using the same architecture as the U-ResNet from MDGCNN~\cite{mdgcnn}, which is based on Residual Networks~\cite{residual-net} and U-Net~\cite{u-net}. The architecture is depicted in \cref{fig:agcnn-unet}, where the ResNet stacks are made of two ResNet blocks, which are described on \cref{fig:agcnn-resnetblock}. For the pooling layer, we subsampled the meshes using quadric decimation~\cite{quadric-decimation} and targeted around a quarter of the faces of the mesh. Like in \cite{mdgcnn}, pooling and un-pooling are performed by matrix multiplying vertex features with sparse matrices obtained during quadric decimation, a common practice in many SotA methods~\cite{bouritsas2019spiralnet, gong2019spiralnet++, foti2022sdvae, foti2023ledvae}. We also doubled the number of filters used on the subsampled meshes (in the stacks 2 and 3).

        As mentioned in \cref{sec:geoconv}, our network takes as input a vector of real values for each vertex. Like in DiffusionNet~\cite{diffusionnet}, we considered two possible input types: the raw 3D coordinates of the vertices, and the Heat Kernel Signature (HKS)~\cite{hks}. When using 3D coordinates, we paired them with rotation and scaling augmentations, using a random rotation and a uniform scaling between $0.85$ and $1.15$. We also normalised the sizes of the meshes beforehand. With HKS as input, the network is invariant to any orientation-preserving isometric deformation of the shape, so rotation augmentation is not necessary. To compute the HKS, we used the same setup as DiffusionNet~\cite{diffusionnet}, with $128$ eigenvectors and heat kernel signatures sampled at $16$ temporal values logarithmically spaced on $[0.01, 1]$.
        
        We fit the model using Adam~\cite{adam} optimiser, and a cosine annealing learning rate scheduler, with an initial learning rate of $10^{-3}$ and an $\eta$ of $10^{-5}$. We used a cross-entropy loss with $0.1$ label smoothing. 
        We trained our model on a Nvidia A100, taking $1$ minute and $21$ seconds per epoch and using around $6.8$ GB of VRAM, while training on the full size meshes, which are made of around $10$k vertices. This means, for each mesh, we compute roughly $30$ million geodesic traces and still maintain a reasonable run time. We trained the model for $40$ epochs on hks, and $60$ epochs on xyz.

    \subsection{Additional Details on MeshFlow}
        \label{sec-supp:otexpflow}

        \begin{figure*}[ht]
                \centering
                \begin{subfigure}[t]{0.32\textwidth}
                    \centering
                    \includegraphics[width=\linewidth]{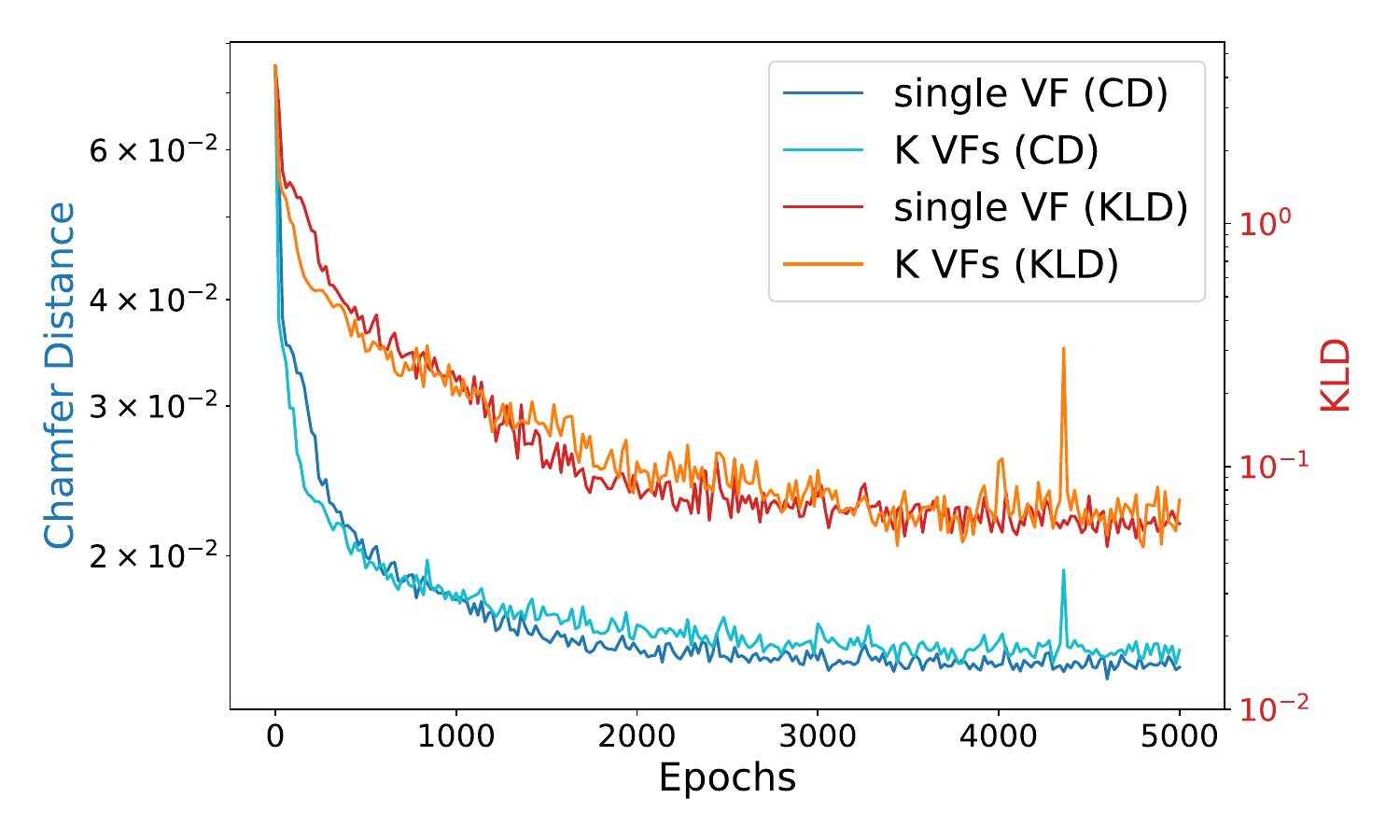}
                    \vspace{-10pt}
                    \caption{Comparison between learning a single vector field and $K$ vector fields.}
                    \label{fig:ot-exp-kmodels}
                \end{subfigure}
                \hfill
                \begin{subfigure}[t]{0.32\textwidth}
                    \centering
                    \includegraphics[width=\linewidth]{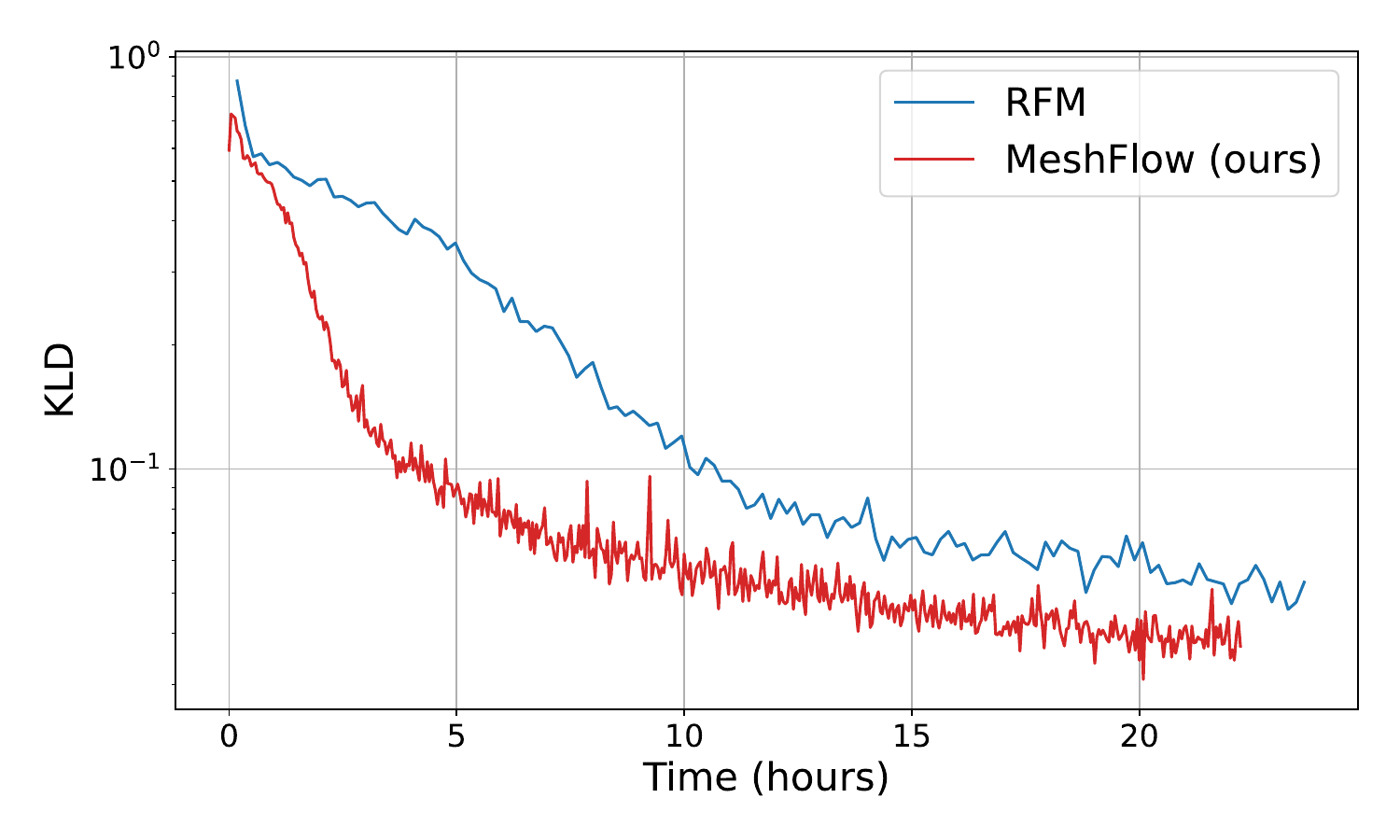}
                    \vspace{-10pt}
                    \caption{KLD over time when training on the Bunny with $k=100$.}
                    \label{fig:ot-rfm-kld}
                \end{subfigure}
                \hfill
                \begin{subfigure}[t]{0.32\textwidth}
                    \centering
                    \includegraphics[width=\linewidth]{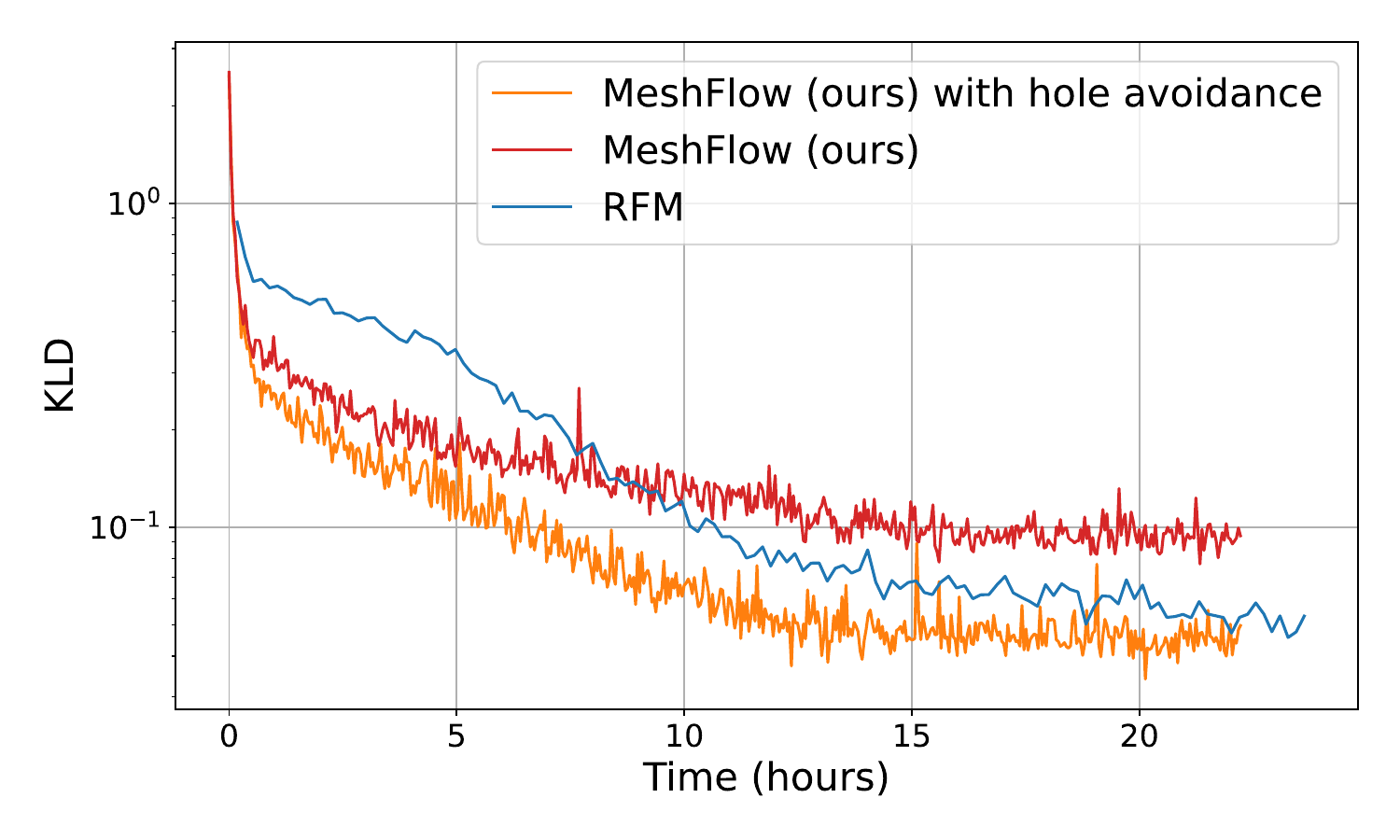}
                    \vspace{-10pt}
                    \caption{KLD over time when training on Spot with holes and $k=10$.}
                    \label{fig:ot-spot-holes}
                \end{subfigure}
                \caption{Comparison of different validation metrics during training.}
                \vspace{-10pt}
            \end{figure*}
            
        \paragraph{Differences and Similarities with RFM}
            As mentioned in \cref{sec:intro,sec:flow}, MeshFlow is inspired by flow matching, but there are multiple fundamental differences that we highlight in this section. The pseudocodes of our MeshFlow and RFM~\cite{riemannian-fm} are reported in \cref{alg:ot-exp,alg:rfm}, respectively. Note that \cref{alg:rfm} is adapted to adhere to the notation of this paper and reports only the case in which RFM is applied to meshes. Note also that $\p_i^{(1)}$ in \cref{alg:rfm} corresponds to our $\q_i$ in \cref{alg:ot-exp}.

            \input{algorithms/ot_exp}
            \input{algorithms/rfm}

            Both methods use a tangent vector field to transport points from a noisy source distribution to the target data distribution. The core differences between the two methods lie in: \emph{(i)} how points are geodetically transported on the surface of a mesh, \emph{(ii)} the choice of vector field, \emph{(iii)} the loss function, and \emph{(iv)} how points are paired. 
            
            MeshFlow leverages our differentiable exponential map to move points from source to target. Since our exponential map is based upon straightest geodesics, to  enable curved paths, we compute $K$ intermediate steps. On the other hand, RFM needs to solve the ODE corresponding to the gradient flow of the distance function, which is normalised to ensure constant speed along the trajectory. We leave the formal definition of the ODE to~\cite{riemannian-fm}. However, intuitively, the ODE dictates how to perform small steps along the trajectory. The movement direction is determined by the gradient of the biharmonic distance towards the target point. This Euler method performs Euclidean displacements, thus points need to be projected onto the mesh after every step. %

            Our method uses a time-invariant vector field $\v_\theta$, which is sampled at every step of the $K$-step exponential map. In contrast, RFM uses a time-conditioned vector field $\v_\theta^{(t)}$. During inference, this field is sampled for each small step of the Euler method. During training, it is optimised at multiple steps of the trajectory connecting source to target samples.  
            
            Since MeshFlow can move samples in a single pass and directly obtain the transported points, they can be directly compared against data samples using the biharmonic distance $d_{BH}$. RFM adopts a completely different strategy and computes the mean squared Euclidean distance between the tangent of the curve ($\dot{\p}^{(t_i)}_i$) at ${\p}^{(t_i)}_i$, and the vector field $\v_\theta^{(t_i)}$ at the arbitrarily selected time step $t_i$ and corresponding position $\p_i^{(t_i)}$.
            
            To find good matches, MeshFlow requires coupling noise ($\p_i \sim \mathcal{P}(\mesh)$) with target ($\q_i \sim \mathcal{Q}(\mesh)$) samples via optimal transport. RFM, on the other hand, can just select random couplings.

        \paragraph{Optimal Transport (OT) Couplings}
            The OT coupling is computed using the Hungarian algorithm~\cite{kuhn1955hungarian} and minimising the linear sum assignment of the pairwise squared Biharmonic distance:
            \begin{equation}
                \label{eq:coupling}
                \sigma = \underset{\hat{\sigma} \in \mathcal{S}}{\text{argmin}} \sum_{i=1}^{B_c} d_{BH}^2(\p_i,\q_{\hat{\sigma}(i)}),
            \end{equation}
            where $\sigma$ is the permutation of $\{1,...,B_c\}$ representing the OT coupling between the noise $\p_1,\dots,\p_{B_c}$ and target $\q_1,\dots,\q_{B_c}$ samples, and $\mathcal{S}$  the set of all possible permutations of the set $\{1,\dots,B_c\}$.
            
            Although we can use the same number of paired samples during training and while establishing the couplings, a higher number of samples is particularly beneficial for OT coupling. Therefore, we enable the selection of two different batch sizes: $B$ for training and $B_c$ for OT coupling, with $B \leq B_c$. In this case, the $K$-step exponential map, loss, and optimiser step in \cref{alg:ot-exp} are repeated $B_c/B$ times, to use all the previously OT-coupled pairs. 
        
        \paragraph{Implementation Details}
            We optimise MeshFlow using Adam~\cite{adam} with a learning rate of $3.10^{-5}$, for $5,000$ epochs. We use $20$k training samples and $5$k test samples per mesh.
            We report results for $B_c=1024$ and $B=256$ (MeshFlow-1024), as well as $B_c=B=256$ (MeshFlow-256) in \cref{tab:rfm-metrics}. Also, as previously reported in \cref{sec:flow}, 
            to learn the vector field, we use an MLP with 3 hidden layers of size $512$, and $K=5$. The MLP takes as input the 3D position of the samples and outputs the corresponding vector. 
            We use the GFD differentiation scheme as we need to back-propagate also through positions.

            We also experimented with learning \(K\) separate vector fields (one per step), and report the corresponding metrics in \cref{fig:ot-exp-kmodels}. However, this approach did not yield improved performance and required longer training time.
            
        \paragraph{Detailed Performance Comparison with RFM}
            We benchmark MeshFlow against RFM on an Nvidia A100 GPU, and use the same batch size, training samples, and test samples. For our method, using an OT batch size of $1024$, each epoch took around $8$ seconds and used $1.6$ GB of VRAM, and was trained for $5,000$ epochs, or around $11$ hours. For RFM, each epoch took around $162$ seconds, and used $38$ GB of VRAM, and was trained for around $400$ epochs, or around $18$ hours. In \cref{fig:ot-rfm-kld} we compare the convergence speed of MeshFlow and RFM when trained for approximately the same time. Our method not only converges significantly faster (\cref{fig:ot-rfm-kld}), but also has significantly faster inference ($98$s for RFM vs $5.8\cdot10^{-3}$s for ours), as it requires less steps and does not rely on projections to stay on the manifold. 
            
            While already significantly faster than RFM, the main bottleneck of MeshFlow is computing the OT coupling. The Hungarian algorithm we currently use runs on the CPU and, with a complexity of $\mathcal{O}(B_c^3)$, it does not scale well with larger batch sizes. Nevertheless, this problem can be efficiently addressed using faster coupling techniques, such as the Sinkhorn algorithm~\cite{sinkhorn1964relationship, feydy2019interpolating}.
        
        \paragraph{MeshFlow with Hole Avoidance}
            We tested our method with and without the hole avoidance mechanism introduced in \cref{sec:dpsg} and further detailed in \cref{sec-supp:geostep}, where the algorithm attempts to parallel transport the direction of the geodesic along the hole, and prevents the trace from stopping abruptly. As observed on \cref{fig:ot-spot-holes}, while RFM proved to be robust to the presence of holes, our method still outperforms RFM when using the hole avoidance mechanism. The gap in performance between our method with and without hole avoidance justifies the introduction of this key modification to the straightest geodesics algorithm.

    \subsection{Additional Details on Mesh-LBFGS and GCVT}
        \label{sec-supp:mlbfgs}

        \begin{figure}[t]
            \includegraphics[width=\linewidth]{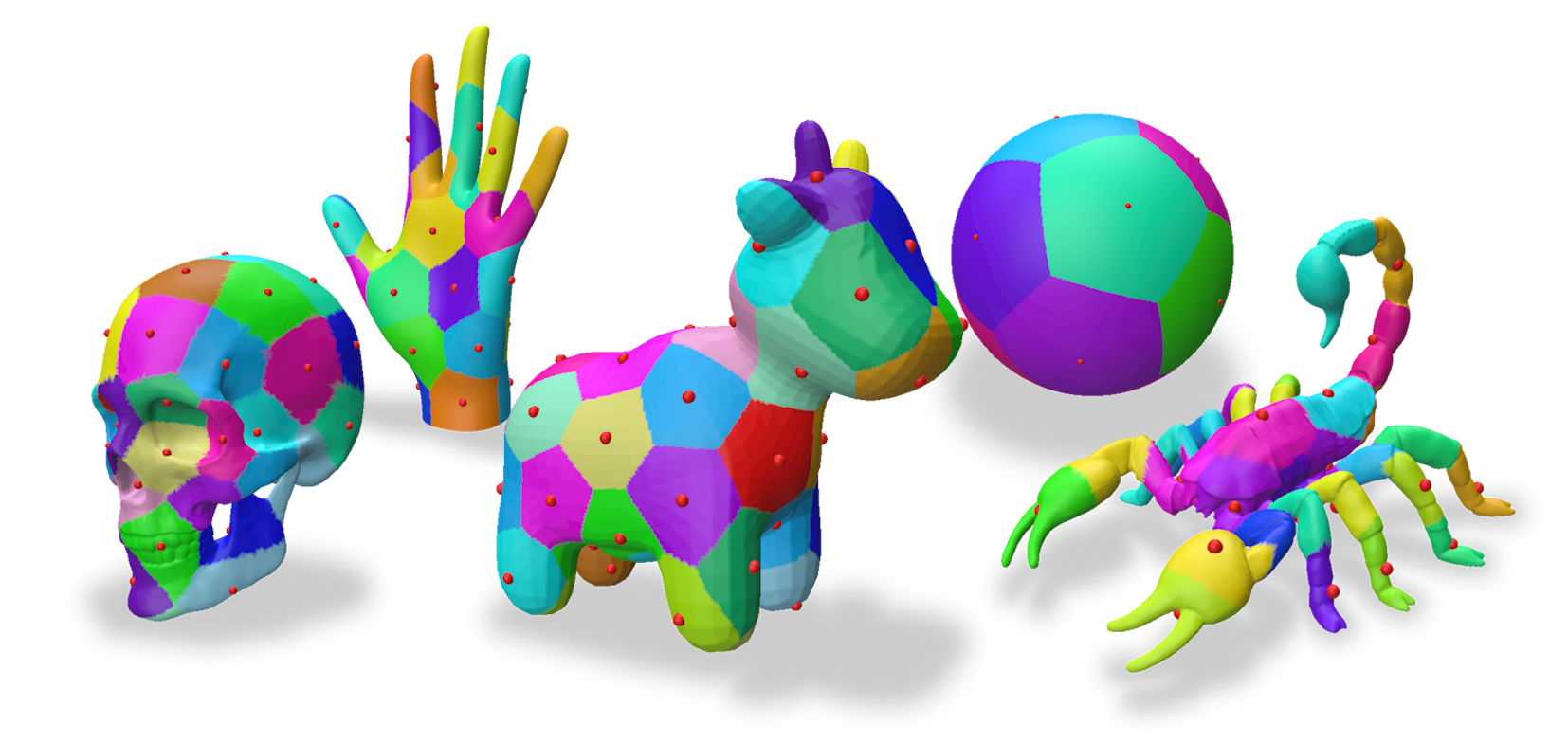}
            \caption{Examples of GCVTs using our method on Skull, Hand, Spot, Ball and Scorpion.}
            \label{fig:gcvt-examples}
        \end{figure}

        The Geodesic Centroidal Voronoi Tessellation (GCVT) on a mesh $\mesh$ aims to move the seeds $\mathbf{S} =\{\mathbf{s}_i \}_{i=1}^S \in \mesh^S$ to the centre of mass of their respective Voronoi cells $\Omega_i$. Centroids are defined as the Karcher mean of the Voronoi cells~\cite{vectorheat}.
        
        Since we are operating on triangular meshes, we discretise the computation of the Karcher mean to make it tractable. We compute the vertex area $A(\x)$ $\forall \x \in \Vertices\in \mesh$, which is defined as a third of the sum of the areas of its connected triangles. We then use the algorithm proposed by \cite{vectorheat} to compute the Karcher mean while also leveraging their heat-based logarithm map formulation. In practice, we only take a single step of the Karcher mean algorithm.
        We define the update vector $\v_i$ pointing towards the centroid of region $\Omega_i$ as:
        \begin{equation*}
            \v_i = \frac{\sum_{\Omega_i} A(\x)\rho(\x) \log_{\mathbf{s}_i}(\x)}{\sum_{\Omega_i} A(\x)\rho(\x)},    
        \end{equation*}
        where $\rho(\x) = \frac{k_t(\mathbf{s}_i, \cdot)}{\sum_j k_t(\mathbf{s}_j, \cdot)}$ is the density function defined at the vertices of the mesh via the scalar heat kernel $k_t$.
        Using the standard Lloyd's algorithm, seeds are then updated via $\mathbf{s}'_{i} = \Exp_{\mathbf{s}_i}(\v_i)$.
        This is equivalent to minimising the following energy function:
        \begin{equation*}
            E(\mathbf{S}) = \frac{1}{2S} \sum_{i=1}^S \sum_{\Omega_i} A(\x) \rho(\x) ||\log_{\mathbf{s}_i}(\x)||^2. 
        \end{equation*}        
        
        To extend this method to a Riemannian second-order optimiser (our Mesh-LBFGS), we leverage the fact that $\mathbf{v}_i$ represents the steepest descent direction. Thus, we approximate the gradient of the energy with the opposite of the directions computed via the Karcher mean, i.e., $\nabla E(\mathbf{S}) \approx (-\v_0, \dots, -\v_n)$. The vector transport $\Pi$ needed for the optimiser is computed via our $\Exp$ map when computing $\mathbf{s}_{i+1}$.
        Its adjoint is simply the inverse of this parallel transport, where a vector will be moved backwards on the curve.
        
        We then apply \cref{alg:mlbfgs} to optimise the seeds. In the general Riemannian L-BFGS framework, a Riemannian metric is required; in our case, however, the mesh is embedded in $\mathbb{R}^3$, so the metric is the one induced by the ambient Euclidean space, $g_{\p}(\u, \v) = \langle \u, \v \rangle$.

        Since we are moving a batch of seeds simultaneously, we operate on the product manifold $\mesh^S$, and treat the collection $\mathbf{S}$ as a single state. The metric, $\Exp$ and vector transport are easily derived on this product manifold, for $\mathbf{S},\mathbf{S}'\in\mesh^S$ and $\mathbf{V},\mathbf{V}'\in\mathcal{T}_\mathbf{S}\mesh^S$:
        \[
        \begin{aligned}
            g_\mathbf{S}(\mathbf{V},\mathbf{V}')
            &=\langle\mathbf{V},\mathbf{V}'\rangle=\langle\v_1,\v'_1\rangle + ... + \langle\v_S,\v'_S\rangle \\
            \Exp_{\mathbf{S}}(\mathbf{V})
            &=(\Exp_{\mathbf{s}_1}(\v_1),...,\Exp_{\mathbf{s}_S}(\v_S)) \\
            \Pi_\mathbf{S}^{\mathbf{S}'}\mathbf{V}
            &=(\Pi_{\mathbf{s}_1}^{\mathbf{s}_1'}\mathbf{v}_1,...,\Pi_{\mathbf{s}_S}^{\mathbf{s}'_S}\mathbf{v}_S)
        \end{aligned}
        \]

        \input{algorithms/mlbfgs}

        We set the base learning rate $\alpha_0$ to $1$ for Lloyd’s algorithm and $0.5$ for Mesh-LBFGS. While the smaller rate slows the initial steps, it yields greater stability and faster overall convergence, outperforming Lloyd’s algorithm after a few iterations.

\begin{figure}[t]
        \centering
    
        \begin{subfigure}{\linewidth}
            \centering
            \begin{subfigure}{0.49\linewidth}
                \centering
                \includegraphics[width=\linewidth]{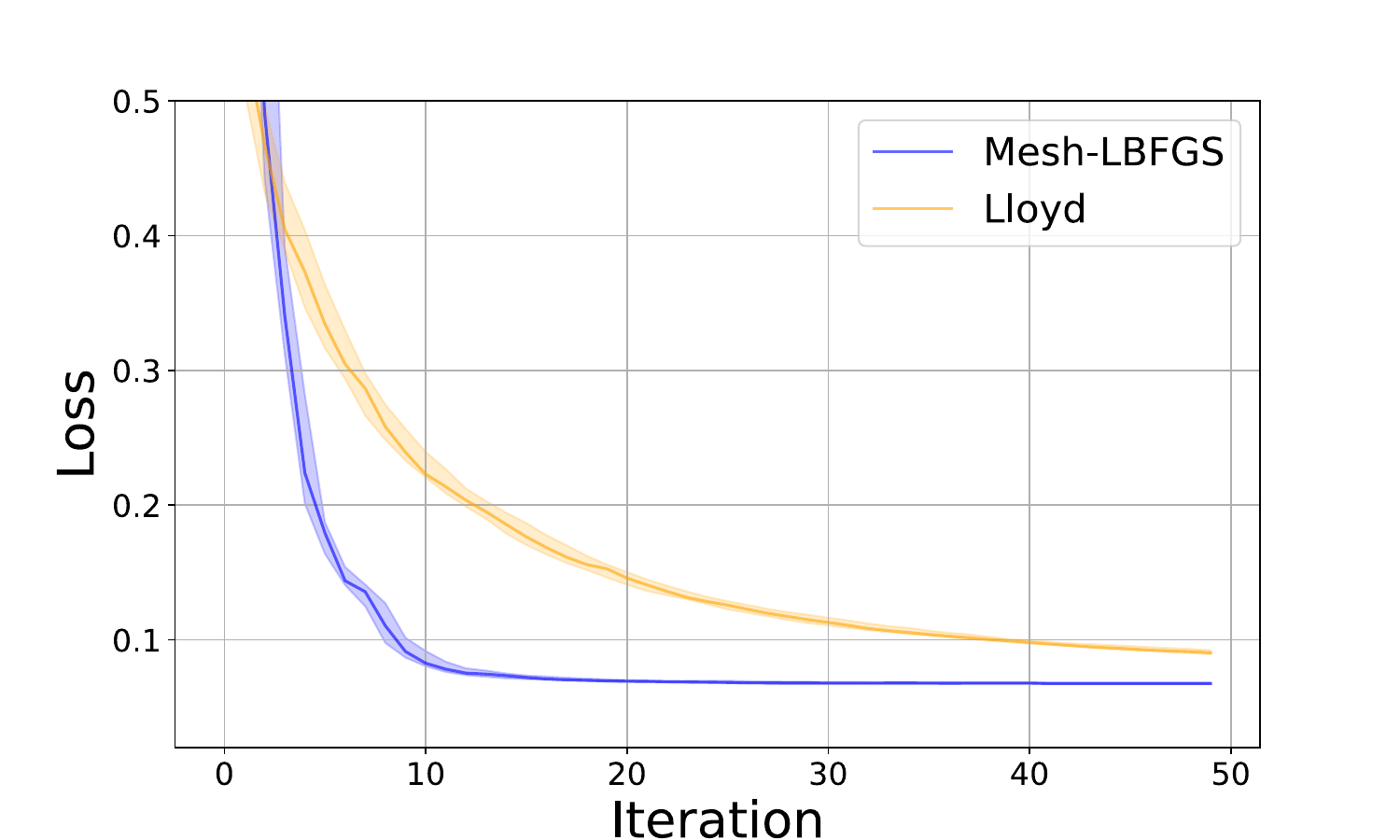}
            \end{subfigure}%
            \hfill
            \begin{subfigure}{0.49\linewidth}
                \centering
                \includegraphics[width=\linewidth]{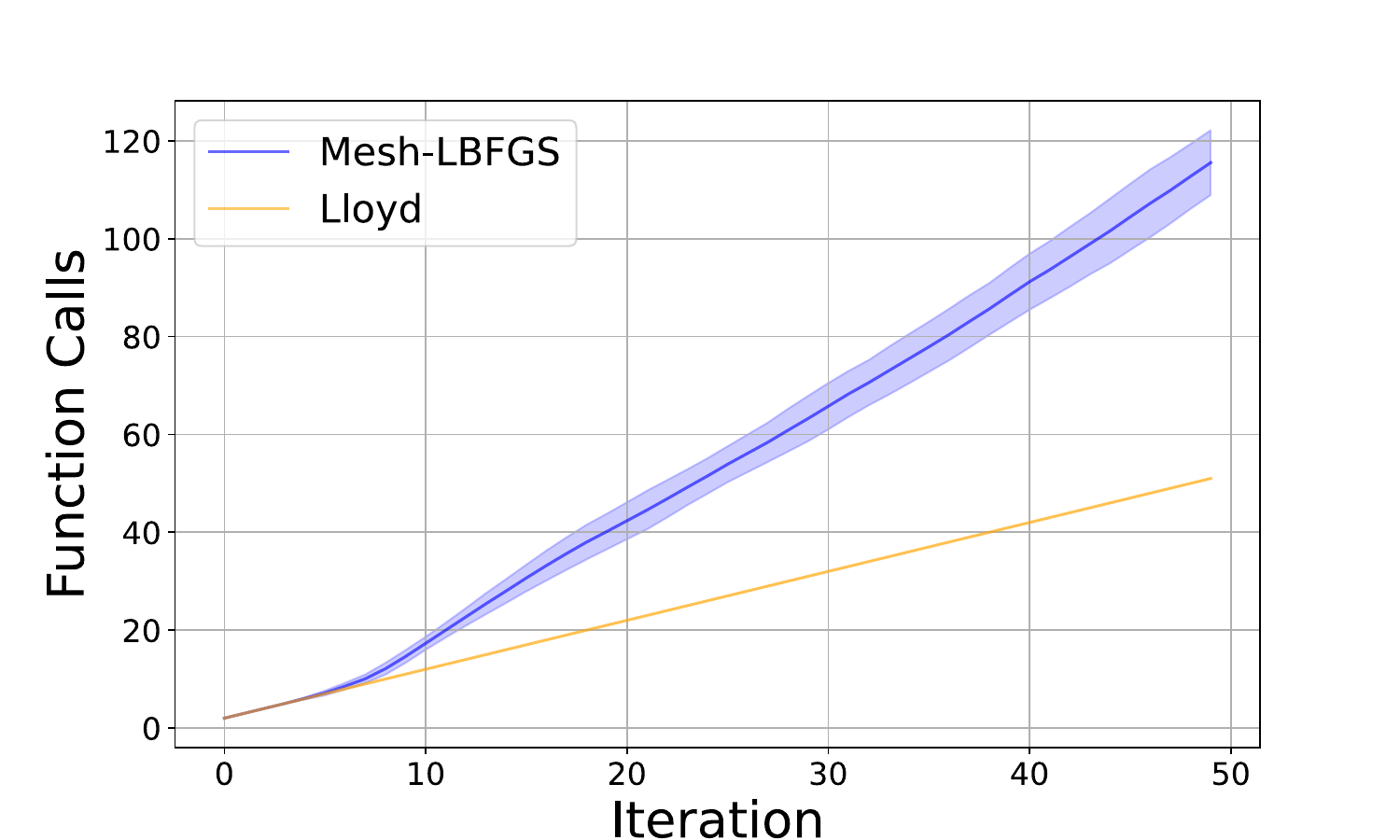}
            \end{subfigure}
            \subcaption{On Spot mesh}
        \end{subfigure}
    
        \vspace{0.5em}
    
        \begin{subfigure}{\linewidth}
            \centering
            \begin{subfigure}{0.49\linewidth}
                \centering
                \includegraphics[width=\linewidth]{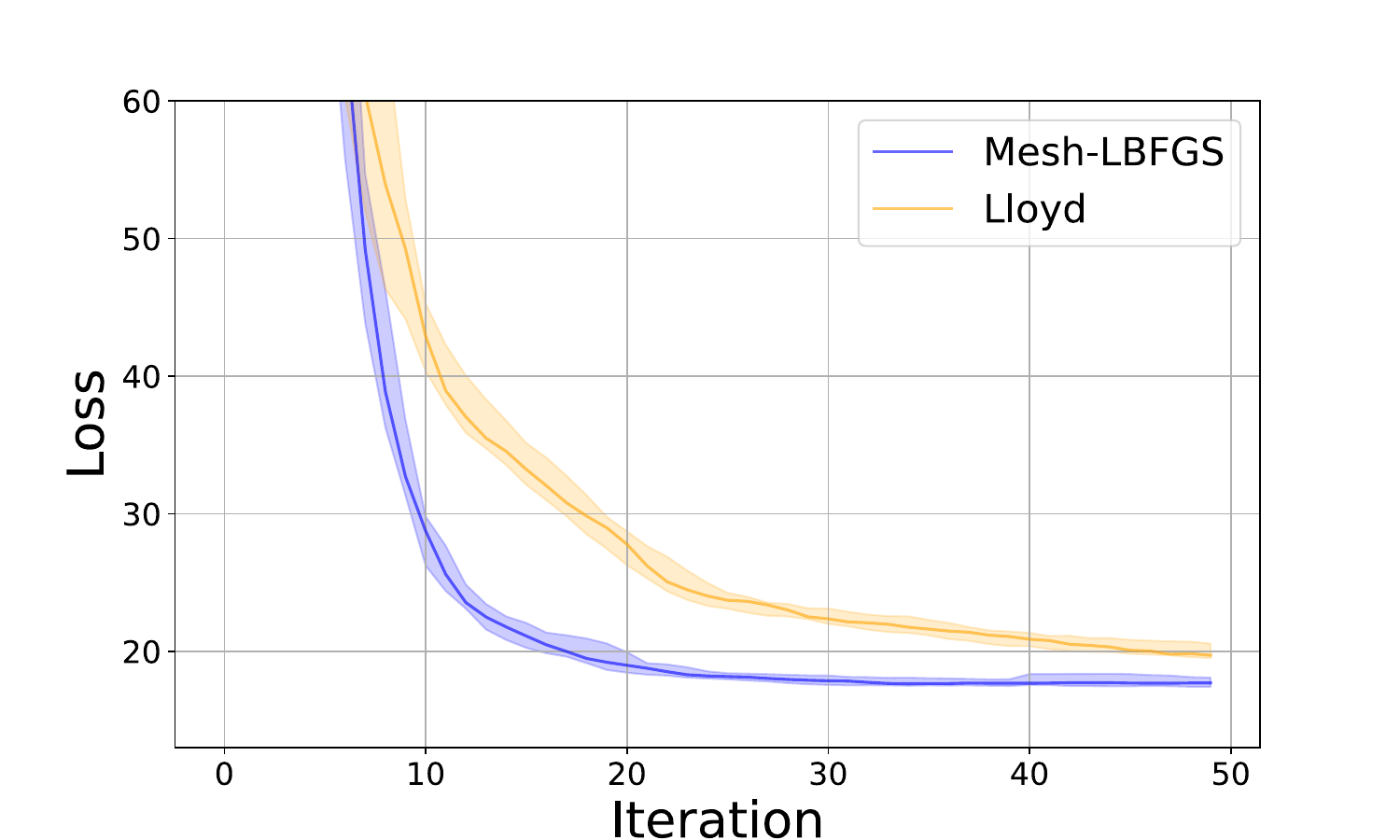}
            \end{subfigure}%
            \hfill
            \begin{subfigure}{0.49\linewidth}
                \centering
                \includegraphics[width=\linewidth]{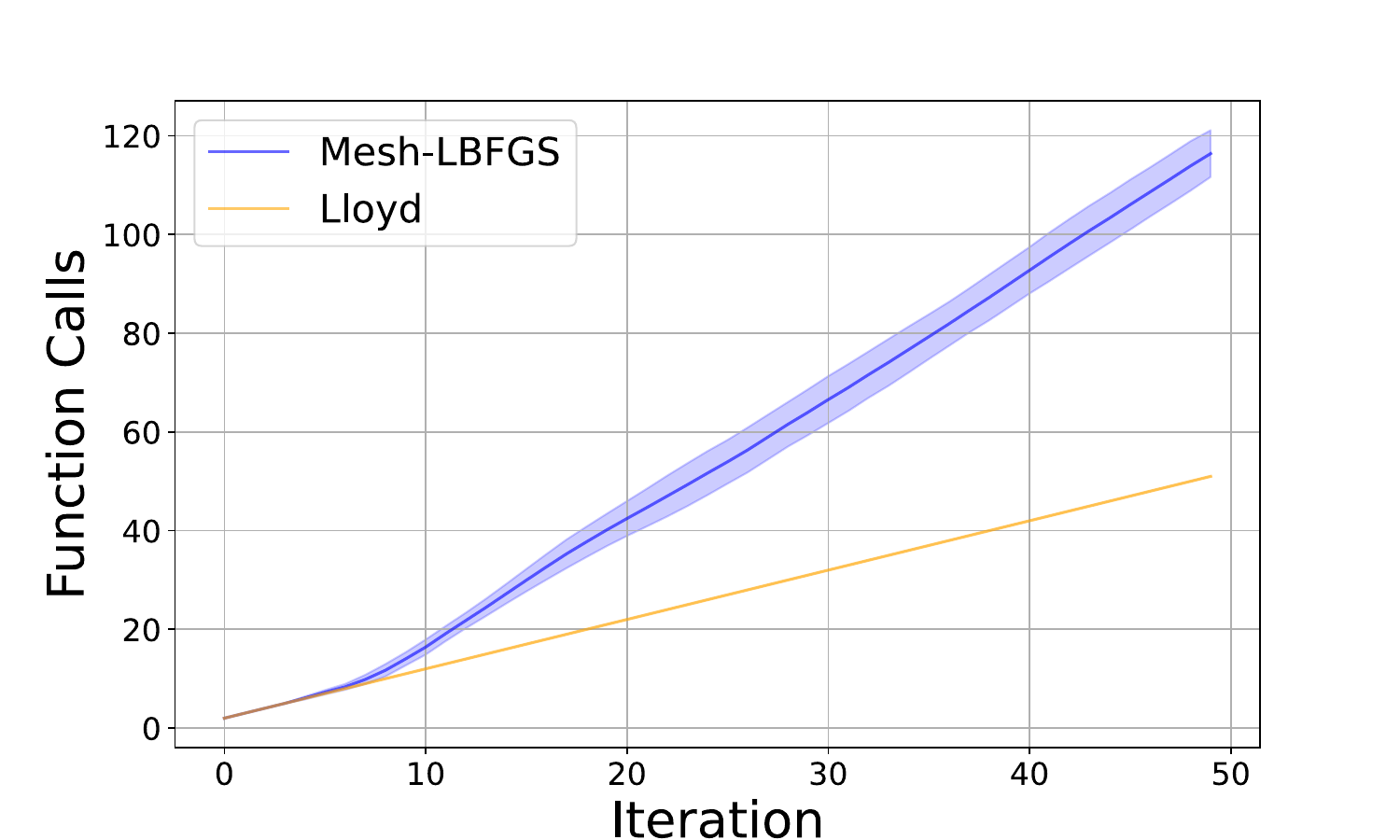}
            \end{subfigure}
            \subcaption{On Scorpion mesh}
        \end{subfigure}
    
        \vspace{0.5em}
    
        \begin{subfigure}{\linewidth}
            \centering
            \begin{subfigure}{0.49\linewidth}
                \centering
                \includegraphics[width=\linewidth]{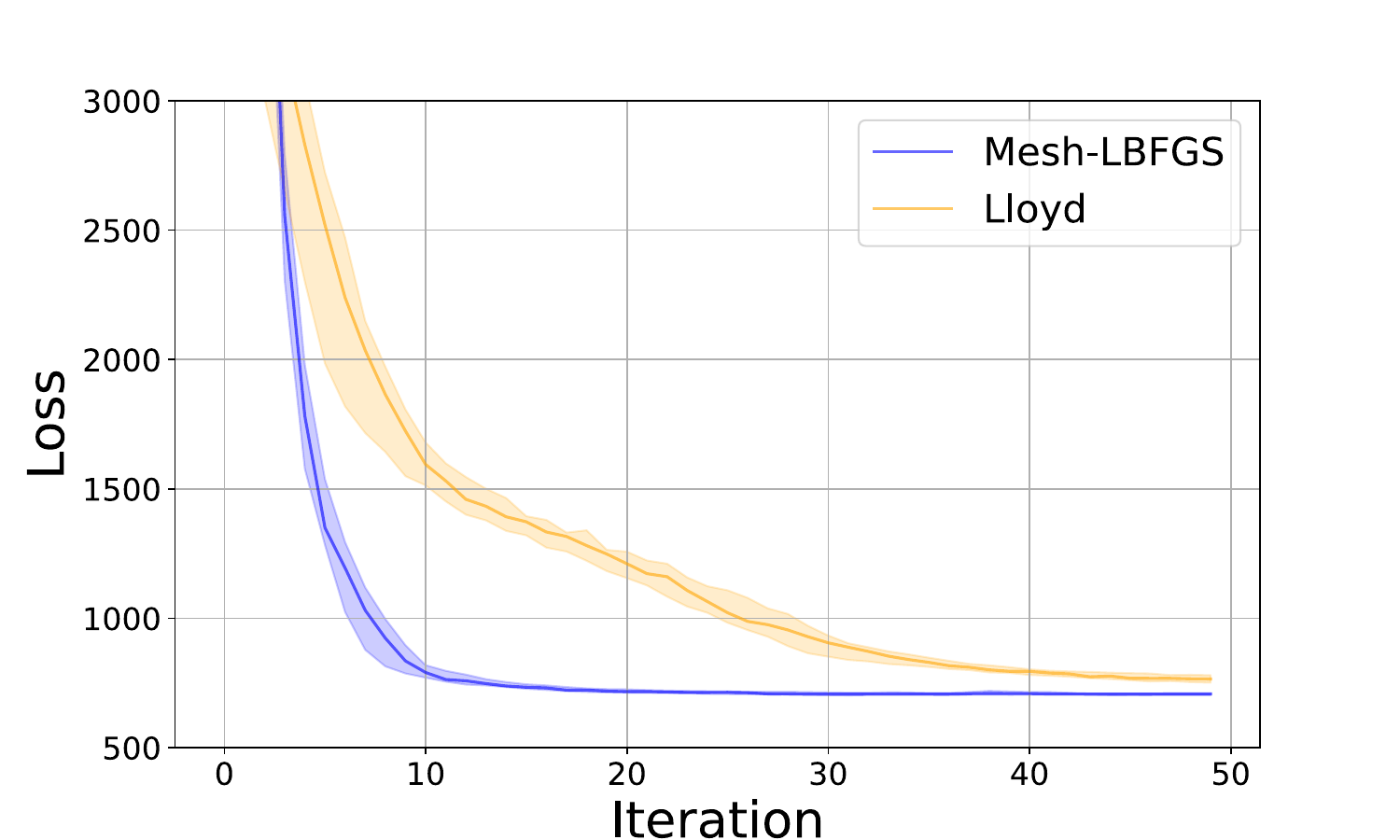}
            \end{subfigure}%
            \hfill
            \begin{subfigure}{0.49\linewidth}
                \centering
                \includegraphics[width=\linewidth]{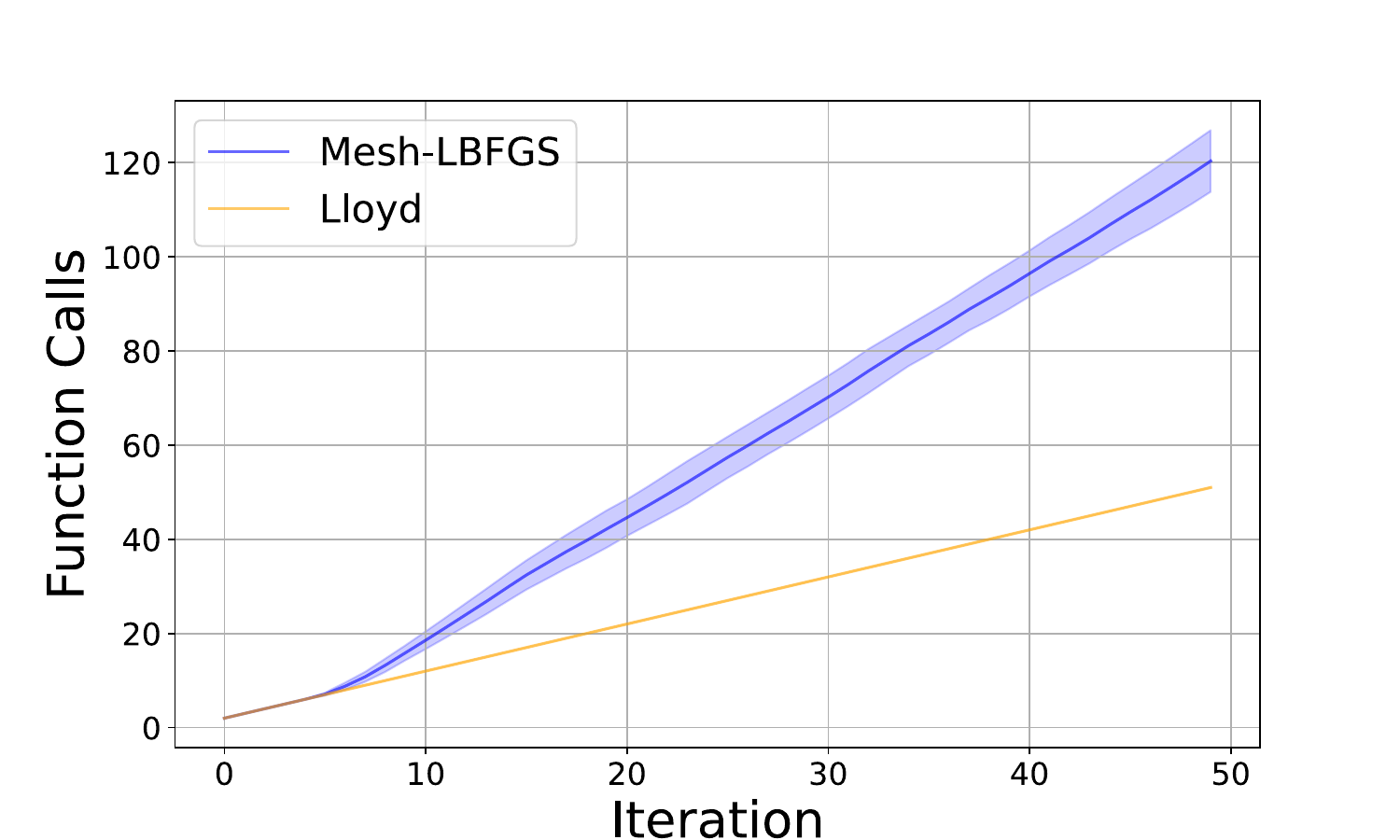}
            \end{subfigure}
            \subcaption{On Skull mesh}
        \end{subfigure}
    
        \vspace{-2pt}
        \caption{Comparison of different GCVT optimisers for 50 \textbf{clustered} seeds over 20 runs on different meshes.
        \textit{Left}: Median loss functions.
        \textit{Right}: Median total function calls.}
        \label{fig:gcvt-cluster-comparison}
    \end{figure}

\begin{figure}[t]
        \centering
    
        \begin{subfigure}{\linewidth}
            \centering
            \begin{subfigure}{0.49\linewidth}
                \centering
                \includegraphics[width=\linewidth]{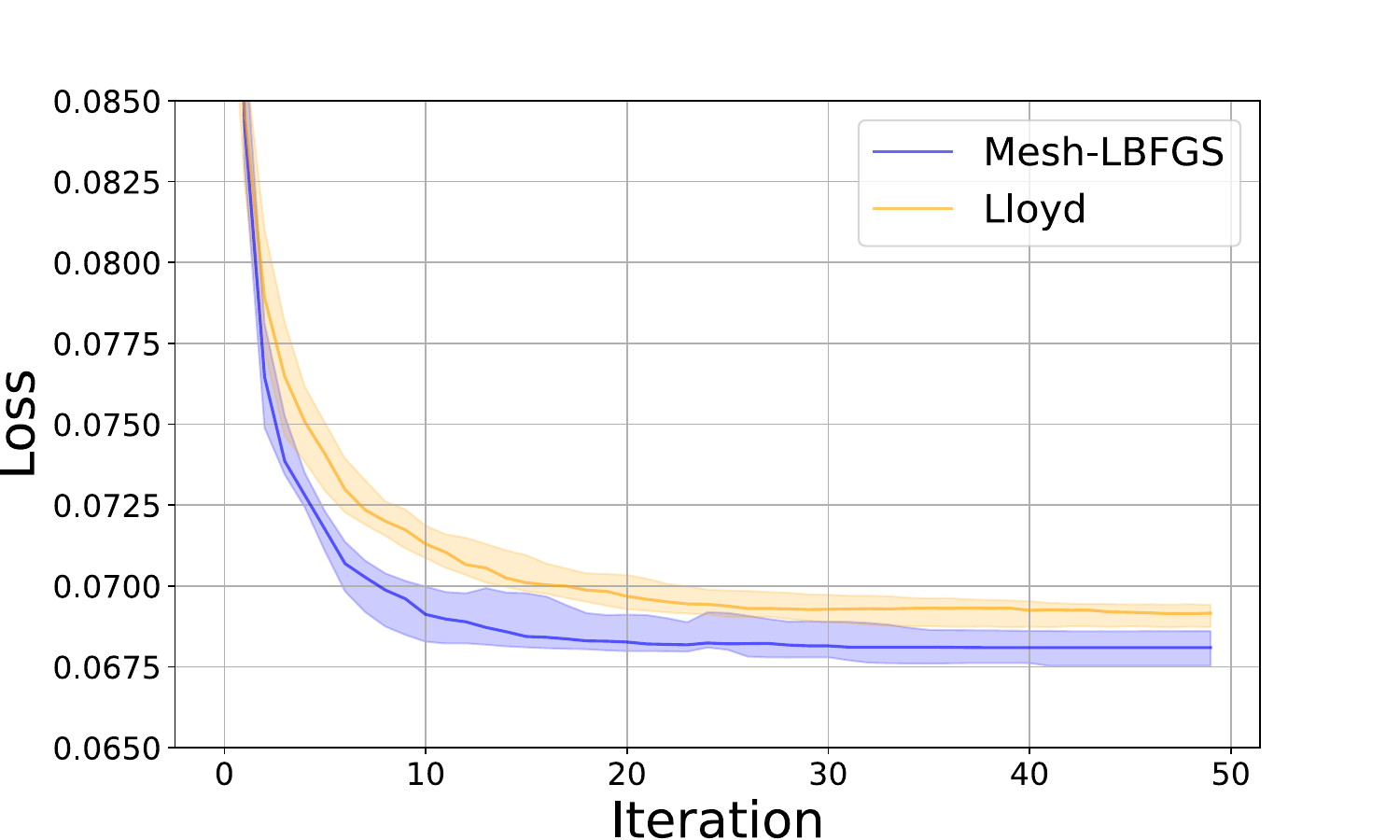}
            \end{subfigure}%
            \hfill
            \begin{subfigure}{0.49\linewidth}
                \centering
                \includegraphics[width=\linewidth]{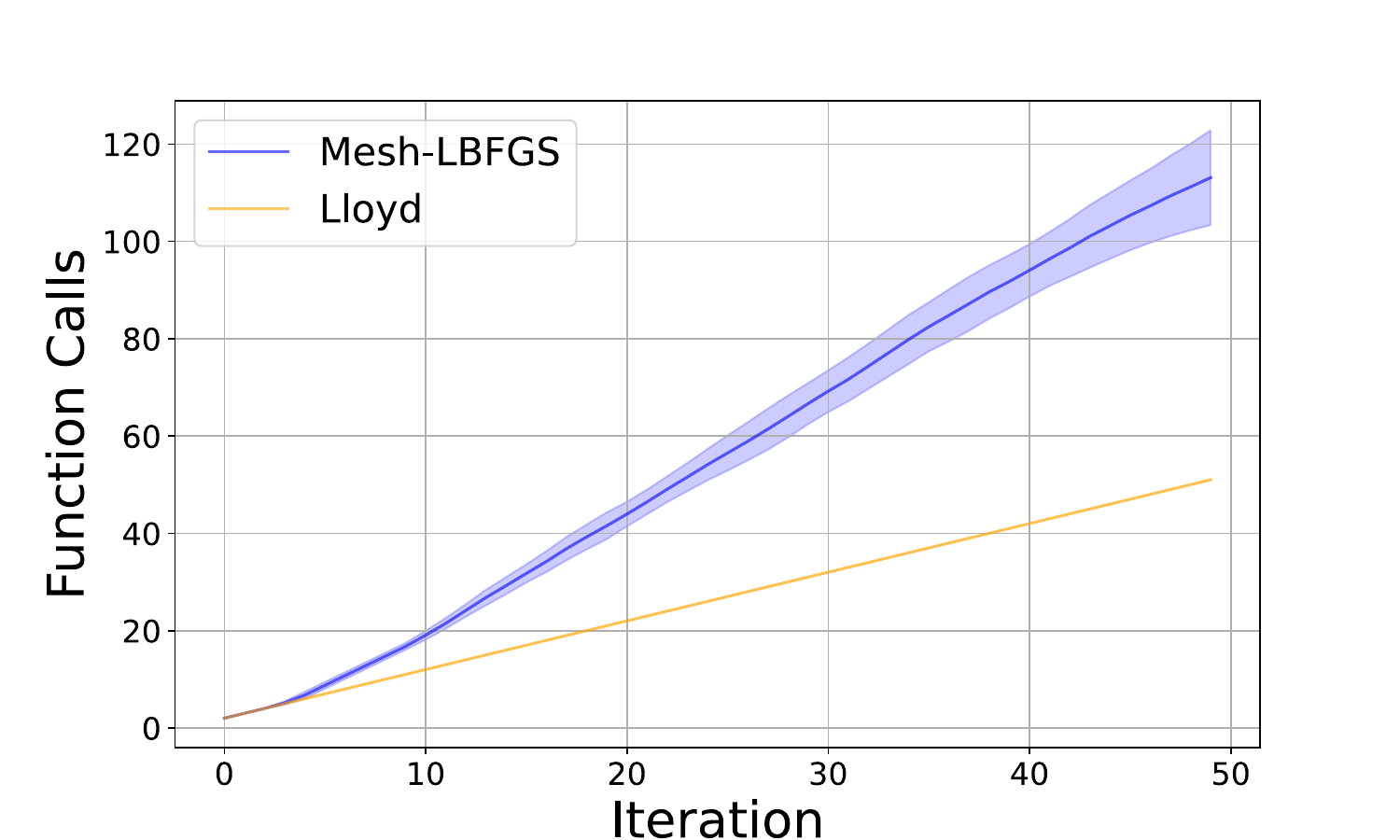}
            \end{subfigure}
            \subcaption{On Spot mesh}
        \end{subfigure}
    
        \vspace{0.5em}
    
        \begin{subfigure}{\linewidth}
            \centering
            \begin{subfigure}{0.49\linewidth}
                \centering
                \includegraphics[width=\linewidth]{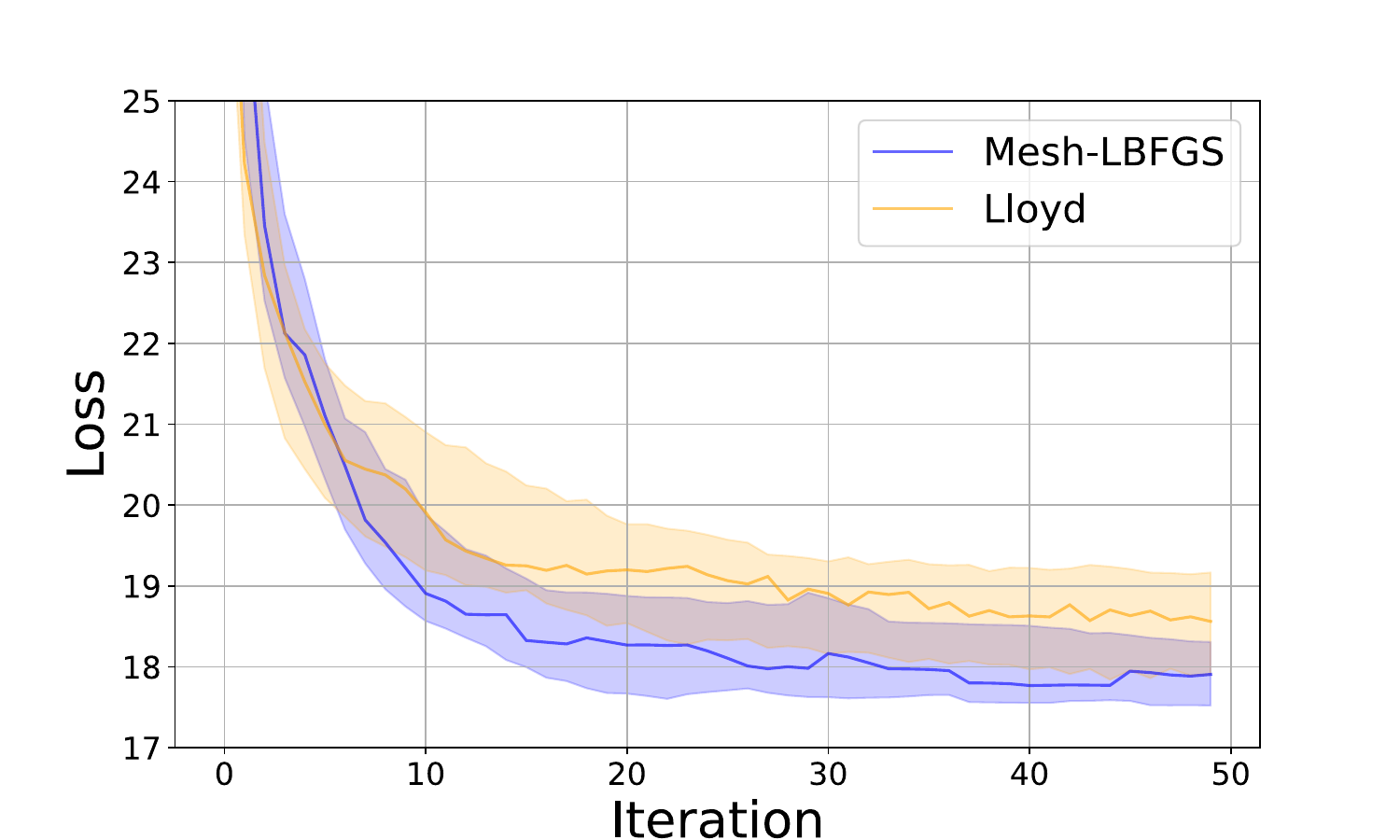}
            \end{subfigure}%
            \hfill
            \begin{subfigure}{0.49\linewidth}
                \centering
                \includegraphics[width=\linewidth]{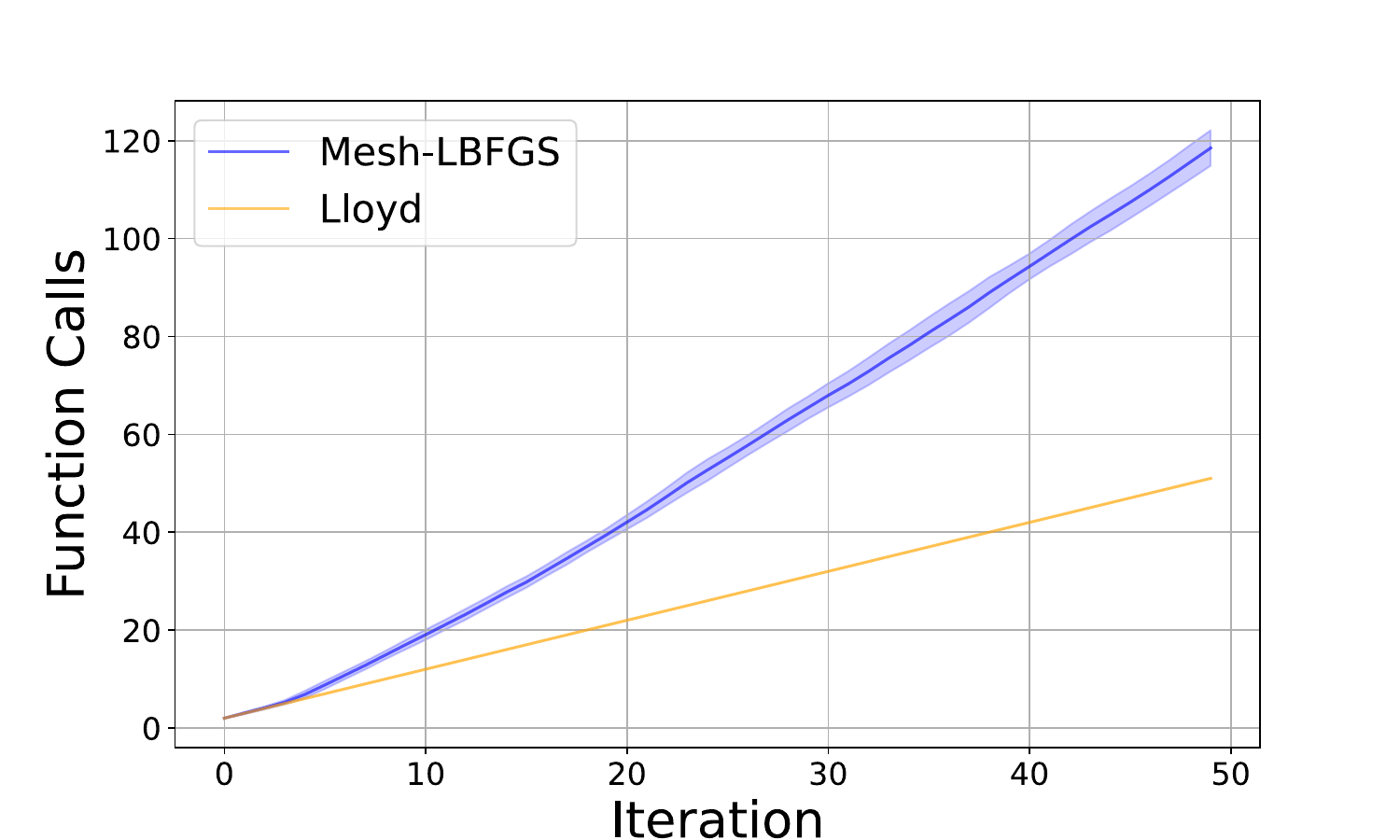}
            \end{subfigure}
            \subcaption{On Scorpion mesh}
        \end{subfigure}
    
        \vspace{0.5em}
    
        \begin{subfigure}{\linewidth}
            \centering
            \begin{subfigure}{0.49\linewidth}
                \centering
                \includegraphics[width=\linewidth]{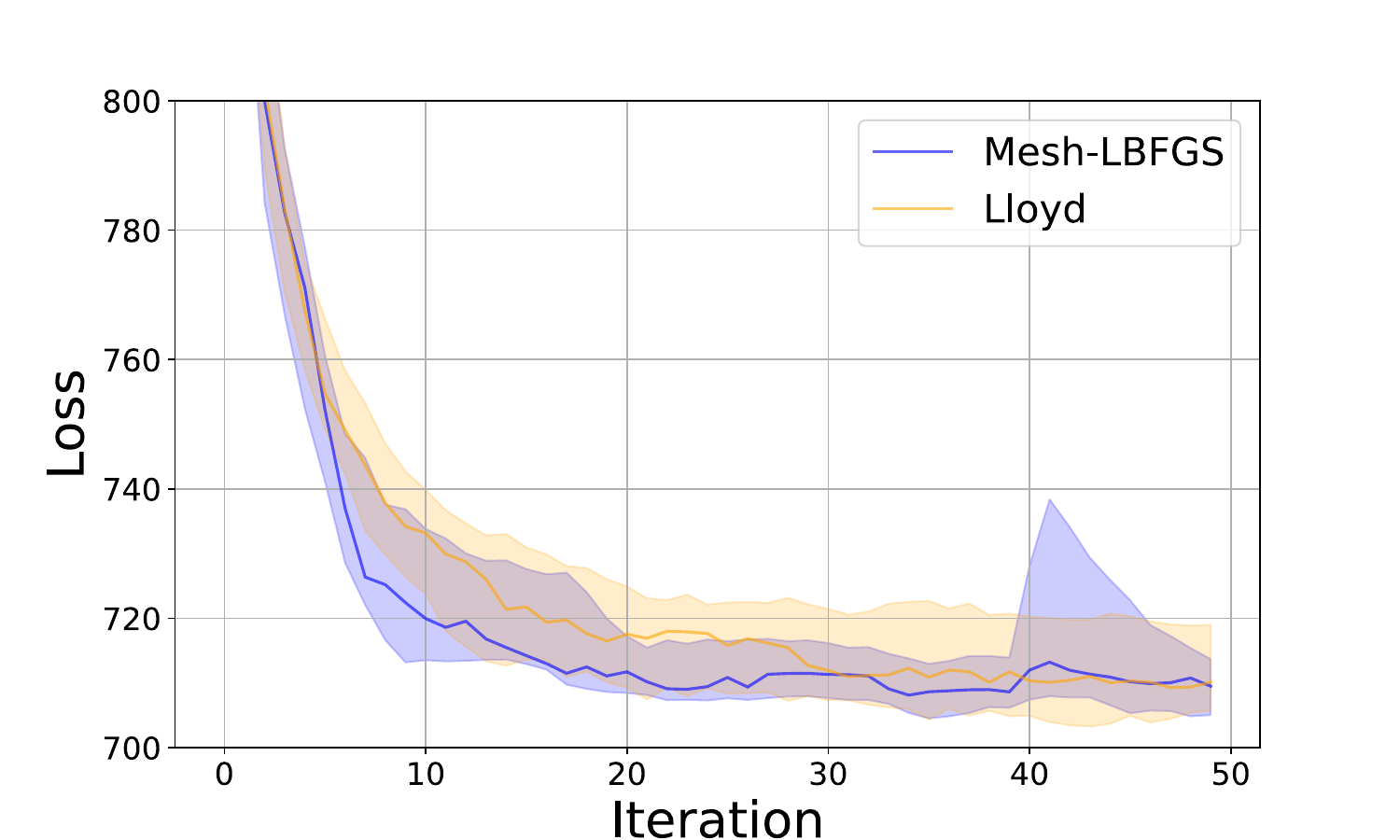}
            \end{subfigure}%
            \hfill
            \begin{subfigure}{0.49\linewidth}
                \centering
                \includegraphics[width=\linewidth]{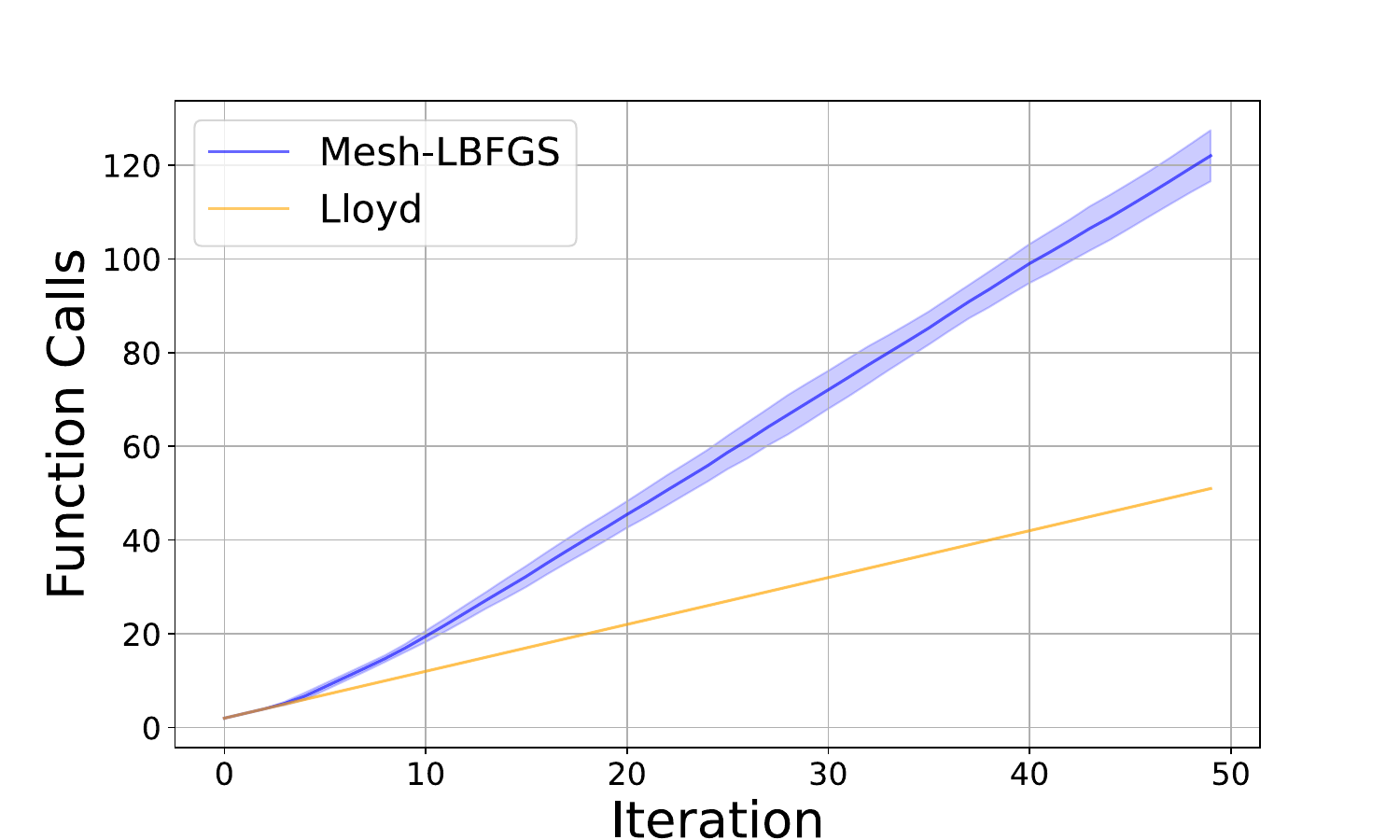}
            \end{subfigure}
            \subcaption{On Skull mesh}
        \end{subfigure}
    
        \vspace{-2pt}
        \caption{Comparison of different GCVT optimisers for 50 \textbf{uniform} seeds over 20 runs on different meshes.
        \textit{Left}: Median loss functions.
        \textit{Right}: Median total function calls.}
        \label{fig:gcvt-uniform-comparison}
    \end{figure}

        We compare our optimiser with Lloyd's algorithm on different meshes with different initial sampling methods, and compare both the loss function and total function calls. In \cref{fig:gcvt-cluster-comparison} we use random clustered seeds, and in \cref{fig:gcvt-uniform-comparison} we use random uniform seeds. We resample the seeds at each run and plot the medians with the 25\% and 75\% quartiles. Overall, our method converges in fewer iterations than Lloyd’s algorithm, but at the same time, it also incurs more function calls. Nevertheless, because it reaches convergence much faster overall, the total number of function calls is ultimately lower, especially when using clustered seeds.

%% file: algorithms/geodesic_step.tex
\begin{algorithm}[b]
    \caption{$\mathtt{geodesic\_step}$}
    \label{alg:sg-step}
    \begin{algorithmic}[0]
    \Require mesh $\mesh$, surface point $\p$, direction $\v$
    \State \hrulefill
    \State $(\faceindex, \bary) \gets \p$
    \If{$\exists k \text{ such that } \bary_k = 1$}
        \State \Return $\mathtt{transport\_over\_vertex}(\mesh, \faceindex,\bary,\v)$
    \ElsIf{$\exists k \text{ such that } \bary_k = 0$}
        \State \Return $\mathtt{transport\_over\_edge}(\mesh, \faceindex,\bary,\v)$
    \Else
        \State Let $\bary^\v$ the barycentric representation of $\v$ in $\faceindex$
        \State $\lambda \gets \min \left\{ -\frac{\bary_i}{\bary^\v_i} \;\middle|\; i \in [0,2],\; -\frac{\bary_i}{\bary^\v_i} > 0 \right\}$
        \State $\bary \gets \bary +\lambda \bary^\v$
        \State $\p \gets (\faceindex, \bary)$
        \State $\Delta \v \gets \lambda \left(\bary^\v_v (\x_1-\x_0) + \bary^\v_w (\x_2-\x_0) \right)$
        \State \Return $\p,\v, ||\Delta \v||$
    \EndIf
    \end{algorithmic}
\end{algorithm}

%% file: algorithms/transport_vertex.tex
\begin{algorithm}
    \caption{$\mathtt{transport\_over\_vertex}$. (we highlight in \textcolor{mypink}{pink} the non differentiable steps and in \textcolor{myteal}{blue} the hole avoidance modifications). }
    \label{alg:sg-step-vertex}
    \begin{algorithmic}[0]
    \Require mesh $\mesh$, face $\faceindex$, barycentric coordinates $\bary$, direction $\v$
    \State \hrulefill

    \State $(\Vertices, \Faces) \gets \mesh$
    \State $\vertexindex_0 \gets \Faces[\faceindex, \text{argmax}_k \bary_k]$

    \If{$\p$ is adjacent to a hole}
        \State \textcolor{myteal}{$\faceindex' \gets \text{argmin}_f \,\mathtt{transport\_error}(\mesh,\v,f)$}
        \State \textcolor{myteal}{$\bary' \gets (0,0,0)$}
        \State \textcolor{myteal}{$k \gets \text{argwhere}_i \Faces[\faceindex,i] = \vertexindex_0$ }
        \State \textcolor{myteal}{$\bary'_k \gets 1$}
        \State \textcolor{myteal}{$\p' \gets (\faceindex',\bary')$}
        \State \textcolor{myteal}{$\mathbf{n'} \gets \mathtt{get\_face\_normal}(\mesh, \faceindex')$}
        \State \textcolor{myteal}{$\v' \gets \v - \langle \v, \mathbf{n'}\rangle\mathbf{n'}$}
        \State \textcolor{myteal}{$\v' \gets \frac{\v'}{||\v'||}$}
        \State \textcolor{myteal}{\Return $\p',\v',0$}
    \EndIf

    \State $\vertexindex_1,\vertexindex_2 \gets \left\{\Faces[\faceindex, \bary_i] \;\middle|\; i \in [0,2],\; i \neq \text{argmax}_k \bary_k \right\}$
    \State $\mathbf{n} \gets \mathtt{face\_normal}(\mesh, \faceindex)$
    
    \State $\alpha \gets \left|\mathtt{signed\_angle}(-\v, \Vertices[\vertexindex_1]-\Vertices[\vertexindex_0], \mathbf{n})\right|$
    \If{$\alpha > \theta(\vertexindex_0) / 2$}
        \State $\alpha \gets \left|\mathtt{signed\_angle}(\Vertices[\vertexindex_2]-\Vertices[\vertexindex_0], \mathbf{v}, \mathbf{n})\right|$
        \State $\vertexindex_1,\vertexindex_2 \gets \vertexindex_2,\vertexindex_1$
    \EndIf
    \While{$\alpha < \theta(\vertexindex_0) / 2$}
        \If{there is no adjacent triangle}
            \State $\p \gets (\faceindex, \bary)$
            \State \Return $\p,\v, 0$
        \EndIf
        \State $\faceindex' \gets \mathtt{connected\_face}(\mesh, \faceindex,\vertexindex_0,\vertexindex_1)$
        \State $\vertexindex_2 \gets \mathtt{next\_edge\_vertex}(\mesh, \faceindex',\vertexindex_0, \vertexindex_1)$
        \State $\mathbf{e}_1 \gets \Vertices[\vertexindex_1]-\Vertices[\vertexindex_0]$; $\mathbf{e}_2 \gets \Vertices[\vertexindex_2]-\Vertices[\vertexindex_0]$
        \State $\alpha \gets \alpha + \left|\mathtt{signed\_angle}(\mathbf{e}_1,\mathbf{e}_2, \mathbf{n})\right|$
        \State $\faceindex \gets \faceindex'$; $\vertexindex_1 \gets \vertexindex_2$
    \EndWhile
    \State $\beta \gets \alpha - \theta(\vertexindex_0) / 2$
    \State $\mathbf{n} \gets \mathtt{face\_normal}(\mesh, \faceindex)$
    \State $\v \gets \mathtt{rotate\_vector}(\v, \mathbf{n}, \beta)$
    \State $\bary \gets (0,0,0)$
    \State \textcolor{mypink}{$k \gets \text{argwhere}_i\; \Faces[\faceindex, i] = \vertexindex_0$}
    \State \textcolor{mypink}{$\bary_k \gets 1$}
    \State $\p \gets (\faceindex, \bary)$
    \State \Return $\p,\v, 0$
    \end{algorithmic}
\end{algorithm}

%% file: algorithms/transport_edge.tex
\begin{algorithm}
    \caption{$\mathtt{transport\_over\_edge}$. (we highlight in \textcolor{mypink}{pink} the non differentiable steps and in \textcolor{myteal}{blue} the hole avoidance modifications).}
    \label{alg:sg-step-edge}
    \begin{algorithmic}[0]
    \Require mesh $\mesh$, face $\faceindex$, barycentric coordinates $\bary$, direction $\v$
    \State \hrulefill 
    \State $(\Vertices, \Faces) \gets \mesh$
    
    \If{there is no adjacent triangle}
        \State \textcolor{myteal}{Select $\vertexindex'$ one of the vertices on the edge}
        \State \textcolor{myteal}{$\bary' \gets (0,0,0)$}
        \State \textcolor{myteal}{$k \gets \text{argwhere}_i \Faces[\faceindex,i] = \vertexindex'$}
        \State \textcolor{myteal}{$\bary'_k \gets 1$}
        \State \textcolor{myteal}{$\p' \gets (\faceindex, \bary')$}
        \State \textcolor{myteal}{\Return $\p',\v, ||\p'-\p||$}
    \EndIf
    
    \State $\vertexindex_0,\vertexindex_1 = \left\{ \Vertices[\Faces[\faceindex, \bary_i]] \;\middle|\; i \in [0,2],\; \mathbf{b}_i \neq 0 \right\}$

    \State $\faceindex' \gets \mathtt{connected\_face}(\mesh, \faceindex,\vertexindex_0,\vertexindex_1)$
    \State $\mathbf{n} \gets \mathtt{face\_normal}(\mesh, \faceindex)$
    \State $\mathbf{n'} \gets \mathtt{face\_normal}(\mesh, \faceindex')$
    \State $\mathbf{e} \gets (\Vertices[\vertexindex_1]-\Vertices[\vertexindex_0]) / ||\Vertices[\vertexindex_1]-\Vertices[\vertexindex_0]||$
    \State $\alpha \gets \mathtt{signed\_angle}(\mathbf{n},\mathbf{n}', \mathbf{e})$
    \State $\v' \gets \mathtt{rotate\_vector}(\v, \mathbf{e}, \alpha)$
    \State \textcolor{mypink}{Let $\bary'$ the new barycentric coordinates of $\p$ in $\faceindex'$}
    \State $\p' \gets (\faceindex', \bary')$
    \State \Return $\p',\v', 0$
    \end{algorithmic}
\end{algorithm}

%% file: algorithms/projection_exp_map.tex
\begin{algorithm}
    \caption{Projection Integration (PI) Exp map~\cite{Madan2025local-parameterizations}}
    \label{alg:projection-exp}
    \begin{algorithmic}[0]
    \Require Mesh $\mesh$, point $\p$, tangent vector $\v$, step size $s$
    \State \hrulefill
    \State $\faceindex, \bary \gets \p$
    \State $l \gets 0; L \gets ||\v||$
    \While{$l < L$}
        \State $\p' \gets \p + s \v$
        \State $\p', \faceindex' \gets \mathtt{project\_to\_mesh}(\mesh, \p')$
        \State $\mathbf{n} \gets \mathtt{face\_normal}(\mesh,\faceindex)$ 
        \State $\mathbf{n}' \gets \mathtt{face\_normal}(\mesh,\faceindex')$
        \State $\alpha \gets \cos^{-1}(\mathbf{n} \cdot \mathbf{n}')$
        \State $\v \gets \mathtt{rotate\_vector}(\v, \mathbf{n} \times \mathbf{n}', \alpha)$
        \State $l \gets l+||\p'-\p||$
        \State $\p\gets\p';\faceindex\gets\faceindex'$
    \EndWhile
    \State \Return $\x$
    \end{algorithmic}
\end{algorithm}

%% file: algorithms/ot_exp.tex
\begin{algorithm}
    \caption{MeshFlow}
    \label{alg:ot-exp}
    \begin{algorithmic}[0]
    \Require Mesh $\mesh$, distance $d_{BH}$, base $\mathcal{P}$, target $\mathcal{Q}$
    \State \hrulefill
    \State Initialize parameters $\theta$ of $\v_\theta$
    \While {not converged}
        \State $\p_1,\dots, \p_{B}  \sim \mathcal{P}(\mesh)$ (sample noise)
        \State $\q_1,\dots, \q_{B} \sim \mathcal{Q}(\mesh)$ (sample training examples)
        \State $\sigma \leftarrow$ \cref{eq:coupling} (OT couplings)
        \State $\hat{\q}_i \leftarrow \Exp^{\circlearrowright K}_{\p_i} (\v_\theta), \forall i = 1,\dots,B$
        \State $\mathcal{L} \leftarrow \frac{1}{B} \sum_{i=1}^B d_{BH}^2(\q_{\sigma(i)},\hat{\q}_i)$
        \State $\theta\leftarrow \mathtt{optimizer\_step}(\nabla_\theta \mathcal{L})$
    \EndWhile
    \end{algorithmic}
\end{algorithm}

%% file: algorithms/rfm.tex
\begin{algorithm}
    \caption{RFM~\cite{riemannian-fm} on $\mesh$}
    \label{alg:rfm}
    \begin{algorithmic}[0]
    \Require Mesh $\mesh$, base $\mathcal{P}$, target $\mathcal{Q}$, scheduler $\kappa$
    \State \hrulefill
    \State Initialize parameters $\theta$ of $\v_\theta$
    \While {not converged}
        \State $t_1,\dots, t_B \sim \mathcal{U}(0,1)$ (sample time)
        \State $\p_1^{(0)},\dots, \p_B^{(0)}  \sim \mathcal{P}(\mesh)$ (sample noise)
        \State $\p_1^{(1)},\dots, \p_B^{(1)} \sim \mathcal{Q}(\mesh)$ (sample training examples)
        \State $\p^{(t_i)}_i \leftarrow \mathtt{solve\_ODE}\big(t_i, \p_i^{(0)}, \mathbf{u}^{(t)}(\p | \p_i^{(1)})\big), \forall i$ 
        
        \State $\mathcal{L} \leftarrow \frac{1}{B} \sum_{i=1}^B \| \v_\theta^{(t_i)}(\p_i^{(t_i)}) - \dot{\p}^{(t_i)}_i \|^2_2$
        \State $\theta\leftarrow \mathtt{optimizer\_step}(\nabla_\theta \mathcal{L})$
    \EndWhile
    \end{algorithmic}
\end{algorithm}

%% file: algorithms/mlbfgs.tex
\begin{algorithm}
    \caption{Mesh-LBFGS}
    \label{alg:mlbfgs}
    \begin{algorithmic}[0]
        \Require{Mesh $\mesh$; vector transport $\prod$; exponential map $\Exp$; initial value $\mathbf{S}_0 \in \mesh^S$; smooth function $f:\mesh^S\to\R$}
        
        \State \hrulefill
        
        \State Set $H_{\mathrm{diag}} \gets 1$\;
        
        \For{$t = 0, 1, \dots$}
          \State $\mathbf{V}_t \gets \mathtt{desc}(-\nabla f(\mathbf{S}_t), t)$ \textcolor{gray}{//descent direction}
          \State Use line-search to find $\alpha$ satisfying Wolfe conditions\;
          \State $\mathbf{S}_{t+1} \gets \Exp_{\mathbf{S}_t}(\alpha \mathbf{V}_t)$\;
          \State $\mathbf{A}_{t+1} \gets \prod_{\mathbf{S}_t}^{\mathbf{S}_{t+1}} (\alpha \mathbf{V}_t)$\;
          \State $\mathbf{B}_{t+1} \gets \nabla f(\mathbf{S}_{t+1}) - \prod_{\mathbf{S}_t}^{\mathbf{S}_{t+1}} (\nabla f(\mathbf{S}_t))$\;
          \State $H_{\mathrm{diag}} \gets \dfrac{\langle\mathbf{A}_{t+1}, \mathbf{B}_{t+1}\rangle}{\langle\mathbf{B}_{t+1}, \mathbf{B}_{t+1}\rangle}$\;
        \EndFor
        \State Return $\mathbf{S}_{t+1}$\;
    \end{algorithmic}
\end{algorithm}

\begin{algorithm}
    \caption{$\mathtt{desc}$}
    \label{alg:desc}
    \begin{algorithmic}[0]
        \Require Vector $\mathbf{V} \in \mathcal{T}_{\mathbf{S_t}}\mesh^S$, iteration $t$
        
        \State \hrulefill
        
        \If{$t > 0$}
            \State $\tilde{\mathbf{V}} \gets \mathbf{V} - \dfrac{\langle\mathbf{A}_t, \mathbf{V}\rangle}{\langle\mathbf{B}_t, \mathbf{A}_t\rangle} \, \mathbf{B}_t$
            \State $\hat{\mathbf{V}} \gets \prod_{\mathbf{S}_{t-1}}^{\mathbf{S}_t}\!\left(
                \mathtt{desc}\bigl(^*\prod_{\mathbf{S}_{t-1}}^{\mathbf{S}_t} \tilde{\mathbf{V}}, t - 1\bigr)\right)$
            \State \textcolor{gray}{
                // with $^*\prod$ the adjoint of $\prod$ 
            }
            \State \Return $\hat{\mathbf{V}}
                - \dfrac{\langle\mathbf{B}_t, \hat{\mathbf{V}}\rangle}{\langle\mathbf{B}_t, \mathbf{A}_t\rangle} \, \mathbf{A}_t
                + \dfrac{\langle\mathbf{A}_t, \mathbf{A}_t\rangle}{\langle\mathbf{B}_t, \mathbf{A}_t\rangle} \, \mathbf{V}$
        \Else
            \State \Return $H_{\mathrm{diag}}\, \mathbf{V}$
        \EndIf %
    \end{algorithmic}
\end{algorithm}